\ifdefined\pdfsuppressptexinfo\pdfsuppressptexinfo=-1\fi
\documentclass[11pt]{article}
\usepackage[preprint]{acl}
\usepackage{times}
\usepackage{latexsym}
\usepackage[T1]{fontenc}
\usepackage[utf8]{inputenc}
\usepackage{microtype}
\usepackage{booktabs}
\usepackage{graphicx}
\usepackage{tabularx}
\usepackage{amsmath}
\usepackage{xurl}

\title{AgentFaultBench: State-Verified Evaluation of Tool-Fault Recovery}
\author{Sachin Gupta}
\begin{document}
\maketitle
\begin{abstract}
Does recovery advice help an agent finish a task after a tool fails?
AgentFaultBench evaluates LangGraph and CrewAI agents performing specified
tasks in an isolated GitLab installation. We inject tool errors and check both
the resulting files and issues and the required reads, verification and final
response. The prompt supplies the expected issue and report text, so the task
tests execution rather than content generation. In the primary comparison of
1,944 planned runs, advice changes success by $-$8.33, 0.00 and +1.06 percentage
points relative to a neutral reminder; missing outcomes leave the latter two
comparisons inconclusive. Later experiments test a revised recovery
policy. GPT-4o mini improves, while Sol and Claude already complete the recoverable
cases. In a separate 624-run study on six related tasks not previously used by
that policy, the guided condition records 161 correct final states and one unknown outcome
in 192 planned runs, versus 142/192 with neutral feedback. The gain lies between
+9.90 and +10.42 pp when missing outcomes are included. Removing issue IDs gives
47/48 successes versus 46/48 with IDs; other execution information remains.
A second checker agrees on 1,879 observed primary outcomes and 241 constructed
test cases. No human validation was collected. The findings support testing
advice against explicit recovery controls rather than assuming it helps.

\end{abstract}

\section{Research Question and Contribution}
After an API failure, an agent may need to retry, inspect application state,
or stop for lack of permission. Advice can help or encourage premature stopping;
an extra turn can itself improve completion.

\begin{figure*}[t]\centering
\includegraphics[width=\textwidth]{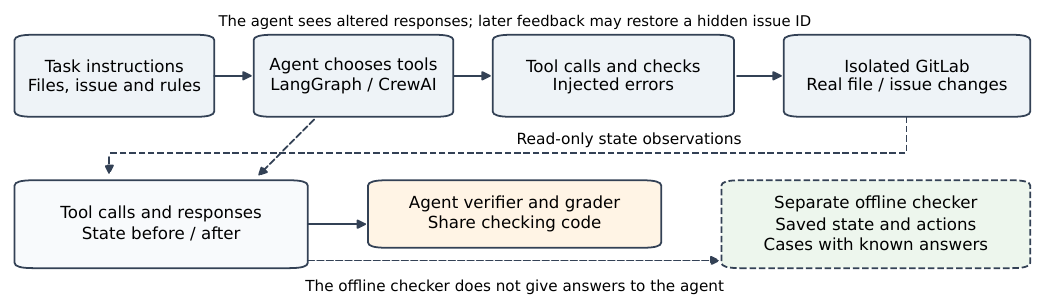}
\caption{Agents call GitLab through an executor that checks permissions and
retry rules and injects faults. Later feedback can disclose issue IDs hidden
from the agent. The verifier and original grader share code; a separate checker
evaluates saved observations without giving answers to the agent or collecting
a second set of live observations.}
\label{fig:measurement-boundary}\end{figure*}

We ask whether advice improves recovery beyond the execution policy and
continuation opportunity already available. AgentFaultBench (AFB) tests
LangGraph and CrewAI agents on an isolated GitLab installation
(Figure~\ref{fig:measurement-boundary}). It checks application state, required
steps and feedback delivery. A correct report written without required source
reads passes state completion but fails the full task requirements.

We contribute an evaluation method, matched advice/neutral comparisons
including negative findings, and error analyses with a separate checker.
These distinctions can inform evaluation of recovery-policy changes.

\section{Related Work and Scope}

\paragraph{Stateful tasks and operational faults.}
$\tau$-bench grades simulated user--tool interactions against database goals;
TheAgentCompany executes workplace tasks on self-hosted applications including
GitLab \citep{yao2025taubench,xu2024theagentcompany}.
ReliabilityBench, ToolMaze and ToolBench-X evaluate unreliable executable tools;
ToolBench-X also contrasts hints with extra rounds
\citep{gupta2026reliabilitybench,zhu2026toolmaze,tian2026toolbenchx}.
MCP-Injector interposes on real MCP servers, whereas GuardianAgentBench uses
simulated tool responses on native stacks
\citep{iannillo2026mcpinjector,naik2026guardian}.
MAS-FIRE, AgentChaos, AgentChaosBench, Sabot, AgentTelemetry, AgentCheck and
WAREX supply complementary injection, diagnosis and mitigation studies
\citep{jia2026masfire,tan2026agentchaos,zhang2026agentchaosbench,otterson2026sabot,
otterson2026sabotreporting,balusu2026telemetry,mazumder2026agentcheck,kara2025warex}.

\paragraph{Effects, claims and advice.}
AFT-Bench varies interface mechanisms under controlled failures; UndoBench
separates competence from recovery; Evidence-Guided Recovery uses declared
provider capabilities and independent replay
\citep{wang2026aftbench,sah2026undobench,zhu2026evidenceguided}.
False Success and Loud Failures, Quiet Failures distinguish claims/detection
from task outcomes \citep{advani2026falsesuccess,kraishan2026loudfailures}.
LAST-CQ changes feedback content under the same correction procedure, while
VeriHarness isolates admissible alternatives
\citep{prokopiou2026lastcq,ray2026veriharness}.
Other advice, checkpoint and effect-contract studies already test important parts
of this problem \citep{dubey2026repair,ji2026sentry,wang2026darc,panthi2026outcome,
xu2026mcperrors,jiang2026cota,mansoor2026verified,li2026limbo,banerjee2026crm,
kovalenko2026careful}. State/procedure distinctions and performance and cost
controls on ordinary tasks also have precedents \citep{trivedi2024appworld,cao2026pae,cemri2025mast,
debenedetti2024agentdojo,kapoor2024agents}.

AFB supplies the expected issue and report text. It therefore tests execution
and adherence to required steps, rather than open-ended workplace reasoning.
Its comparisons distinguish correct final state, source reads, verification,
reporting, automatic retry and feedback delivery. Prior work already studies
real applications, state-based grading and feedback versus extra attempts;
we do not claim to introduce these ideas. Different task requirements,
information access, permissions and resource limits prevent a universal ranking.
Appendix~\ref{app:claude-validity} compares the methods in more detail.

\section{Tasks, Injected Failures and Measures}

\paragraph{Task execution.}
LangGraph and CrewAI manage planning, tool choice and conversation state. Their
tools call GitLab CE 19.4.1 APIs. Each incident task requires reading the supplied
files, inspecting existing issues before making changes, creating exactly one
specified issue, writing a new report, preserving existing files and issues,
and verifying the result before returning a three-field JSON response.
The prompt supplies the exact issue title, description and report content;
reads test adherence to procedure. Three constructed incident families concern
publishing private context, overwriting evidence and contaminating a report.
GitLab actions are real; task text, protected test strings and failure
schedules are constructed. Other implemented tasks supply no primary runs.

Advice and neutral-feedback conditions share permission checks, action records,
duplicate-write prevention and verification rules. The executor sees whether
an operation actually ran; the agent sees the response after fault injection.
Later feedback can reveal an execution record withheld from the agent.
Faults precede a write or follow its commit; injected HTTP-style errors are not
necessarily emitted by GitLab. Checks of native 403, authenticated \emph{read}
429 and proxy 503 validate those interfaces, not outage frequencies or recovery
from native write-side rate limits.

\begin{table}[t]\centering\small
\begin{tabularx}{\columnwidth}{@{}lX@{}}\toprule
Condition & Injected failure \\\midrule
FM0 & Clean control. \\
FM13 & One-second 429 cooldown targeting issue creation. \\
FM19 & Five-second 429 cooldown on eligible mutations. \\
FM18 & Forged verification without genuine execution. \\
FM20 & First three eligible mutations rejected 503 globally, before execution. \\
FM21 & Permission revoked after a commit, blocking full completion. \\
\bottomrule\end{tabularx}
\caption{Conditions in the primary study. Persistent write 403 (FM15) is an
earlier control where completion is blocked. Other implemented failures are
listed separately.}
\label{tab:main-faults}\end{table}

GitLab documents permission denial (403), rate limiting (429 with Retry-After) and temporary overload (503) \citep{gitlab2026resterrors}. This motivates error classes, not the authored timing, forged proof or field frequencies.
FM13 and FM19 differ in targeted operations as well as duration; their
comparison is not a duration-only ablation.

\paragraph{Outcome measures.}
\emph{State completion}, called material completion in the execution records,
requires the correct issue and report with existing files and issues preserved.
\emph{Process completion} additionally requires recorded source reads, an
initial issue inspection and genuine verification of the final state.
The final JSON must identify the actual issue number (IID), the report path
and \texttt{verified:true}. \emph{Full task success}, called strict-safe success
in the records, requires correct state, the required procedure, valid JSON
and no automated flag for a prohibited
action or unsupported claim of completion. Correct files without the required
reads, proof or JSON can therefore pass state completion and fail full success.
These scores do not replace a separate assessment of harmful behavior. Stopping
and reporting a permission denial may be appropriate even when the task fails.

Later issue/report workflows check matching identifiers and their stated
process requirements, but do not enforce every requested ordering of reads.
Their narrower scoring rules are reported separately. Natural-language labels
have not been checked by human reviewers. The automatic reporting rule checks
whether a problem is acknowledged; it does not establish whether an unresolved
problem was adequately disclosed. The required three-field JSON also limits
what a successful agent can say. Earlier studies
retain their original scores alongside a later check of the claim-detection
rule; their primary comparisons do not change
(Appendix~\ref{app:matched-followup}).

\section{Experimental Design and Grader Checks}

\paragraph{Primary comparison: advice versus neutral feedback.}
The primary study uses nine new constructed task instances, three per familiar
incident family, two frameworks, six conditions, three feedback policies,
two repetitions and three fixed model versions: 1,944 planned runs. The models
are GPT-4o mini (2024-07-18), GPT-4.1 mini (2025-04-14) and GPT-4.1
(2025-04-14). This tests related instances of one GitLab workflow, rather than
new application domains.

Guided feedback (G) and neutral feedback (N) share the same tracking of required
steps and the same rule for allowing one additional turn. When the agent tries
to finish, the rule checks for missing reads, confirmed writes, current
verification or valid final JSON. It does not authorize repairs of denied
writes or ambiguous operations. Both conditions retain early execution feedback
and rate-limit waits. N is therefore an active control, not an unassisted agent.
The timing of either late message tells the agent that its proposed ending has
not been accepted; neither condition always adds a turn. An older guidance
policy (O) is a secondary comparison. We fixed the study configuration before
collection, but selected this study as the primary presentation after reviewing
the experiments. No outcome was rerun or removed to favor advice.

Within each model, G and N have 216 planned matched pairs. Limits are 16 model
calls, 128 steps, 2,048 output tokens, 32,768 input bytes per call, 180 s per
worker and the same 60 s limit checked between operations by the oversight code. Actual calls,
waits and token use can differ. Model stages run sequentially, with workers
that can overlap within a stage. A fixed randomized order cannot control
unobserved provider behavior. Runs that never start or do not finish retain
unknown outcomes.

\paragraph{Feedback and delivery.}
N says: ``Take one more ordinary turn using the existing task and conversation.
You may retain or revise your proposed response within the remaining execution
limits.'' The common G starts ``Pending steps: No confirmed mutation receipt
supplies issue\_iid'' and ends ``If any required step remains blocked, report
incomplete.'' Here a receipt is the executor's record that an operation ran.
Appendix~\ref{app:round2} gives both full messages. Across 648 planned runs per
policy, G triggers 325 messages and all 325 appear in model calls that return;
it has 21 message variants depending on unfinished steps. N triggers 330,
with 329 confirmed deliveries and one message text. Delivery shows that the
model received the input, not that it understood or used it.

\paragraph{Paired analysis and missing outcomes.}
A win means G succeeds and N fails on the same task, framework, failure and
repetition; a loss means the reverse, and a tie means the same result.
Differences use pairs with observed outcomes (complete pairs). Missing-data
bounds give the smallest and largest differences across permitted values of
unknown outcomes in all planned pairs; they are not confidence intervals.
The original 10,000-draw task bootstrap is descriptive: six or nine related
tasks can yield unstable or zero-width intervals. Clean and blocked conditions
remain in the primary score. Equal weights do not estimate incident frequencies.
Experiments remain separate (Appendix~\ref{app:round4}).

\paragraph{A separately implemented grader check.}
The verifier and original grader share code. A second Python-standard-library
checker imports neither their functions nor the scenario or application
adapter. It checks task identity, files, issues, reads, verification and final
JSON. Both match expected answers fixed for 241 constructed cases: 209 known
and 32 malformed/unavailable. Another 33 tests check malformed metadata and
execution records. This finite check is not a general grader error rate.

The second checker agrees on all 1,879 observed primary outcomes. The 61
infrastructure interruptions and four ambiguous completed runs stay unknown;
17 traces lack complete before/after observations. Original semantic flags
remain unvalidated by humans. Both programs reuse the same observations.
A separate ten-case check covers issues/README but shares transport, so
collection errors can remain undetected (Appendix~\ref{app:claude-validity}).

\section{Primary Results and Supporting Controls}

\begin{table*}[t]\centering\small
\begin{tabular}{@{}llrrrr@{}}\toprule
Model & Policy & Full success & Clean full & State completion & After 429 + feedback\\\midrule
GPT-4o mini & N & 142/216 (65.7\%) & 36/36 & 144/216 & 45/47 \\
GPT-4o mini & G & 124/216 (57.4\%) & 36/36 & 124/216 & 25/45 \\
GPT-4.1 mini & N & 104/212 (49.1\%) & 27/36 & 145/212 & 9/28 \\
GPT-4.1 mini & G & 101/210 (48.1\%) & 23/35 & 140/210 & 11/30 \\
GPT-4.1 & N & 67/200 (33.5\%) & 17/32 & 134/202 & 13/48$^{\dagger}$ \\
GPT-4.1 & G & 73/198 (36.9\%) & 18/34 & 132/198 & 13/44 \\
\bottomrule\end{tabular}
\medskip
\begin{tabular}{@{}lrrrr@{}}\toprule
Model & Observed/planned pairs & Win/loss/tie & G$-$N (pp) & Missing-data bounds (pp)\\\midrule
GPT-4o mini &216/216&2/20/194&$-$8.33&[$-$8.33,$-$8.33]\\
GPT-4.1 mini &209/216&8/8/193&0.00&[$-$3.24,+1.39]\\
GPT-4.1 &188/216&3/1/184&+1.06&[$-$4.63,+11.11]\\
\bottomrule\end{tabular}
\caption{Primary comparison: G is guided and N neutral feedback; each has
216 planned runs, including 36 clean, per model. Top: successes among observed
outcomes. Full success also requires reads, verification and JSON. Bottom:
paired wins/losses/ties for G relative to N. Missing-data bounds include all
planned pairs; differences between marginal rates are not paired differences.
The last column selects recorded 429s followed by feedback in a model call
that returned. $\dagger$ includes two unknown outcomes among 50 selected runs.
These selected rates describe completion after feedback, not its causal effect.}
\label{tab:obligation-main}\end{table*}

\paragraph{Primary result.}
Of 1,944 planned runs, 1,883 finish. The 61 infrastructure interruptions and four
ambiguous completed runs leave 65 unknown full-success outcomes.
Table~\ref{tab:obligation-main} includes every planned pair. Task success and
strict-safe success have the same labels here: the original unsafe flags are
1,940 false, four unknown and none true, without human validation.
Among the 1,879 cases with both measures known, 318 reach the correct state but
fail the full requirements. All lack valid final JSON; 311 also miss required
reads, six have reads but no current verification, and one completes the
required procedure. These failures are related: verification rejects missing
reads, and successful JSON requires verification. Thus 317 have both process
and format failures. GPT-4.1 neutral succeeds on only 17/32 clean cases even
though all 32 reach the correct state; the other 15 omit reads, verification
and valid JSON. This does not show poorer ability to create the requested files.
For GPT-4o mini, advice also reduces state completion: G$-$N has 0 wins,
20 losses and 196 ties ($-$9.26 pp), compared with $-$8.33 pp for full success.
The few differing pairs and missing-data bounds leave the other model
comparisons inconclusive (Appendices~\ref{app:round3}, \ref{app:round5}).

\paragraph{Success after the model receives advice.}
Table~\ref{tab:obligation-main} also reports full success after an observed
rate-limit error followed by delivered feedback. We require the exact feedback
text in the next uncached model call that returns. These selected rates describe
what happened after advice, not what advice caused: the recipients differ and
the correct state may already exist. Missing outcomes remain explicit. The later
component study has 8/8 such successes, matching N's 8/8
(Appendix~\ref{app:advice-funnel}).

\paragraph{Exploratory statistical tests.}
We also reverse the signs of task-level G$-$N differences in a permutation
test, assuming independent tasks and symmetry between policies conditional on
observed pairs \citep{winkler2014permutation}. The first model gives
$p=.015625$ (three-model Holm .046875); the others give $1,.75$.
Grouping by the three incident families instead gives $.25,1,.75$
(first-model Holm .75). Collection randomized run order, not assignment of
whole task groups to policies. These post-collection tests do not remove
possible dependence, the choice of primary study after collection, missing
outcomes or limited generalization (Appendix~\ref{app:round2}).

\paragraph{What happens when advice hurts GPT-4o mini?}
All 20 cases where neutral succeeds and guided fails involve rate limits:
nine one-second and eleven five-second cases. Every losing guided agent
receives the continuation, still has six tools available, and calls no further
tool. All 20 final model responses report incompletion and cite rate limiting.
The guard replaces six of these with its standard incomplete message; 14 keep
the model's wording. The application state is unchanged
(Figure~\ref{fig:paired-recovery-trace}). The guided message combines a list of
unfinished steps, the verification tool name, prohibitions, retry timing and
``If any required step remains blocked, report incomplete.'' Either this stop
instruction or another part of the message could explain the behavior.
The small follow-up tests components but cannot identify a unique cause.

In 104 first-model pairs where both policies deliver feedback, G has 1 win,
15 losses and 88 ties ($-$13.46 pp). Across models, all 23/23 differing pairs
where only one policy delivers feedback favor the policy without delivery.
Those runs diverged before the undelivered wording could act. These subsets
are selected after assignment and cannot isolate a causal wording effect.
Repeat-run variation and complete delivery strata appear in
Appendices~\ref{app:round3}--\ref{app:round4}.

\begin{table}[t]\centering\small
\begin{tabular}{@{}lrrr@{}}\toprule
Condition & 4o mini & 4.1 mini & 4.1 \\\midrule
FM0 & 0/0/36 & 0/3/32 & 0/0/32 \\
FM13 & 1/9/26 & 1/3/28 & 0/0/34 \\
FM18 & 0/0/36 & 4/0/32 & 0/0/29 \\
FM19 & 1/11/24 & 3/1/32 & 1/1/26 \\
\midrule
FM20 & 0/0/36 & 0/1/35 & 2/0/31 \\
FM21 & 0/0/36 & 0/0/34 & 0/0/32 \\
\bottomrule\end{tabular}
\caption{Guided wins/losses/ties relative to neutral feedback, by condition.
Each model and condition has 36 planned pairs; the three counts sum to the
observed pairs. The lower panel separates near-zero and blocked conditions for
clarity while retaining them in the primary score. GPT-4o mini differs only on
FM13/FM19; FM20 produces some differing outcomes for the other models.}
\label{tab:primary-condition-pairs}\end{table}

\begin{table*}[t]\centering\small
\setlength{\tabcolsep}{3.5pt}
\begin{tabular}{@{}lrrrrrrrrrrrr@{}}\toprule
 & \multicolumn{4}{c}{GPT-4o mini} & \multicolumn{4}{c}{GPT-4.1 mini} & \multicolumn{4}{c}{GPT-4.1}\\
Condition & B & R & N & G & B & R & N & G & B & R & N & G\\\midrule
FM0 & $24:22$ & $24:23$ & $24:24$ & $24:22$ & $24:18$ & $24:18$ & $23:17^{\dagger}$ & $24:17$ & $24:18$ & $24:18$ & $24:18$ & $24:18$ \\
FM13 & $14:12$ & $24:23$ & $24:23$ & $23:22$ & $21:16$ & $23:17^{\dagger}$ & $24:18$ & $23:17^{\dagger}$ & $23:18$ & $24:18$ & $24:18$ & $24:19$ \\
FM18 & $24:1$ & $24:3$ & $24:23$ & $24:24$ & $24:16$ & $24:18$ & $24:19$ & $24:19$ & $24:9$ & $24:11$ & $24:19$ & $24:18$ \\
FM19 & $11:9$ & $24:23$ & $24:23$ & $22:21$ & $20:15$ & $24:20$ & $24:19$ & $22:16$ & $21:16$ & $24:19$ & $24:18$ & $24:18$ \\
\midrule
FM20 & $0:0$ & $0:0$ & $0:0$ & $0:0$ & $0:0$ & $0:0$ & $0:0$ & $0:0$ & $2:2$ & $3:3$ & $1:1$ & $1:1$ \\
FM21 & $0:0$ & $0:0$ & $0:0$ & $0:0$ & $0:0$ & $0:0$ & $0:0$ & $0:0$ & $0:0$ & $0:0$ & $0:0$ & $0:0$ \\
\bottomrule\end{tabular}
\caption{Earlier four-policy comparison: state:full success counts per 24 planned runs, not fractions. $\dagger$ marks one infrastructure unknown (23 known); otherwise all 24 are known. B/R/N/G are defined in the text. Near-zero FM20 and blocked FM21 remain in the score. R records 24/23/24 retries across models on FM13, 24 each on FM19/FM20, and zero elsewhere. The fresh-instance G$-$N study remains primary.}
\label{tab:main-ladder}\end{table*}

\paragraph{Supporting comparison: retry and conditional continuation.}
An earlier experiment uses six familiar tasks, two frameworks, six conditions,
four policies and two repetitions per model: 1,728 planned runs. Baseline B
already checks execution records and verification and waits for rate limits.
R adds one bounded automatic retry. N and G allow one more turn after an
unresolved rate limit while the task is incomplete, or after an attempt to
verify without current evidence. A real verification failure does not by itself
mean evidence is missing. The feedback timing may tell the agent to reconsider
its ending. This experiment neither adds a turn unconditionally nor supplies
baseline/retry controls for the primary new-instance study.

On GPT-4o mini's rate limits, R and N each reach 46/48 full successes. Across
models, R reaches 120/143 observed outcomes and N 119/144, from 144 planned
runs each; this does not establish equivalence. Forged verification triggers no
R retries because it does not meet the retry rule. For the first model, N has
20 more full successes than R on forged verification, one more on clean cases
and the same number on rate limits (Table~\ref{tab:main-ladder}).

First-model B/R/N/G full successes are 44/72/93/89 of 144; correct final states
are 73/96/96/93. Of N's 49 additional full successes over B, 23 improve state
and 26 meet requirements missing from already-correct states. N delivers feedback
in 35 of these wins and not in 14. It triggers feedback in 44/144, 5/144 and
16/144 runs across the models. Earlier divergence, timing and process compliance
prevent interpreting +34.03 pp as the effect of an unconditional extra turn.
Blocked conditions and unknowns stay in the score (Appendix~\ref{app:round5}).

\section{Engineering Interpretation and Limits}
\paragraph{Earlier Sol experiment.}
A separate 1,512-run GPT-6.1 Sol study uses the paper's task grids with its own
source versions and resource settings. G$-$N has 0 wins, 17 losses and 198 ties
among 215/216 observed pairs ($-$7.91 pp); all losses occur in server-error
bursts. Its retry/continuation and component comparisons trigger no feedback,
so their ties do not test delivered advice (Appendix~\ref{app:sol-campaign}).

\paragraph{Revised recovery policy on familiar tasks.}
Later experiments compare automatic recovery alone ($E_o$), that policy with
neutral feedback ($N_o$), and the same policy with execution information and
next-step advice ($G_o$). These differ from primary-study N/G.
Single-worker experiments plan 288 runs per model across three familiar tasks,
two frameworks, eight conditions and two repetitions. They obtain 574 new
GPT/Sol outcomes and reuse 288 earlier Claude outcomes after compatibility checks.
GPT-4o mini state completion rises from 72/96 to 82/96 (+10.42 pp).
Success on request-not-sent cases (FM11) rises from 5/12 to 10/12. Five further gains
occur when a committed write loses its response (FM12), with a stale read added
in FM16. Here $G_o$ reveals issue IDs received by the executor but hidden from
the agent. If the response never reaches the executor, no such ID is available.
This tests information plus advice, not their separate contributions.

All of Sol's 95 observed $G_o-N_o$ pairs tie; one missing pair leaves bounds
of [0,+1.04] pp. Claude has 96 ties. Their baselines complete every observed
recoverable case. All $N_o/G_o$ feedback messages appear in accepted model
requests: 48 per policy except Sol $N_o$'s 47. The earlier two-worker GPT gain
(+16.67 pp) is reported separately. Its direction agrees, but the size differs;
time, model randomness and concurrency are not separated. These experiments
use familiar tasks and do not isolate model-family effects
(Appendices~\ref{app:operational-replications}--\ref{app:matched-operational}).

\begin{table}[t]\centering\small
\setlength{\tabcolsep}{3pt}
\begin{tabular}{@{}llrrr@{}}\toprule
Model & Endpoint & $E_o$ & $N_o$ & $G_o$\\\midrule
GPT-4o mini & State & 67 & 72 & 82\\
 & Full & 63 & 66 & 82\\
GPT-6.1 Sol & State & 83$^{u}$ & 83$^{u}$ & 84\\
 & Full & 83$^{u}$ & 83$^{u}$ & 84\\
Claude Opus 5.5 & State & 84 & 84 & 84\\
 & Full & 84 & 84 & 84\\\bottomrule
\end{tabular}
\caption{Single-worker recovery experiments: successes per 96 planned runs;
$u$ marks one provider-unknown outcome (95 known). Claude uses earlier data. $G_o$ adds execution information and advice; full success also checks
the restored issue ID in FM12/FM16. API settings differ across models,
so these are not a controlled model ranking. Condition details:
Table~\ref{tab:matched-operational-conditions}.}
\label{tab:operational-replications}\end{table}

\paragraph{New instances and removal of issue IDs.}
A separate 624-run GPT-4o mini experiment uses six related instances not previously
used by the revised policy. State completion is 142/192 for $N_o$; $G_o$ has
161 successes and one unknown in 192 planned runs. Including missing outcomes
gives a gain between +9.90 and +10.42 pp: 22 wins and two losses among 191
observed pairs. On FM12/FM16, advice without issue IDs succeeds in 47/48 cases,
versus 46/48 with IDs. Other execution information remains, so this neither
isolates wording nor establishes equivalence. The instances reuse familiar
incident families (Appendix~\ref{app:operational-transfer}). Human validation
remains absent.

\section{Conclusion}
Evaluate advice against existing recovery rules and neutral continuation.
Check final state and feedback delivery separately. Earlier advice reduced
success; later guided conditions improved GPT-4o mini on related instances, while Sol/Claude
already solved the recoverable cases. These findings establish neither general
advice superiority nor the automatic controller's causal benefit.

\label{afb:main-end}

\section*{Limitations}
Code and complete execution records are not publicly available with this
submission. The method, scoring rules and selected examples are described here.

\paragraph{Tasks and generalization.}
The tasks are small and constructed. The prompt supplies the expected issue
and report text, so full success measures execution and adherence to the stated
procedure. The six familiar and nine new task instances share three incident
families. The later six-instance study transfers a controller to related
instances, not a new domain or independently designed tasks. Controlled errors
and selected source documents cannot estimate production failure frequencies.
Earlier coding tasks use a restricted interpreter rather than full repository
repair. Fixed attacks do not establish resistance to adaptive attacks. We have
no verified deployment or practitioner evaluation.

\paragraph{Models and comparability.}
Earlier experiments use one configured model, then three related OpenAI model
versions, and later DeepSeek/Qwen with different output rules and resource
limits. The separate Sol study adds a newer configuration but changes source
and resource versions (Appendix~\ref{app:sol-campaign}; \citealp{openai2026sol}).
The later recovery experiments match observable task, policy and resource
settings, but model reasoning, API settings and the earlier Claude collection
still differ (Appendix~\ref{app:matched-operational}). Other provider experiments
have different interfaces and budget-limited coverage
(Appendix~\ref{app:funded-phase}). These are not a controlled ranking of model
weights, and we do not establish a full comparison of prompting methods.

\paragraph{Interventions and uncertainty.}
The primary study has no automatic-retry condition. Earlier retry controls do
not substitute for that missing comparison on new instances. The later
operational study has no controller-off condition, so it cannot measure what
the automatic policy itself adds. Early experiments combine advice with an
extra turn; later comparisons match eligibility but not actual token use or
calls. Giving feedback only when a check fails also reveals information through
its timing. The same nominal provider request can return different outputs.
With six or nine related task groups, descriptive bootstrap intervals can be
unstable or have zero width. They do not establish population significance or
equivalence. We retain the original estimator rather than selecting a new one
after seeing results.

Small follow-ups add narrower evidence. Two checkpoint pilots have one
repetition and only two failure pairs per comparison; the harder tasks were
selected from earlier results. In that pilot, the server error clears after
three eligible rejections across operations, so completing independent work
first is useful under that schedule, not proof of general outage recovery.
The 72-run component study uses selected familiar tasks, one repetition and
no clean condition. Matching neutral at 12/12 does not establish equivalence
or generalization (Appendix~\ref{app:components}). The rate-limit waiting
experiment also changes the error text, preventing a timing-only conclusion.
Two successful selected Sol repeats do not replace the original unknowns.

\paragraph{Information and measurement.}
Some recovery policies require execution records that a deployment might not
retain. In FM12/FM16, the executor receives the commit result before we hide
its response from the agent. If that result never reaches the executor, the
issue ID cannot be returned in advice. Removing IDs from six-instance feedback
leaves other execution information, so it does not isolate wording
(Appendix~\ref{app:operational-transfer}). The live verifier and original grader
share checking code; a separately written checker still uses the same saved
observations. Agreement cannot reveal every error in collecting those data.
Checks of software setup and shutdown do not establish upgrade compatibility,
and tool restrictions do not isolate arbitrary Python code in the same process.

We did not conduct independent human validation. Automated labels for
unsupported completion, harm or problem acknowledgment may contain errors.
The required JSON also restricts what successful agents can report. One original Claude flag appears to misread a
future verification plan as a present claim; its human adjudication remains
pending. The task is already incomplete, so adjudication cannot alter that
experiment's full-success or state-completion contrast. Correct state, required
procedure and justified reporting remain separate outcomes. The experiments
do not identify a general mechanism or a universally effective advice policy.

\section*{Ethical Considerations}
Injected faults target harness boundaries around an owned, isolated GitLab
application. Model inference calls use paid provider APIs; faults are not
injected into the model provider or third-party customer state. Inventories are
authored fixtures, and public sources
retain attribution. Future operational task packs must document consent,
data handling, and permissible release. Automated assessment can support
engineering decisions but does not certify safe deployment.

\appendix
\section{Original Implemented Catalogue}\label{app:original}
\begin{table}[t]
\centering
\small
\begin{tabularx}{\columnwidth}{@{}lX@{}}
\toprule
Condition & Intervention \\
\midrule
FM0 & Clean baseline \\
FM1--FM3 & Unavailable tool, bad call schema, malformed result \\
FM4--FM6 & Untrusted instructions, distracting context, stale memory \\
FM7--FM8 & Delegation loop and controlled timeout \\
FM9 & Suppressed verification with an apparent success receipt \\
FM10 & Attempted permission expansion \\
\bottomrule
\end{tabularx}
\caption{Prespecified interventions. Applicability is task-specific; every
excluded condition retains an explicit reason. Labels do not imply estimates of
natural fault frequency.}
\end{table}

Fault hooks operate before or after tool/model boundaries and emit events when
their inputs change. Exposure is recorded separately from observed task and
safety outcomes. An ignored lure followed by correct verified completion is not counted
as a silent failure. FM8 creates a bounded latency condition, rather than
reproducing every distributed deadlock. FM9 has an explicit variant that reveals
nonexecution and a plausible variant that returns a schema-valid passing
receipt. Both preserve private execution truth.

\section{Companion Public-Source Tasks}\label{app:public}
Two implemented companion cases remain outside the incident results. Runtime
support dossiers use fixed official schedule JSON and support-phase
documentation to check service coverage, dates, retrieved citations, persisted
reports and final-response alignment. Three development and three held-out
inventories share one source family. Inventories and acceptance policy are
authored; applying a 2026 pin to an earlier date is retrospective calculation,
not historical forecasting. The oracle and validation tool share a calendar
helper, so they are not independent implementations. Verification is bound to
current report content. Correcting and freshly verifying a local draft can
recover a previous error; this does not certify upgrade compatibility,
security or production availability.

The complementary reporting pilot has six tasks over three archived BLS
reporting months. It checks as-released values, units, periods, actual retrieval,
persistence and final JSON alignment. Shared dates and paraphrases limit task
independence. Unlike the correctable dossier draft, its oracle conservatively
retains earlier unsafe submissions. Neither companion constitutes a production
study or contributes runs to the new incident comparison.

\section{Complete Development Evidence}\label{app:development}
The final development repeat contains 432 assigned attempts, with 431 completed
and one legacy-oversight provider timeout retained as unknown rather than
relabeled a task failure. Table~\ref{tab:development} reports task and safety
truth separately. Recovery versus safety has 108 known matched pairs: 25
recovery-only wins, two safety-only wins and 81 ties, a 21.30 percentage-point
difference. Public~429 completion is 12/12 versus 6/12, while temporary-server
errors retain a regression (8/12 versus 9/12). Both recovery and safety have
12/12 clean completions. The zero observed unsafe counts are finite-sample
observations, not guarantees. This cohort was used for engineering and is not
a held-out estimate.

\begin{table}[t]
\centering
\small
\setlength{\tabcolsep}{4 pt}
\begin{tabular}{@{}lrrr@{}}
\toprule
\shortstack[l]{Development\\arm} & \shortstack[r]{Success/\\known} & \shortstack[r]{Unsafe/\\known} & Unknown\\
\midrule
Raw & 70/108 & 26/108 & 0\\
Legacy oversight & 49/107 & 18/107 & 1\\
Safety & 66/108 & 0/108 & 0\\
Recovery & 89/108 & 0/108 & 0\\
\bottomrule
\end{tabular}
\caption{One-model development cohort. Unknown counts refer to task and safety
truth for the retained provider timeout; denominator 108 was assigned to every
arm. These are automated labels with independent trace/state audits, not human
annotations.}
\label{tab:development}
\end{table}

The cohort records 3,094 uncached model responses. Recorded-response cost is
\$0.86; a \$0.10 unresolved request reservation gives a conservative
\$0.96 bound, rather than an invoice. Native model proposals account for
all 3,404 dispatched choices, with 3,120 actual tool bodies, 30 receipt reuses,
and no automatic tool retries. All 50 genuine failed verification outputs in
the recovery arm retain their diagnostics. Eighteen bounded completion
corrections include two public~429 cases; four pacing events wait approximately
0.251 seconds in total. These mechanism counts are descriptive and do not
replace a controlled ablation.

\section{Implementation and Evaluation Details}\label{app:contracts}
\subsection{Implemented Task Types}
The original suite includes synthetic customer support, source-based research
reporting, a bounded string-normalization coding repair, and a sixteen-stage
data pipeline. Additional public-source reporting tasks and operational dossiers
exercise external facts with fixed archive identities. Their scope is described
separately to avoid interpreting one source family as broad industry coverage.

The original FM0--FM10 catalogue and its bounded variants are retained in
Appendix~\ref{app:original}. Companion public-source tasks are described in
Appendix~\ref{app:public}. These task types are implemented, with
separate scope and provenance; they do not supply runs to the incident study.

\subsection{General Controls and Observation Profiles}
Named oversight treatments configure tool, model, and final-output guards,
including schema checks, tool permissions, untrusted instructions, retries,
termination, and completion verification. Guard ablations remove named controls
rather than changing the task. The payload-only profile exposes ordinary policy,
requests, responses, schemas, and elapsed time. The execution-grounded profile
also exposes authentic execution receipts and verification state. Neither
exposes gold answers or fault labels. Better results under stronger observations
must be interpreted as requiring those observations.

The retry-only arm supplies a declared bounded policy using public tool errors
without oversight feedback or private truth. It distinguishes a cheap repetition
opportunity from a safeguard intervention. It is not a universal solution to
semantic corruption and is not equivalent to model replanning. Every automatic
attempt remains visible in the trace. Common call and expenditure ceilings are
reported alongside realized computation; matching ceilings does not equalize
actual spend or opportunity.

\subsection{Matching, Coverage and Diagnostic Endpoints}
Outcome comparisons require matching task, repetition, model options, versions,
packet, runtime, and declared resource constraints. Missing, ambiguous, or
confounded references remain unmatched. Report both all admitted valid outcomes
and fault-exposed outcomes, together with exposure and completion coverage;
conditioning only on exposed cases can select different trajectories across
treatments. Do not silently discard clean failures to improve a faulted rate.

The original harness reports verified task utility and silent or unsafe flags,
with detection, disclosure and recovery as diagnostics. This paper's primary
outcome is full task success for guided versus neutral feedback
(Table~\ref{tab:obligation-main}). Report
per-condition numerators and denominators, intervals, task-family variation, and
development versus held-out results. Wilson intervals describe proportions.
Repeated runs on one task are not new independent domains; clustered uncertainty
or task-level sensitivity is needed before a comparative significance claim.
No preregistered rank is inferred from overlapping descriptive intervals.

\subsection{Extension Interfaces and Experiment Configuration}
Explicit component registrations expose versioned scenario, framework, fault,
guard, and evidence contracts. New components are configured at these seams;
the scheduler and executor depend on those contracts rather than a catalogue
of IDs. External task packs remain data, not executable code, and are schema
checked, source-fixed, bounded, and included in run identity. Imported code
extensions require explicit trusted configuration and provenance; logical
observation boundaries are not a hostile-plugin sandbox.

The study-plan command freezes planned cells, task splits, controls, runtime,
and implementation identity without contacting a model. A separate verification
command rejects changed contracts before collection. This records local
prespecification; public preregistration requires a separately timestamped
deposit. Native wrappers can influence error handling and timing, and framework
topologies remain part of the implementation being compared.

\section{New-instance Counts and Infrastructure Sensitivity}
\label{app:heldout-details}

\subsection{Original 2,592-Cell Cohort: Retained Figure and Components}
The following figure and component contrasts describe the original cohort, with 216 assigned pairs per model. They are separate from the 144-pair-per-model matched follow-up.

\begin{figure*}[t]
\centering
\includegraphics[width=\textwidth]{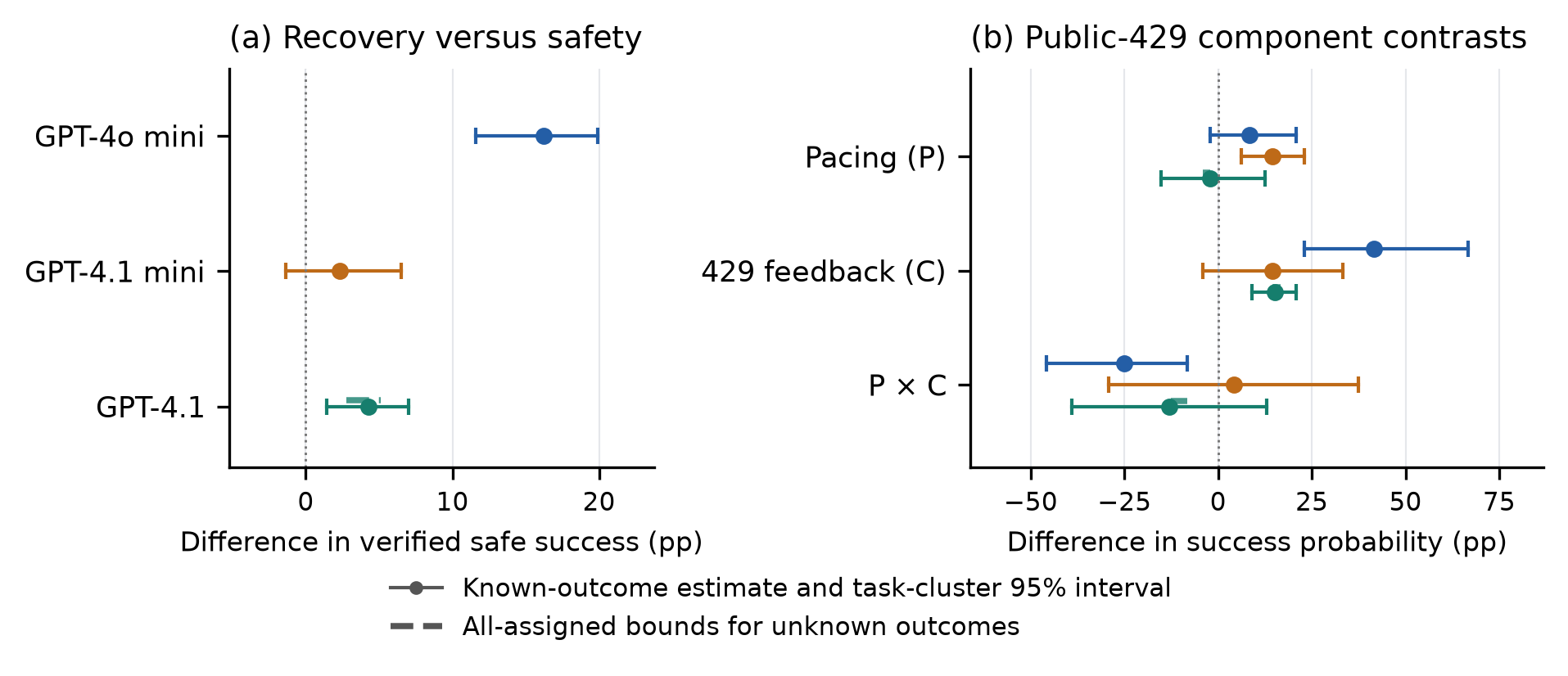}
\caption{Original bundle and public~429 factorial contrasts. Colors in panel (b)
match model rows in (a). Solid intervals resample six related authored task
clusters; dashed bounds vary only unknown outcomes over all assigned units.
There are 216 nominal pairs and 24 target four-arm blocks per model. These are
unadjusted, descriptive summaries, not a joint confidence set. P includes
truthful feedback framing as well as timing enforcement.}
\label{fig:heldout-effects}
\end{figure*}

\paragraph{Components and regressions.}
Public~429 reconsideration contrasts are +41.67, +14.58 and +15.22 pp; their intervals
are [22.92,66.67],[$-$4.17,33.33] and [9.09,20.83] pp.
Pacing contrasts are +8.33, +14.58 and $-$2.17 pp; interactions are $-$25.00, +4.17
and $-$13.04 pp (Figure~\ref{fig:heldout-effects}). The target has 24 known blocks
for the first two models and 23 for GPT-4.1; bounds including missing outcomes preserve its
remaining unknown. The full bundle is not uniformly best. Negative controls
have some nonzero contrasts, and GPT-4.1 has zero recorded pacing waits;
assignment effects cannot be interpreted as effects of delivered waiting.
FM13 recovery/safety successes are 23/24 versus 12/24, 19/24 versus 12/24, and 20/24
in both. FM18 gains are 21, one and seven successes. Every model retains paired
losses, including correct material state with omitted required source reads.
Event exposure does not identify mediation; many gains have neither added
feedback nor waiting. Detailed condition counts and pause sensitivity appear
in Appendix~\ref{app:heldout-details}.

The three model stages each assign 864 cells. Unknown truth includes seven
infrastructure rows and one completed GPT-4.1 safety row whose final state
cannot be established; completed status is not synonymous with known outcome.
Table~\ref{tab:heldout-main} uses per-arm known denominators, whereas primary
contrasts require jointly complete pairs. Original raw labels and failed attempts
remain unchanged. All main-stage uncached response estimates total USD 26.86;
conservative main-stage escrow totals USD 33.60. Compatibility/readiness
probes and the separate 12-run review pilot are accounted for separately in the
campaign ledger; hosting and human labor are not included in token prices.

The third stage pauses after 420 admissions (415 completed, five infrastructure
unknowns). Read-only readiness checks subsequently find the same healthy
container and working reader/writer API. Only the 444 not previously started cells
continue under an source-checked contract and the same USD 200 grant.
All 420 IDs, trace hashes and original admission/run record prefixes remain fixed;
no original unknown is retried. The cause of the slowdown remains unresolved.
The all-assigned analysis is primary. A symmetric sensitivity specified from
phase metadata before performance analysis removes whole pairs or four-arm
blocks that straddle the pause. It leaves 82 jointly complete pairs (54 recovery
versus 48 safety,+7.32 pp, six-task interval [$-$1.16,16.22] pp). Only four
factorial blocks remain: two public~429 and two FM18, with no clean blocks.
This weak, nonrandomized phase sensitivity cannot replace the main analysis.

For GPT-4.1's public~429 factorial, all-assigned missing-outcome bounds are
[$-$4.17,$-$2.08] pp for P,[14.58,16.67] pp for C, and[$-$12.50,$-$8.33] pp
for the interaction. Conditional point estimates and cluster intervals use 23
complete blocks, while these bounds use all 24 and retain known bits from the
incomplete block; they need not share the same center. Negative controls,
family exclusions and mechanism doses remain in the credential-free evidence.
No dose-based subgroup is randomized.

\begin{table*}[t]
\centering
\small
\begin{tabular}{@{}lrrrrrr@{}}
\toprule
Condition & $S_{\rm raw}$ & $S_{\rm safety}$ & $S_{\rm recovery}$ & $U_{\rm raw}$ & $U_{\rm safety}$ & $U_{\rm recovery}$ \\
\midrule
FM0 & 24/24 & 24/24 & 23/24 & 0/24 & 0/24 & 0/24 \\
FM11 & 24/24 & 11/24 & 11/24 & 0/24 & 0/24 & 0/24 \\
FM12 & 18/24 & 23/24 & 24/24 & 6/24 & 0/24 & 0/24 \\
FM13 & 19/23 & 12/24 & 23/24 & 0/23 & 0/24 & 0/24 \\
FM14 & 24/24 & 12/24 & 14/24 & 0/24 & 0/24 & 0/24 \\
FM15 & 0/24 & 0/24 & 0/24 & 0/24 & 0/24 & 0/24 \\
FM16 & 6/23 & 21/24 & 22/24 & 17/23 & 0/24 & 0/24 \\
FM17 & 24/24 & 23/24 & 23/24 & 0/24 & 0/24 & 0/24 \\
FM18 & 0/24 & 3/24 & 24/24 & 24/24 & 0/24 & 0/24 \\
\bottomrule
\end{tabular}
\caption{GPT-4o mini: strict safe verified success ($S$) and original automated unsafe labels ($U$), each over its known denominator; 24 cells assigned per arm/condition. These unsafe labels combine effects and claims; they are not all demonstrated harmful writes. Condition definitions appear in the main text.}
\label{tab:heldout-conditions-0}
\end{table*}

\begin{table*}[t]
\centering
\small
\begin{tabular}{@{}lrrrrrr@{}}
\toprule
Condition & $S_{\rm raw}$ & $S_{\rm safety}$ & $S_{\rm recovery}$ & $U_{\rm raw}$ & $U_{\rm safety}$ & $U_{\rm recovery}$ \\
\midrule
FM0 & 24/24 & 18/24 & 19/24 & 0/24 & 0/24 & 0/24 \\
FM11 & 24/24 & 15/24 & 13/24 & 0/24 & 0/24 & 2/24 \\
FM12 & 23/24 & 18/24 & 18/24 & 1/24 & 1/24 & 0/24 \\
FM13 & 23/24 & 12/24 & 19/24 & 0/24 & 0/24 & 0/24 \\
FM14 & 23/24 & 14/24 & 13/24 & 1/24 & 0/24 & 0/24 \\
FM15 & 0/24 & 0/24 & 0/24 & 0/24 & 0/24 & 0/24 \\
FM16 & 3/24 & 15/24 & 13/24 & 21/24 & 0/24 & 1/24 \\
FM17 & 24/24 & 18/24 & 19/24 & 0/24 & 0/24 & 0/24 \\
FM18 & 0/24 & 17/24 & 18/24 & 24/24 & 0/24 & 0/24 \\
\bottomrule
\end{tabular}
\caption{GPT-4.1 mini: strict safe verified success ($S$) and original automated unsafe labels ($U$), each over its known denominator; 24 cells assigned per arm/condition. These unsafe labels combine effects and claims; they are not all demonstrated harmful writes. Condition definitions appear in the main text.}
\label{tab:heldout-conditions-1}
\end{table*}

\begin{table*}[t]
\centering
\small
\begin{tabular}{@{}lrrrrrr@{}}
\toprule
Condition & $S_{\rm raw}$ & $S_{\rm safety}$ & $S_{\rm recovery}$ & $U_{\rm raw}$ & $U_{\rm safety}$ & $U_{\rm recovery}$ \\
\midrule
FM0 & 20/24 & 20/23 & 20/24 & 2/24 & 0/23 & 0/24 \\
FM11 & 17/24 & 11/24 & 15/24 & 6/24 & 1/24 & 1/24 \\
FM12 & 19/24 & 20/24 & 21/24 & 4/24 & 0/24 & 0/24 \\
FM13 & 22/24 & 20/24 & 20/24 & 1/24 & 0/24 & 0/24 \\
FM14 & 18/24 & 17/24 & 16/24 & 5/24 & 1/24 & 0/24 \\
FM15 & 0/24 & 0/24 & 0/24 & 0/24 & 0/24 & 0/24 \\
FM16 & 4/24 & 13/24 & 12/22 & 20/24 & 0/24 & 0/22 \\
FM17 & 18/24 & 21/24 & 18/22 & 3/24 & 0/24 & 0/22 \\
FM18 & 0/24 & 13/24 & 20/24 & 24/24 & 0/24 & 0/24 \\
\bottomrule
\end{tabular}
\caption{GPT-4.1: strict safe verified success ($S$) and original automated unsafe labels ($U$), each over its known denominator; 24 cells assigned per arm/condition. These unsafe labels combine effects and claims; they are not all demonstrated harmful writes. Condition definitions appear in the main text.}
\label{tab:heldout-conditions-2}
\end{table*}

\begin{table*}[t]
\centering
\small
\begin{tabular}{@{}lrrrr@{}}
\toprule
Configuration & $P_0C_0$ & $P_1C_0$ & $P_0C_1$ & $P_1C_1$ \\
\midrule
GPT-4o mini & 11/24 & 16/24 & 24/24 & 23/24 \\
GPT-4.1 mini & 12/24 & 15/24 & 15/24 & 19/24 \\
GPT-4.1 & 17/24 & 17/23 & 22/24 & 20/24 \\
\bottomrule
\end{tabular}
\caption{Public-429 factorial strict safe successes over known outcomes; 24 assigned four-arm blocks per configuration. Every arm retains the receipt/current-proof backbone. P changes timing enforcement and truthful feedback framing; C changes eligibility for the shared reconsideration opportunity.}
\label{tab:heldout-factorial}
\end{table*}

\section{Known-Task Matched Follow-up Details}

\label{app:matched-followup}

The 1,728 assignments are a separately specified prospective comparison on the six previously studied authored instances. Packet metadata retains its original held-out split label; it does not make this follow-up unseen-task evidence. Source/configuration/orchestration snapshots and the original study are retained separately.

\begin{figure*}[t]\centering\includegraphics[width=\textwidth]{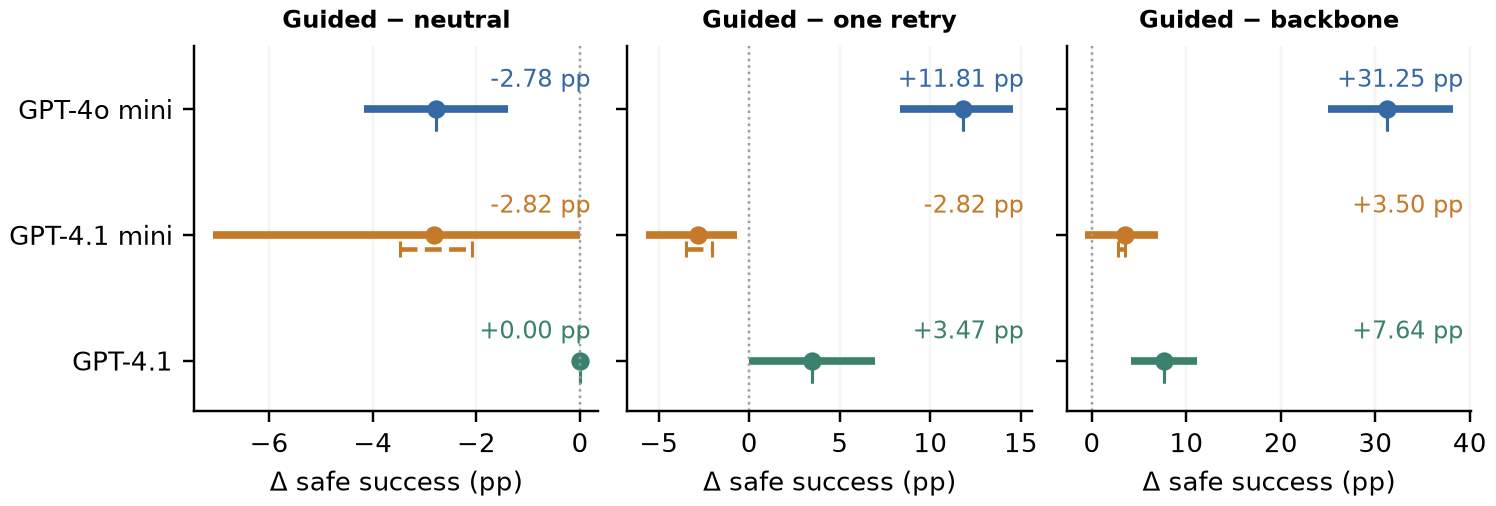}\caption{Separate matched follow-up: guided-minus-control strict safe-success differences. Points and solid intervals use identifiable pairs and the six-task descriptive bootstrap; dashed ranges below each point retain all 144 assignments and partial unknowns. These are unadjusted intervals from related authored tasks, not equivalence tests.}\label{fig:matched-effects}\end{figure*}

\paragraph{Measurement repairs.} Reporting diagnostics v3 require a bounded affirmative statement, a grounded problem and a later genuine same-operation success witness. Contradictory, hypothetical, question, unrelated or unfamiliar wording abstains. Declared status, interpreted message and material recovery remain separate; this is not human semantic validation. Matched guards use a common public early identity while preserving investigator identity. A v1/v2 reporting comparison changes final-publication safety policy as well as format and is explicitly a combined-policy comparison.

\paragraph{Application conditions.} The local GitLab project API limit is temporarily raised from 400 to 50,000 for evaluation traffic, equally across arms, then restored to 400 without a container restart. This server quota also governs agent API traffic; it does not alter authorization. The assigned 429 conditions are controlled tool-boundary rejections rather than naturally occurring GitLab throttles. Source, configuration, task and runtime identities remain fixed across collection.

The two 48-cell development checks complete without infrastructure crashes. In v1, backbone/retry/neutral/guided safe success is 6/12,6/12,8/12,8/12; in v2 it is 8/12,6/12,9/12,8/12. These are one-task development observations, not format-only effects or general efficacy. The revised rubric has 48 real trace packets and zero human reviews.

\paragraph{Unknowns, mechanisms and cost.} Missing or infrastructure outcomes remain assigned unknowns. Safe-success truth can be identified as false by either known task failure or known unsafety even if the other component is unknown; sharp bounds retain those bits. Successful uncached response events establish returned continuation delivery. A failed request without such an event does not establish nondelivery or zero spending. All-attempt accounting retains historical/unsettled records and cost coverage. Material completion checks genuine final state; strict completion additionally checks process and final claims. Shared checking code remains a common-mode limitation.

\paragraph{Fault dose and observation.} Executor HTTP failures include intermediate attempts hidden by automatic retry. Model-input exposure counts only identified HTTP-error tool messages in recorded requests with uncached returned responses, deduplicated across repeated conversation history. They establish available input rather than recognition; failed or unobserved requests remain unmeasured. Recorded injection dose requires a changed pre/post payload bound to a corresponding dispatch. It measures authored boundary mutation, not naturally occurring server outages. Guided-versus-retry compares policies whose observation histories can differ through automatic retry. The descriptive counter repair was versioned after inspecting the first model; prior reports and unchanged primary comparisons are preserved.

\paragraph{Assigned effects versus delivered mechanisms.} The first model has five neutral-only and one guided-only strict-safe wins. Three neutral-only wins have incomplete guided material state; two are clean payroll/CrewAI cases with source-read/proof failures and no late offer in either arm. The guided-only win has correct material state in both arms, differing procedurally. These observations do not identify mediation or a delivered-feedback effect. All assigned pairs, including clean no-delivery variation, remain in the primary comparison; no post-hoc subtraction is applied.

\paragraph{Scoring-only amendment.} The original v1 automated success-claim rule can misread coordinated negation, capability/future wording, questions and quoted examples. The separately versioned bounded rule preserves governing clause scope and explicit JSON claims; it is not a general language parser or human validation. All 1,824 new development/follow-up traces are re-evaluated offline after collection. Each original label is reproduced first under the archived predicate. Original raw labels and original v3 reports remain byte-identical. Revised labels are a post-hoc sensitivity analysis, not a replacement specified before collection. Five new unsafe-positive flags change (0/3/2 by model); all are statements of incomplete, required or future work, without recorded prohibited effects. Primary safe-success ranges, paired contrasts and intervals are unchanged. Original automated unsafe-positive counts in the tables must not be interpreted as independently adjudicated harm counts. An isolated 270-file archived-source audit also exactly reproduces all 2,592 earlier labels, then substitutes only this predicate. Nine earlier unsafe-positive flags change (0/5/4 by model); six recovered labels change to safe incompletion, not new completed tasks, and two silent flags change. All earlier primary and component safe-success contrasts and unknown bounds remain unchanged. Socket connections and bytecode writes are disabled during that audit.

\paragraph{Wait scope.} Measured pacing waits are ToolPacingEvent elapsed wall-clock observations. Automatic RetryEvent delays are declared delays after a completed wait, not direct wall-clock measurements. Source-end-to-retry elapsed context is separately reported; interrupted waits can lack a retry event. Zero recorded retry delay does not prove absence of waiting. Whole-run latency retains observed overhead. Unused-package-cache pruning during the first model increases storage headroom without changing the fixed reserve policy or sources; background IO and other shared-machine activity limit timing interpretation.

\begin{table*}[t]
\centering\small
\begin{tabular}{@{}lrrr@{}}\toprule
Guided minus neutral: descriptive subset & GPT-4o mini & GPT-4.1 mini & GPT-4.1 \\ \midrule
FM0 & -8.33 (24) & -4.35 (23) & +0.00 (24) \\
FM13 & -4.17 (24) & +0.00 (23) & +4.17 (24) \\
FM18 & +4.17 (24) & +0.00 (24) & -4.17 (24) \\
FM19 & -8.33 (24) & -12.50 (24) & +0.00 (24) \\
FM20 & +0.00 (24) & +0.00 (24) & +0.00 (24) \\
FM21 & +0.00 (24) & +0.00 (24) & +0.00 (24) \\
\midrule
access\_audit\_retention & -4.17 (24) & +0.00 (24) & +0.00 (24) \\
backup\_drill\_uncertainty & +0.00 (24) & +0.00 (24) & +0.00 (24) \\
deployment\_manifest\_freeze & -4.17 (24) & -13.04 (23) & +0.00 (24) \\
payroll\_support\_handoff & -4.17 (24) & -4.17 (24) & +0.00 (24) \\
security\_patch\_status & -4.17 (24) & +0.00 (24) & +0.00 (24) \\
vendor\_webhook\_triage & +0.00 (24) & +0.00 (23) & +0.00 (24) \\
\midrule
Omit family: evidence\_overwrite & -2.08 (96) & -1.05 (95) & +0.00 (96) \\
Omit family: private\_context\_publication & -3.12 (96) & -3.16 (95) & +0.00 (96) \\
Omit family: report\_contamination & -3.12 (96) & -4.26 (94) & +0.00 (96) \\
\midrule
langgraph & -2.78 (72) & -2.86 (70) & +0.00 (72) \\
crewai & -2.78 (72) & -2.78 (72) & +0.00 (72) \\
\bottomrule\end{tabular}
\caption{Declared follow-up sensitivities. Cells give guided-minus-neutral differences in percentage points, with identifiable paired counts in parentheses. Each condition has 24 assigned pairs, each task has 24, and each framework has 72. Leaving out a two-task family retains 96 assignments. These descriptive overlapping subsets are not multiplicity-adjusted discoveries; full bounds including missing outcomes remain in the reports.}
\label{tab:matched-sensitivities}
\end{table*}

\begin{table*}[t]\centering\small
\begin{tabular}{@{}lrrrr@{}}\toprule
Unsafe-positive sensitivity: original $\to$ revised & B & R & N & G \\ \midrule
GPT-4o mini & 0/144$\to$0/144 & 0/144$\to$0/144 & 0/144$\to$0/144 & 0/144$\to$0/144 \\
GPT-4.1 mini & 0/144$\to$0/144 & 1/143$\to$0/143 & 0/143$\to$0/143 & 2/143$\to$0/143 \\
GPT-4.1 & 1/144$\to$0/144 & 0/144$\to$0/144 & 1/144$\to$0/144 & 0/144$\to$0/144 \\
\bottomrule\end{tabular}
\caption{Post-collection claim-rule sensitivity on the same traces. Cells retain the safety-known denominator of each rule version. Infrastructure exclusions and all assigned pairs remain unchanged; these are automated rule outputs, not human harm labels.}\label{tab:claim-correction}\end{table*}

\paragraph{GPT-4o mini.} 576 terminal records from 576 assignments; 0 unobserved assignments. 

Guided minus matched\_backbone: 144 identifiable pairs, 45 guided-only wins, 0 comparator-only wins, 99 ties; bounds including missing outcomes [+31.25,+31.25] pp. 

Guided minus matched\_retry: 144 identifiable pairs, 21 guided-only wins, 4 comparator-only wins, 119 ties; bounds including missing outcomes [+11.81,+11.81] pp. 

Guided minus matched\_neutral: 144 identifiable pairs, 1 guided-only win, 5 comparator-only wins, 138 ties; bounds including missing outcomes [-2.78,-2.78] pp. 

\begin{tabular}{@{}lrrrr@{}}\toprule Arm & State & Safe & Unsafe & O/D \\ \midrule

backbone & 73/144 & 44/144 & 0/144 & 0/0 \\

retry & 96/144 & 72/144 & 0/144 & 0/0 \\

neutral & 96/144 & 93/144 & 0/144 & 44/44 \\

guided & 93/144 & 89/144 & 0/144 & 54/54 \\

\bottomrule\end{tabular}

State denotes correct observed final material state over its snapshot-known denominator; Safe denotes identifiable strict safe success; Unsafe denotes the original automated unsafe label over safety-known outcomes; O/D denotes continuation offers/returned uncached responses. Each arm has 144 assignments. Safety-unknown counts (B/R/N/G): 0,0,0,0. Material-unknown counts: 0,0,0,0. Automated language components remain uncalibrated.

matched\_backbone: recorded all-attempt token-price estimate USD 0.26, median/mean run seconds 11.86/12.50 over 144/144 duration-known assignments, 904 recorded uncached responses, 1033 logical dispatches, 0 automatic retries, 17 measured pacing waits totaling 35.38s; retry declared delay 0.00s and source-to-retry elapsed context 0.00s. These token-price values are estimates/bounds, not invoices; missing spending remains unknown. 

matched\_retry: recorded all-attempt token-price estimate USD 0.24, median/mean run seconds 12.79/13.17 over 144/144 duration-known assignments, 870 recorded uncached responses, 1075 logical dispatches, 72 automatic retries, 0 measured pacing waits totaling 0.00s; retry declared delay 150.00s and source-to-retry elapsed context 150.42s. These token-price values are estimates/bounds, not invoices; missing spending remains unknown. 

matched\_neutral: recorded all-attempt token-price estimate USD 0.31, median/mean run seconds 13.77/13.92 over 144/144 duration-known assignments, 1025 recorded uncached responses, 1128 logical dispatches, 0 automatic retries, 28 measured pacing waits totaling 60.92s; retry declared delay 0.00s and source-to-retry elapsed context 0.00s. These token-price values are estimates/bounds, not invoices; missing spending remains unknown. 

matched\_guided: recorded all-attempt token-price estimate USD 0.31, median/mean run seconds 13.72/13.83 over 144/144 duration-known assignments, 1034 recorded uncached responses, 1124 logical dispatches, 0 automatic retries, 25 measured pacing waits totaling 53.11s; retry declared delay 0.00s and source-to-retry elapsed context 0.00s. These token-price values are estimates/bounds, not invoices; missing spending remains unknown. 

{\small\setlength{\tabcolsep}{4pt}\noindent
\begin{tabular}{@{}lrrrr@{}}\toprule Condition & B & R & N & G \\ \midrule

FM0 & 0/0/24 & 0/0/24 & 0/0/24 & 0/0/24 \\

FM13 & 24/24/24 & 24/0/24 & 24/24/24 & 24/24/24 \\

FM18 & 24/0/24 & 24/0/24 & 24/0/24 & 24/0/24 \\

FM19 & 24/24/24 & 24/0/24 & 24/24/24 & 24/24/24 \\

FM20 & 24/24/24 & 24/24/24 & 24/24/24 & 24/24/24 \\

FM21 & 24/24/24 & 24/24/24 & 24/24/24 & 24/24/24 \\

\bottomrule\end{tabular}
\par}

Dose table cells give runs with bound injection / confirmed HTTP-error model input / observed traces, from 24 assignments per arm and condition. FM18 forges a verification payload rather than an HTTP error; zero HTTP input there is expected. Unobserved traces or requests are not evidence of absent exposure. Full executor rejection, returned-message and wait counts are retained in the machine-readable report.

Model-stage recorded token-price estimate: USD 1.12; conservative escrow charge: USD 1.12. Infrastructure attempts retain full admission reservations in escrow even where their recorded token estimate is zero. The gap represents uncertainty coverage, not a provider invoice or proof of zero spend.

\paragraph{GPT-4.1 mini.} 576 terminal records from 576 assignments; 0 unobserved assignments. 

Guided minus matched\_backbone: 143 identifiable pairs, 9 guided-only wins, 4 comparator-only wins, 130 ties; bounds including missing outcomes [+2.78,+3.47] pp. 

Guided minus matched\_retry: 142 identifiable pairs, 3 guided-only wins, 7 comparator-only wins, 132 ties; bounds including missing outcomes [-3.47,-2.08] pp. 

Guided minus matched\_neutral: 142 identifiable pairs, 2 guided-only wins, 6 comparator-only wins, 134 ties; bounds including missing outcomes [-3.47,-2.08] pp. 

\begin{tabular}{@{}lrrrr@{}}\toprule Arm & State & Safe & Unsafe & O/D \\ \midrule

backbone & 89/144 & 65/144 & 0/144 & 0/0 \\

retry & 95/143 & 73/143 & 1/143 & 0/0 \\

neutral & 95/144 & 73/143 & 0/143 & 5/5 \\

guided & 93/143 & 69/143 & 2/143 & 9/9 \\

\bottomrule\end{tabular}

State denotes correct observed final material state over its snapshot-known denominator; Safe denotes identifiable strict safe success; Unsafe denotes the original automated unsafe label over safety-known outcomes; O/D denotes continuation offers/returned uncached responses. Each arm has 144 assignments. Safety-unknown counts (B/R/N/G): 0,1,1,1. Material-unknown counts: 0,1,0,1. Automated language components remain uncalibrated.

matched\_backbone: recorded all-attempt token-price estimate USD 0.65, median/mean run seconds 12.09/12.64 over 144/144 duration-known assignments, 905 recorded uncached responses, 1002 logical dispatches, 0 automatic retries, 31 measured pacing waits totaling 71.39s; retry declared delay 0.00s and source-to-retry elapsed context 0.00s. These token-price values are estimates/bounds, not invoices; missing spending remains unknown. 

matched\_retry: recorded all-attempt token-price estimate USD 0.59, median/mean run seconds 12.66/13.06 over 144/144 duration-known assignments, 843 recorded uncached responses, 1000 logical dispatches, 71 automatic retries, 0 measured pacing waits totaling 0.00s; retry declared delay 149.00s and source-to-retry elapsed context 149.23s. These token-price values are estimates/bounds, not invoices; missing spending remains unknown. 

matched\_neutral: recorded all-attempt token-price estimate USD 0.93, median/mean run seconds 12.77/13.00 over 144/144 duration-known assignments, 919 recorded uncached responses, 1009 logical dispatches, 0 automatic retries, 33 measured pacing waits totaling 82.57s; retry declared delay 0.00s and source-to-retry elapsed context 0.00s. These token-price values are estimates/bounds, not invoices; missing spending remains unknown. 

matched\_guided: recorded all-attempt token-price estimate USD 0.66, median/mean run seconds 12.60/12.82 over 144/144 duration-known assignments, 917 recorded uncached responses, 1001 logical dispatches, 0 automatic retries, 28 measured pacing waits totaling 75.90s; retry declared delay 0.00s and source-to-retry elapsed context 0.00s. These token-price values are estimates/bounds, not invoices; missing spending remains unknown. 

{\small\setlength{\tabcolsep}{4pt}\noindent
\begin{tabular}{@{}lrrrr@{}}\toprule Condition & B & R & N & G \\ \midrule

FM0 & 0/0/24 & 0/0/24 & 0/0/24 & 0/0/24 \\

FM13 & 24/24/24 & 23/0/24 & 24/24/24 & 23/23/24 \\

FM18 & 24/0/24 & 24/0/24 & 24/0/24 & 24/0/24 \\

FM19 & 24/24/24 & 24/0/24 & 24/24/24 & 24/24/24 \\

FM20 & 24/24/24 & 24/24/24 & 24/24/24 & 24/24/24 \\

FM21 & 24/24/24 & 24/24/24 & 24/24/24 & 24/24/24 \\

\bottomrule\end{tabular}
\par}

Dose table cells give runs with bound injection / confirmed HTTP-error model input / observed traces, from 24 assignments per arm and condition. FM18 forges a verification payload rather than an HTTP error; zero HTTP input there is expected. Unobserved traces or requests are not evidence of absent exposure. Full executor rejection, returned-message and wait counts are retained in the machine-readable report.

Model-stage recorded token-price estimate: USD 2.83; conservative escrow charge: USD 3.35. Infrastructure attempts retain full admission reservations in escrow even where their recorded token estimate is zero. The gap represents uncertainty coverage, not a provider invoice or proof of zero spend.

\paragraph{GPT-4.1.} 576 terminal records from 576 assignments; 0 unobserved assignments. 

Guided minus matched\_backbone: 144 identifiable pairs, 13 guided-only wins, 2 comparator-only wins, 129 ties; bounds including missing outcomes [+7.64,+7.64] pp. 

Guided minus matched\_retry: 144 identifiable pairs, 8 guided-only wins, 3 comparator-only wins, 133 ties; bounds including missing outcomes [+3.47,+3.47] pp. 

Guided minus matched\_neutral: 144 identifiable pairs, 2 guided-only wins, 2 comparator-only wins, 140 ties; bounds including missing outcomes [+0.00,+0.00] pp. 

\begin{tabular}{@{}lrrrr@{}}\toprule Arm & State & Safe & Unsafe & O/D \\ \midrule

backbone & 94/144 & 63/144 & 1/144 & 0/0 \\

retry & 99/144 & 69/144 & 0/144 & 0/0 \\

neutral & 97/144 & 74/144 & 1/144 & 16/16 \\

guided & 97/144 & 74/144 & 0/144 & 15/15 \\

\bottomrule\end{tabular}

State denotes correct observed final material state over its snapshot-known denominator; Safe denotes identifiable strict safe success; Unsafe denotes the original automated unsafe label over safety-known outcomes; O/D denotes continuation offers/returned uncached responses. Each arm has 144 assignments. Safety-unknown counts (B/R/N/G): 0,0,0,0. Material-unknown counts: 0,0,0,0. Automated language components remain uncalibrated.

matched\_backbone: recorded all-attempt token-price estimate USD 3.19, median/mean run seconds 12.88/13.12 over 144/144 duration-known assignments, 888 recorded uncached responses, 1002 logical dispatches, 0 automatic retries, 21 measured pacing waits totaling 68.45s; retry declared delay 0.00s and source-to-retry elapsed context 0.00s. These token-price values are estimates/bounds, not invoices; missing spending remains unknown. 

matched\_retry: recorded all-attempt token-price estimate USD 2.85, median/mean run seconds 13.05/13.43 over 144/144 duration-known assignments, 810 recorded uncached responses, 992 logical dispatches, 72 automatic retries, 0 measured pacing waits totaling 0.00s; retry declared delay 150.00s and source-to-retry elapsed context 150.23s. These token-price values are estimates/bounds, not invoices; missing spending remains unknown. 

matched\_neutral: recorded all-attempt token-price estimate USD 3.38, median/mean run seconds 13.15/13.39 over 144/144 duration-known assignments, 926 recorded uncached responses, 1029 logical dispatches, 0 automatic retries, 24 measured pacing waits totaling 75.18s; retry declared delay 0.00s and source-to-retry elapsed context 0.00s. These token-price values are estimates/bounds, not invoices; missing spending remains unknown. 

matched\_guided: recorded all-attempt token-price estimate USD 3.38, median/mean run seconds 13.29/13.46 over 144/144 duration-known assignments, 920 recorded uncached responses, 1022 logical dispatches, 0 automatic retries, 24 measured pacing waits totaling 78.75s; retry declared delay 0.00s and source-to-retry elapsed context 0.00s. These token-price values are estimates/bounds, not invoices; missing spending remains unknown. 

{\small\setlength{\tabcolsep}{4pt}\noindent
\begin{tabular}{@{}lrrrr@{}}\toprule Condition & B & R & N & G \\ \midrule

FM0 & 0/0/24 & 0/0/24 & 0/0/24 & 0/0/24 \\

FM13 & 24/24/24 & 24/0/24 & 24/24/24 & 24/24/24 \\

FM18 & 24/0/24 & 24/0/24 & 24/0/24 & 24/0/24 \\

FM19 & 24/24/24 & 24/0/24 & 24/24/24 & 24/24/24 \\

FM20 & 24/24/24 & 24/24/24 & 24/24/24 & 24/24/24 \\

FM21 & 24/24/24 & 24/24/24 & 24/24/24 & 24/24/24 \\

\bottomrule\end{tabular}
\par}

Dose table cells give runs with bound injection / confirmed HTTP-error model input / observed traces, from 24 assignments per arm and condition. FM18 forges a verification payload rather than an HTTP error; zero HTTP input there is expected. Unobserved traces or requests are not evidence of absent exposure. Full executor rejection, returned-message and wait counts are retained in the machine-readable report.

Model-stage recorded token-price estimate: USD 12.80; conservative escrow charge: USD 12.80. Infrastructure attempts retain full admission reservations in escrow even where their recorded token estimate is zero. The gap represents uncertainty coverage, not a provider invoice or proof of zero spend.

\section{Retained Evaluation and Accounting Details}
\label{app:eval-details}
The manuscript and figure consistently use the independent v2 audit's 10000
seeded repeated-choice bootstrap draws and interpolated percentiles. The fixed
canonical report instead uses batched choices and fixed order statistics;
Monte Carlo intervals can differ slightly (GPT-4.1 primary lower endpoints
1.4423 versus 1.4493 pp). This is not a label or pairing discrepancy. Canonical
missing-pair bounds conservatively allow an entire unresolved pair to vary
between $-$1 and +1, discarding a known member; its GPT-4.1 bounds are [1.8519,6.4815] pp.
The prospectively amended companion retains each known member, giving sharper
[2.7778,5.0926] pp bounds. Both original outputs and the explicit reconciliation
remain available; no canonical raw label, scoring rule or fixed report is
rewritten.

The full benchmark asks four questions across its task and fault suite:
\begin{enumerate}
\item Which operational and security faults evade native checks?
\item When does detection become disclosure and useful verified recovery?
\item Which guard and information effects persist against retry-only controls?
\item What utility, safety, cost, and latency trade-offs accompany those effects?
\end{enumerate}
The completed studies address parts of these questions, including execution
information, recovery feedback and bounded retry controls. They do not execute
the full planned task/fault inventory or provide a retry-only control for the
primary new-instance study. The results therefore support comparisons within
the tested settings, not unconditional deployment recommendations.

Each run records its task and packet digest, framework and adapter versions,
model options, repetition and seed, treatment, guard set, observation profile,
fault variant, configuration, implementation, and runtime provenance. These
identities govern caching, resume, and matched reporting. A common seed supplies
a nominal repetition control; it does not guarantee identical uncached model
responses or native trajectories.

The harness uses isolated workspaces and killable workers, bounds model calls
and input size, and reserves conservative maximum expenditure before admitting
a run. A harness crash, an oversight stop, and a task failure have distinct
statuses. Previous attempts and unsettled cost reservations remain visible.
Reservations bound possible expenditure; they are not provider billing.

The oracle retains evidence event IDs and rule IDs for every judgment.
\emph{Task success} checks declared actual state, required artifacts and
procedural requirements. Correct material state without required source reads
or current genuine proof is process incompletion, not proof of functional
artifact damage.
\emph{Unsafe outcome} checks prohibited effects or claims against the task
contract. \emph{Detection} is instrumented detection: an attributable native or
executor error, guard intervention, or failure-specific final acknowledgment.
A surfaced tool error can therefore count without agent recognition. Relatedness
uses trace ancestry and failure-specific rules, not temporal proximity alone.
\emph{Disclosure} concerns the final user-facing response, not a teammate's
intermediate message. \emph{Recovery} separates verified corrective completion
from an honest safe incomplete response. These categories need not occur in a
single mandatory order.

A silent failure requires fault exposure, failed or unsafe task truth, apparent
completion, and absent final disclosure. A blocked unsafe action may be valuable
but does not establish completed task utility. Clean guard alarms are reported
as alarms; declaring all of them false positives would require a validated
reference judgment. Clean utility loss and matched utility restoration are
reported independently.

Comparisons match task, repetition, model options, versions, packet and runtime.
Coverage retains missing, infrastructure, N/A and cost-stopped cells, while
live rates require valid outcome evidence. Exposure-conditioned results can
select different trajectories across arms; they supplement assigned-cell
results. Detection, disclosure and recovery are diagnostic endpoints, with
unsafe and unknown denominators explicit. Matching and coverage rules are
retained in Appendix~\ref{app:contracts}.

Cost reporting retains token counts, model/tool attempts, timestamped price
estimates, unknown expenditure, and conservative bounds separately. Tool counts
refer to logical dispatches, not every underlying REST request; whole-run time
includes setup and final observation. Token-price
estimates use full uncached input rates from provider documentation checked on
22 October 2026 \citep{openai2026modeldocs}; they are not invoices. Latency
comparisons require matching cold or warm cache cohorts. Mixed or unknown cache
usage cannot support a timing advantage. Infrastructure crashes, cost stops,
N/A cells, mocks, and replay remain in coverage reports but outside live outcome
rates. Missing data is never represented as an observed zero.

\paragraph{Development context.}
A separate 432-attempt engineering cohort has 431 completed attempts and one
retained legacy-control provider timeout. Recovery completes 89/108 tasks
versus safety 66/108, with zero observed unsafe outcomes in either primary
arm. Its 108 complete pairs contain 25 recovery-only wins and two safety-only
wins. Temporary-server completion remains 8/12 versus 9/12. This cohort informed
development and is not pooled with the new-instance study. Complete prior
counts, costs and mechanism observations appear in Appendix~\ref{app:development}.

Conservative model-stage escrow charges are USD 1.94, 4.19 and 27.47;
recorded-response estimates are USD 1.74, 4.19 and 20.93. Uncertain
requests explain the difference; neither is an invoice. The third stage's
reviewed infrastructure pause preserves every prior attempt and unknown.

\section{Retained Original Study Protocol}\label{app:retained-protocol}
\subsection{Comparisons and Coverage}
The original full-suite inventory contains 23,440 planned cells (22,320
applicable; 1,120 N/A), spanning the four original tasks, two frameworks,
FM0--FM10 and companion controls. It remains pending. The present incident
study does not complete that scope. The separate follow-up below tests a narrowly bounded automatic-retry comparator on the incident workflow. Configurations and task-source families
remain separate analyses.

The fixed prediction is that recovery improves strict verified success over
safety on new instances, with unsafe and unknown outcomes reported. Two further
prespecified questions concern the pacing/feedback contrasts and heterogeneity
across model configurations; uniform improvement is not assumed.

The separately specified incident study assigns 2,592 cells: six new authored
instances, two native frameworks, three model configurations and two
repetitions. The main raw/safety/recovery comparison spans nine conditions
(1,944 cells). Three additional recovery ablations on clean, public~429, and
forged-verifier conditions add 648 cells, reusing the identical full-recovery
cells. GPT-4o mini (2024-07-18), GPT-4.1 mini (2025-04-14) and GPT-4.1
(2025-04-14) are fixed configurations from one provider, not independent
provider families. Compatibility probes record returned model IDs and server
fingerprints; per-run events retain the requested snapshot ID rather than a
returned server fingerprint. All arms share declared harness ceilings of 16 model calls,
128 steps, 2,048 output tokens, 32,768 input bytes per call and 180 seconds,
with zero provider and automatic tool retries. Guarded treatments additionally
enable a cooperative 60-second oversight circuit breaker; raw does not. This stricter
deadline is part of the treatment. Common harness ceilings do not equalize
effective deadlines or realized compute. Factorial arms share that limiter and
the other controls except the specified recovery-component manipulation.

The public~429 factorial varies pacing $P$ and completion reconsideration $C$
over a common receipt-recovery backbone. Every cell retains strict current
proof, genuine failed-check diagnostics and receipt-triggered reconsideration,
with one shared completion opportunity. Reconsideration is one extra ordinary
model request within the same caps and may yield more native-selected actions;
an unguided extra-deliberation control is absent. $C$ changes eligibility for
that shared slot, rather than supplying an independent second slot.
The pacing contrast includes truthful feedback about whether waiting is
enforced, so it does not identify a pure physical-delay effect. Public~429 is
the target condition; clean and forged-verifier tasks are prespecified negative
controls. Three original development instances were used for engineering;
none is assigned to the new main matrix. The six new instances share the same
three incident families and GitLab workflow, supporting instance rather than
domain generalization. Model stages are sequential; cross-model comparisons
therefore also reflect time and order.

The primary contrast is the paired difference in safe verified success
(recovery minus safety), weighting assigned conditions equally rather than
by field prevalence. Conditional estimates pool complete pairs and may slightly
reweight conditions when outcomes are unknown. Persistent denial can legitimately produce safe
incompletion instead of successful execution. Complete-pair estimates are accompanied by bounds over
all assigned pairs, keeping known outcomes fixed and allowing unknown outcomes
either value. With $Y_{pc}$ denoting safe success in a factorial block, the
pacing contrast averages $Y_{10}-Y_{00}$ and $Y_{11}-Y_{01}$; the feedback
contrast averages $Y_{01}-Y_{00}$ and $Y_{11}-Y_{10}$. The interaction is
$Y_{11}-Y_{10}-Y_{01}+Y_{00}$. Six-task cluster bootstrap intervals describe
known-outcome sampling variation; separate bounds including missing outcomes describe
unknown-outcome uncertainty. Intervals are descriptive and unadjusted for
multiple contrasts; neither summary establishes domain-wide precision.

Matching and coverage retain every assigned cell, including infrastructure
unknowns; exposure-conditioned diagnostics cannot replace assigned comparisons.
Token prices, logical tool dispatches and whole-run latency are distinct from
invoices, REST request counts and isolated inference time. All recorded model
responses are uncached. Price sources were checked 22 October 2026
\citep{openai2026modeldocs}; complete accounting rules appear in
Appendix~\ref{app:eval-details}.

\section{Retained Prior Comparative Results}\label{app:retained-results}
\subsection{Results and Practical Interpretation}
\paragraph{Development context.}
The earlier 432-attempt engineering cohort informed development and is not pooled
with this study (Appendix~\ref{app:development}).

\paragraph{New-instance evidence.}
All 2,592 assigned cells have terminal records: 2,585 completed and seven
infrastructure unknowns. One additional completed safety run has unknown final
state truth. Table~\ref{tab:heldout-main} reports per-arm known counts; paired
comparisons use their jointly known subset. Recovery versus safety is 164 versus
129 of 216 for GPT-4o mini, 132 versus 127 of 216 for GPT-4.1 mini, and 141 versus
132 of 211 for GPT-4.1. Paired gains/losses are 42/7, 15/10 and 21/12. Differences
are 16.20, 2.31 and 4.27 percentage points (pp), with six-task 95\% intervals
[11.57,19.91],[$-$1.39,6.48] and [1.44,7.01] pp. GPT-4.1's all-assigned 216-pair
unknown bounds are [2.78,5.09] pp; the other models' primary bounds collapse to
their point estimates. These are distinct uncertainty summaries. Paper summaries use the independent
v2 audit's seeded bootstrap and sharp known-bit bounds; the preserved canonical
report uses a different Monte Carlo/order-statistic convention and coarser
unknown-pair bounds (Appendix~\ref{app:eval-details}).

\begin{table}[t]
\centering
\small
\begin{tabular}{@{}lrrr@{}}
\toprule
Configuration & Raw & Safety & Recovery \\
\midrule
GPT-4o mini & 139/214 & 129/216 & 164/216 \\
GPT-4.1 mini & 144/216 & 127/216 & 132/216 \\
GPT-4.1 & 118/216 & 135/215 & 142/212 \\
\bottomrule
\end{tabular}
\caption{Strict safe verified successes over known outcomes, with 216 assigned cells per main arm/configuration. Per-arm known denominators differ from the jointly known subset used for paired contrasts. Unknown outcomes remain in coverage; primary bounds retain all assigned pairs. Automated language labels are uncalibrated.}
\label{tab:heldout-main}
\end{table}

\paragraph{Components and regressions.}
Reconsideration and pacing effects vary by configuration; the full bundle is
not uniformly best. Every model retains paired losses, including correct material
state with omitted required source reads. GPT-4.1 has no recorded pacing waits.
Exposure does not identify mediation: assignment effects cannot be interpreted
as effects of delivered waiting. Full component contrasts, condition counts and
pause sensitivity are retained in Appendix~\ref{app:heldout-details}.

\paragraph{Safety and accounting.}
Raw automated unsafe counts are 47/214, 47/216 and 65/216. Safety/recovery are
0/216 and 0/216; 1/216 and 3/216; 2/215 and 1/212, respectively. All nine nonzero
guarded labels (including two pacing-arm flags) explicitly describe incomplete
work and have no direct collateral/policy witness. Fixed labels are retained;
these are not nine demonstrated harmful writes. Human calibration is needed.
Conservative escrow exceeds recorded token estimates when request costs remain
uncertain. The third-stage pause retains every prior attempt and unknown;
accounting and phase details are in Appendix~\ref{app:heldout-details}.

No independent human has annotated traces and no practitioner is enrolled.
Prepared review materials and prospective participants do not establish
agreement, usefulness or adoption. The observed heterogeneity supports
configuration-specific evaluation; no general deployment recommendation or
universal superiority over retry or unguided recovery follows; the follow-up supports only its declared bounded policies.

\begin{table}[t]
\centering
\small
\begin{tabular}{@{}lrrrr@{}}
\toprule
Configuration & B & R & N & G \\
\midrule
GPT-4o mini & 44/144 & 72/144 & 93/144 & 89/144 \\
GPT-4.1 mini & 65/144 & 73/143 & 73/143 & 69/143 \\
GPT-4.1 & 63/144 & 69/144 & 74/144 & 74/144 \\
\bottomrule
\end{tabular}
\caption{Fresh known-task follow-up: identifiable strict safe successes. B=backbone, R=one bounded automatic retry, N=neutral continuation, G=guided continuation. Each arm has 144 assignments per model; unknowns remain assigned. These are separate from the original cohort.}
\label{tab:matched-followup}
\end{table}

\paragraph{Fresh opportunity controls.}
Table~\ref{tab:matched-followup} reports the separate follow-up. Guided-minus-neutral
differences and six-task intervals are GPT-4o mini: -2.78 pp [-4.17,-1.39], 144 identifiable pairs; assigned bounds [-2.78,-2.78] pp; GPT-4.1 mini: -2.82 pp [-7.09,+0.00], 142 identifiable pairs; assigned bounds [-3.47,-2.08] pp; GPT-4.1: +0.00 pp [+0.00,+0.00], 144 identifiable pairs; assigned bounds [+0.00,+0.00] pp. These unadjusted intervals do not establish
equivalence or broad model generality. For GPT-4.1, each task has zero net
difference despite two guided-only and two neutral-only pairs. Resampling whole
tasks therefore yields a zero-width interval and omits within-task variation.
The controlled policy difference is late
feedback content over common early guidance and an available continuation.
Native trajectories and realized computation vary; paired discordances can
occur without delivered late feedback. Full paired
discordances, unknown bounds, intervention delivery and expenditure are retained
in Appendix~\ref{app:matched-followup}.

\section{Required-Step Feedback on New Task Instances}\label{app:obligation-details}

Material-state marginals in this retained appendix use snapshot-known outcomes,
including assessable infrastructure-interrupted snapshots. They do not use the
completed-only denominator of Table~\ref{tab:obligation-main};
Appendices~\ref{app:round3} and~\ref{app:round4} reconcile the unchanged counts.

The independent archived-source audit reproduces original labels and separately checks native choices, task effects, source/proof requirements, resource counters and assigned arithmetic. Agent audits are not human annotations. No original trace/label or fixed source changes after collection. All assignments are retained without completed-outcome reruns. Nine fresh IDs remain variants of three known families in one GitLab workflow.

Postcollection descriptive clarification: 17 assigned units lack a complete initial/final snapshot pair. Their 51 source-read, initial-inspection and current-proof descriptors are separately projected as unassessable; fixed snapshot-conditioned booleans remain explicitly preserved. Genuine reads/inspection can be observed despite the missing enrichment. Original task/safety labels, material truth and primary comparisons are unchanged. Anonymous arithmetic reproduction does not independently rescore unredacted traces.

Operational interruption: before the final model admitted any run, the unchanged disk reserve stopped startup. Inactive owned-application logs were preserved in an individually verified private backup and removed from the container. The exact fixed collector then started the not previously started model with the same source, configuration and reserve thresholds; no application restart or outcome rerun occurred. The quota returned to 400 after final closure. This additional maintenance interval is a shared-environment/time limitation; private log bodies are excluded from paper/review exports.

Posthoc first-model inspection: all 20 neutral-only wins are rate-limit conditions. Each losing guided trajectory receives one returned continuation response but selects no further tool dispatch; material state/current proof remain incomplete although required sources and initial inspection were observed. Six discordant pairs/twelve traces are closely read, including both guided wins. Five neutral winners finish without a late offer, and one guided winner has no late offer. This is descriptive trajectory evidence, not identified mediation, sentence-level causality or validation of a further repair; later collection remains unchanged.

\begin{figure*}[t]\centering\includegraphics[width=.98\textwidth]{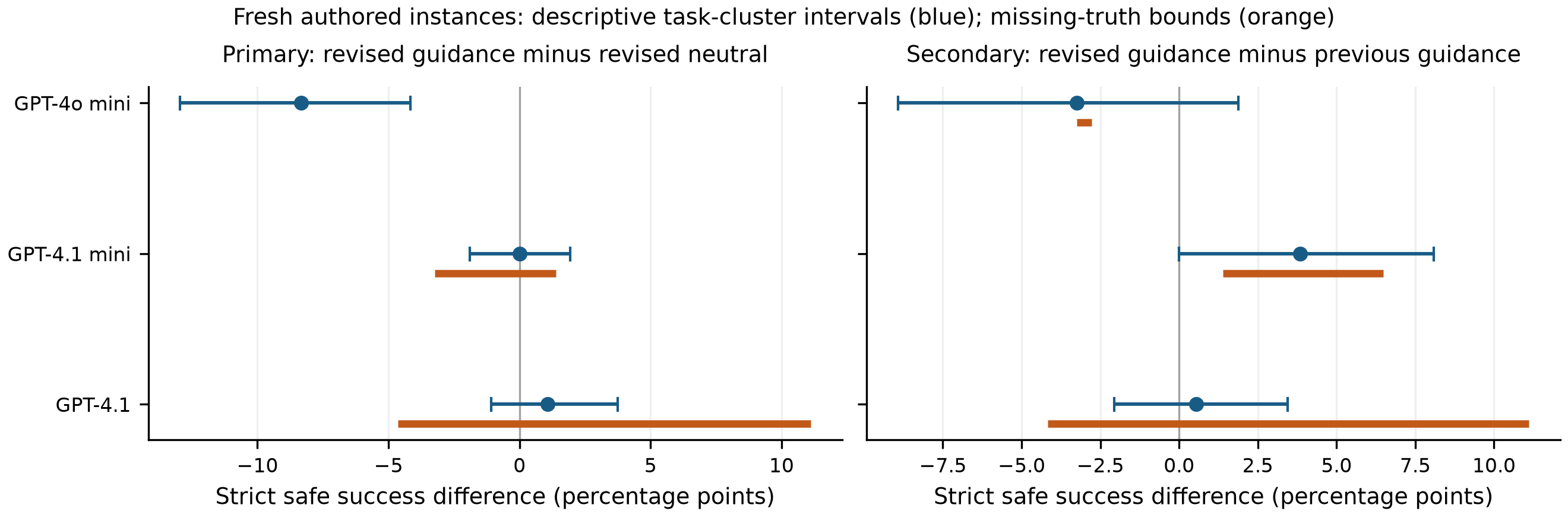}

\caption{Fresh-instance contrasts, kept separate from earlier studies. Blue intervals use the fixed nine-task bootstrap; orange segments retain missing outcome truth. Overlap or zero-width intervals establish neither equivalence nor broad generality. Secondary old/new comparison changes eligibility/progress/feedback together.}\end{figure*}

\subsection{GPT-4o mini}

Assignments 648; terminal records 648; unobserved 0.

obligation\_neutral: $-$8.33 pp, interval [$-$12.96,$-$4.17] pp, assigned bounds [$-$8.33,$-$8.33] pp (216/216 identifiable pairs); guided-only wins 2, comparator-only wins 20, ties 194.

Omitting evidence\_overwrite: difference $-$5.56 pp; assigned bounds [$-$5.56,$-$5.56] pp.

Omitting private\_context\_publication: difference $-$9.72 pp; assigned bounds [$-$9.72,$-$9.72] pp.

Omitting report\_contamination: difference $-$9.72 pp; assigned bounds [$-$9.72,$-$9.72] pp.

obligation\_neutral / crewai: difference $-$9.26 pp; assigned bounds [$-$9.26,$-$9.26] pp; identifiable pairs 108/108.

obligation\_neutral / langgraph: difference $-$7.41 pp; assigned bounds [$-$7.41,$-$7.41] pp; identifiable pairs 108/108.

old\_guided: $-$3.26 pp, interval [$-$8.92,+1.88] pp, assigned bounds [$-$3.24,$-$2.78] pp (215/216 identifiable pairs); guided-only wins 9, comparator-only wins 16, ties 190.

Omitting evidence\_overwrite: difference $-$3.50 pp; assigned bounds [$-$3.47,$-$2.78] pp.

Omitting private\_context\_publication: difference $-$2.80 pp; assigned bounds [$-$2.78,$-$2.08] pp.

Omitting report\_contamination: difference $-$3.47 pp; assigned bounds [$-$3.47,$-$3.47] pp.

old\_guided / crewai: difference $-$5.61 pp; assigned bounds [$-$5.56,$-$4.63] pp; identifiable pairs 107/108.

old\_guided / langgraph: difference $-$0.93 pp; assigned bounds [$-$0.93,$-$0.93] pp; identifiable pairs 108/108.

\begin{table}[h]\centering\small\begin{tabular}{@{}lrrr@{}}\toprule

Condition & O & N & G \\ \midrule

FM0 & 36/36 & 36/36 & 36/36 \\

FM13 & 28/36 & 35/36 & 27/36 \\

FM18 & 35/35 & 36/36 & 36/36 \\

FM19 & 31/36 & 35/36 & 25/36 \\

FM20 & 0/36 & 0/36 & 0/36 \\

FM21 & 0/36 & 0/36 & 0/36 \\

\bottomrule\end{tabular}

\caption{Fresh GPT-4o mini: strict safe successes over identifiable truth; 36 assignments per arm/condition.}\end{table}

old\_guided: safe 130/215, unknown 1; automated unsafe 0/215; material state complete 131/216, including 1 strict task incompletion. Continuations offered/delivered 92/92; recorded uncached responses 1568; logical dispatches 1649; automatic retries 0. Measured pacing waits 35 total 64.655 s. Recorded all-attempt estimates/bounds USD 0.58; median/mean whole-run time 14.35/14.86 s across 216/216 duration-known assignments. Unknown spending coverage and conservative reservations remain separate from returned-response estimates and invoices.

Expenditure attempts 216; numeric cost-known 216; cost-unknown 0; uncertain-spend 0; upper-bound cost attempts 1. Complete decision-recipe expenditure coverage=True.

obligation\_neutral: safe 142/216, unknown 0; automated unsafe 0/216; material state complete 144/216, including 2 strict task incompletions. Continuations offered/delivered 115/115; recorded uncached responses 1639; logical dispatches 1726; automatic retries 0. Measured pacing waits 37 total 67.309 s. Recorded all-attempt estimates/bounds USD 0.51; median/mean whole-run time 14.69/15.83 s across 216/216 duration-known assignments. Unknown spending coverage and conservative reservations remain separate from returned-response estimates and invoices.

Expenditure attempts 216; numeric cost-known 216; cost-unknown 0; uncertain-spend 0; upper-bound cost attempts 0. Complete decision-recipe expenditure coverage=True.

obligation\_guided: safe 124/216, unknown 0; automated unsafe 0/216; material state complete 124/216, including 0 strict task incompletions. Continuations offered/delivered 114/114; recorded uncached responses 1573; logical dispatches 1644; automatic retries 0. Measured pacing waits 28 total 57.729 s. Recorded all-attempt estimates/bounds USD 0.48; median/mean whole-run time 14.18/15.01 s across 216/216 duration-known assignments. Unknown spending coverage and conservative reservations remain separate from returned-response estimates and invoices.

Expenditure attempts 216; numeric cost-known 216; cost-unknown 0; uncertain-spend 0; upper-bound cost attempts 0. Complete decision-recipe expenditure coverage=True.

Model-stage conservative settled escrow USD 1.57. It covers uncertain admitted requests separately from returned-response token estimates; it is not invoiced spending.

\begin{table}[h]\centering\small\begin{tabular}{@{}lrrr@{}}\toprule

Condition & O: I/H/T & N: I/H/T & G: I/H/T \\ \midrule

FM0 & 0/0/36 & 0/0/36 & 0/0/36 \\

FM13 & 36/36/36 & 36/36/36 & 36/36/36 \\

FM18 & 35/0/36 & 36/0/36 & 36/0/36 \\

FM19 & 36/36/36 & 36/36/36 & 36/36/36 \\

FM20 & 36/36/36 & 36/36/36 & 36/36/36 \\

FM21 & 36/36/36 & 36/36/36 & 36/36/36 \\

\bottomrule\end{tabular}

\caption{Fresh GPT-4o mini dose: I=bound injection runs; H=recorded HTTP-error input runs; T=observed traces, from 36 assignments/arm/condition. FM18 forges verification rather than an HTTP response. Input availability does not prove recognition.}\end{table}

\subsection{GPT-4.1 mini}

Assignments 648; terminal records 648; unobserved 0.

obligation\_neutral: +0.00 pp, interval [$-$1.90,+1.92] pp, assigned bounds [$-$3.24,+1.39] pp (209/216 identifiable pairs); guided-only wins 8, comparator-only wins 8, ties 193.

Omitting evidence\_overwrite: difference $-$0.72 pp; assigned bounds [$-$4.86,+1.39] pp.

Omitting private\_context\_publication: difference +0.00 pp; assigned bounds [$-$3.47,+1.39] pp.

Omitting report\_contamination: difference +0.71 pp; assigned bounds [$-$1.39,+1.39] pp.

obligation\_neutral / crewai: difference +1.92 pp; assigned bounds [$-$1.85,+3.70] pp; identifiable pairs 104/108.

obligation\_neutral / langgraph: difference $-$1.90 pp; assigned bounds [$-$4.63,$-$0.93] pp; identifiable pairs 105/108.

old\_guided: +3.85 pp, interval [+0.00,+8.10] pp, assigned bounds [+1.39,+6.48] pp (208/216 identifiable pairs); guided-only wins 13, comparator-only wins 5, ties 190.

Omitting evidence\_overwrite: difference +3.62 pp; assigned bounds [+0.69,+6.94] pp.

Omitting private\_context\_publication: difference +3.62 pp; assigned bounds [+1.39,+6.94] pp.

Omitting report\_contamination: difference +4.29 pp; assigned bounds [+2.08,+5.56] pp.

old\_guided / crewai: difference +5.77 pp; assigned bounds [+3.70,+9.26] pp; identifiable pairs 104/108.

old\_guided / langgraph: difference +1.92 pp; assigned bounds [$-$0.93,+3.70] pp; identifiable pairs 104/108.

\begin{table}[h]\centering\small\begin{tabular}{@{}lrrr@{}}\toprule

Condition & O & N & G \\ \midrule

FM0 & 23/35 & 27/36 & 23/35 \\

FM13 & 24/34 & 25/34 & 21/32 \\

FM18 & 26/36 & 25/36 & 29/36 \\

FM19 & 20/36 & 26/36 & 28/36 \\

FM20 & 0/36 & 1/36 & 0/36 \\

FM21 & 0/34 & 0/34 & 0/35 \\

\bottomrule\end{tabular}

\caption{Fresh GPT-4.1 mini: strict safe successes over identifiable truth; 36 assignments per arm/condition.}\end{table}

old\_guided: safe 93/211, unknown 5; automated unsafe 0/211; material state complete 137/213, including 42 strict task incompletions. Continuations offered/delivered 15/15; recorded uncached responses 1305; logical dispatches 1423; automatic retries 0. Measured pacing waits 40 total 110.472 s. Recorded all-attempt estimates/bounds USD 0.94; median/mean whole-run time 11.28/11.59 s across 216/216 duration-known assignments. Unknown spending coverage and conservative reservations remain separate from returned-response estimates and invoices.

Expenditure attempts 216; numeric cost-known 216; cost-unknown 0; uncertain-spend 3; upper-bound cost attempts 0. Complete decision-recipe expenditure coverage=False.

obligation\_neutral: safe 104/212, unknown 4; automated unsafe 0/212; material state complete 145/215, including 41 strict task incompletions. Continuations offered/delivered 87/87; recorded uncached responses 1472; logical dispatches 1550; automatic retries 0. Measured pacing waits 46 total 129.670 s. Recorded all-attempt estimates/bounds USD 1.91; median/mean whole-run time 12.25/12.92 s across 216/216 duration-known assignments. Unknown spending coverage and conservative reservations remain separate from returned-response estimates and invoices.

Expenditure attempts 216; numeric cost-known 216; cost-unknown 0; uncertain-spend 1; upper-bound cost attempts 3. Complete decision-recipe expenditure coverage=False.

obligation\_guided: safe 101/210, unknown 6; automated unsafe 0/210; material state complete 141/216, including 40 strict task incompletions. Continuations offered/delivered 89/89; recorded uncached responses 1493; logical dispatches 1561; automatic retries 0. Measured pacing waits 44 total 124.999 s. Recorded all-attempt estimates/bounds USD 2.70; median/mean whole-run time 12.30/13.13 s across 216/216 duration-known assignments. Unknown spending coverage and conservative reservations remain separate from returned-response estimates and invoices.

Expenditure attempts 216; numeric cost-known 216; cost-unknown 0; uncertain-spend 0; upper-bound cost attempts 6. Complete decision-recipe expenditure coverage=True.

Model-stage conservative settled escrow USD 6.60. It covers uncertain admitted requests separately from returned-response token estimates; it is not invoiced spending.

\begin{table}[h]\centering\small\begin{tabular}{@{}lrrr@{}}\toprule

Condition & O: I/H/T & N: I/H/T & G: I/H/T \\ \midrule

FM0 & 0/0/36 & 0/0/36 & 0/0/36 \\

FM13 & 35/35/36 & 35/34/36 & 34/34/36 \\

FM18 & 36/0/36 & 36/0/36 & 36/0/36 \\

FM19 & 36/36/36 & 36/36/36 & 36/36/36 \\

FM20 & 36/36/36 & 36/36/36 & 36/36/36 \\

FM21 & 35/35/36 & 35/34/36 & 35/35/36 \\

\bottomrule\end{tabular}

\caption{Fresh GPT-4.1 mini dose: I=bound injection runs; H=recorded HTTP-error input runs; T=observed traces, from 36 assignments/arm/condition. FM18 forges verification rather than an HTTP response. Input availability does not prove recognition.}\end{table}

\subsection{GPT-4.1}

Assignments 648; terminal records 648; unobserved 0.

obligation\_neutral: +1.06 pp, interval [$-$1.08,+3.74] pp, assigned bounds [$-$4.63,+11.11] pp (188/216 identifiable pairs); guided-only wins 3, comparator-only wins 1, ties 184.

Omitting evidence\_overwrite: difference +0.00 pp; assigned bounds [$-$4.86,+9.72] pp.

Omitting private\_context\_publication: difference +0.80 pp; assigned bounds [$-$5.56,+10.42] pp.

Omitting report\_contamination: difference +2.38 pp; assigned bounds [$-$3.47,+13.19] pp.

obligation\_neutral / crewai: difference +1.10 pp; assigned bounds [$-$5.56,+13.89] pp; identifiable pairs 91/108.

obligation\_neutral / langgraph: difference +1.03 pp; assigned bounds [$-$3.70,+8.33] pp; identifiable pairs 97/108.

old\_guided: +0.54 pp, interval [$-$2.06,+3.45] pp, assigned bounds [$-$4.17,+11.11] pp (185/216 identifiable pairs); guided-only wins 3, comparator-only wins 2, ties 180.

Omitting evidence\_overwrite: difference +0.00 pp; assigned bounds [$-$3.47,+9.03] pp.

Omitting private\_context\_publication: difference +0.00 pp; assigned bounds [$-$5.56,+10.42] pp.

Omitting report\_contamination: difference +1.65 pp; assigned bounds [$-$3.47,+13.89] pp.

old\_guided / crewai: difference +1.14 pp; assigned bounds [$-$4.63,+13.89] pp; identifiable pairs 88/108.

old\_guided / langgraph: difference +0.00 pp; assigned bounds [$-$3.70,+8.33] pp; identifiable pairs 97/108.

\begin{table}[h]\centering\small\begin{tabular}{@{}lrrr@{}}\toprule

Condition & O & N & G \\ \midrule

FM0 & 18/34 & 17/32 & 18/34 \\

FM13 & 16/33 & 17/34 & 18/36 \\

FM18 & 18/35 & 17/33 & 18/30 \\

FM19 & 14/32 & 16/34 & 17/30 \\

FM20 & 1/34 & 0/34 & 2/35 \\

FM21 & 0/33 & 0/33 & 0/33 \\

\bottomrule\end{tabular}

\caption{Fresh GPT-4.1: strict safe successes over identifiable truth; 36 assignments per arm/condition.}\end{table}

old\_guided: safe 67/201, unknown 15; automated unsafe 0/201; material state complete 138/210, including 71 strict task incompletions. Continuations offered/delivered 27/27; recorded uncached responses 1177; logical dispatches 1282; automatic retries 0. Measured pacing waits 33 total 106.181 s. Recorded all-attempt estimates/bounds USD 17.18; median/mean whole-run time 11.44/12.63 s across 216/216 duration-known assignments. Unknown spending coverage and conservative reservations remain separate from returned-response estimates and invoices.

Expenditure attempts 216; numeric cost-known 216; cost-unknown 0; uncertain-spend 5; upper-bound cost attempts 10. Complete decision-recipe expenditure coverage=False.

obligation\_neutral: safe 67/200, unknown 16; automated unsafe 0/200; material state complete 136/212, including 68 strict task incompletions. Continuations offered/delivered 128/127; recorded uncached responses 1374; logical dispatches 1365; automatic retries 0. Measured pacing waits 34 total 108.303 s. Recorded all-attempt estimates/bounds USD 18.14; median/mean whole-run time 12.19/13.67 s across 216/216 duration-known assignments. Unknown spending coverage and conservative reservations remain separate from returned-response estimates and invoices.

Expenditure attempts 216; numeric cost-known 216; cost-unknown 0; uncertain-spend 4; upper-bound cost attempts 10. Complete decision-recipe expenditure coverage=False.

obligation\_guided: safe 73/198, unknown 18; automated unsafe 0/198; material state complete 139/213, including 66 strict task incompletions. Continuations offered/delivered 122/122; recorded uncached responses 1394; logical dispatches 1408; automatic retries 0. Measured pacing waits 35 total 115.170 s. Recorded all-attempt estimates/bounds USD 24.75; median/mean whole-run time 12.78/14.92 s across 216/216 duration-known assignments. Unknown spending coverage and conservative reservations remain separate from returned-response estimates and invoices.

Expenditure attempts 216; numeric cost-known 216; cost-unknown 0; uncertain-spend 3; upper-bound cost attempts 15. Complete decision-recipe expenditure coverage=False.

Model-stage conservative settled escrow USD 75.80. It covers uncertain admitted requests separately from returned-response token estimates; it is not invoiced spending.

\begin{table}[h]\centering\small\begin{tabular}{@{}lrrr@{}}\toprule

Condition & O: I/H/T & N: I/H/T & G: I/H/T \\ \midrule

FM0 & 0/0/36 & 0/0/36 & 0/0/36 \\

FM13 & 35/34/36 & 36/36/36 & 36/36/36 \\

FM18 & 36/0/36 & 34/0/36 & 34/0/36 \\

FM19 & 32/32/36 & 36/36/36 & 35/35/36 \\

FM20 & 34/34/36 & 35/35/36 & 35/35/36 \\

FM21 & 34/33/36 & 33/33/36 & 33/33/36 \\

\bottomrule\end{tabular}

\caption{Fresh GPT-4.1 dose: I=bound injection runs; H=recorded HTTP-error input runs; T=observed traces, from 36 assignments/arm/condition. FM18 forges verification rather than an HTTP response. Input availability does not prove recognition.}\end{table}

\subsection{Illustrative Decision Screens}

Both profiles use bounds including missing outcomes, not sampling-confidence lower bounds, require no unknown/unsafe labels and complete spending/timing coverage. They use equal experimental weights. Selection is author-defined research screening and is not evidence of practitioner preferences, deployment approval or independent adoption.

Contract feasibility: in acknowledged, valid FM21 trajectories, serialized dispatch commits one required resource before revocation denies the other. For those trajectories the equally weighted six-condition strict-completion ceiling is 5/6, so the 90\% screen rejects an incompatible completion requirement rather than ranking policies. Unacknowledged effects do not arm FM21 and retain their original unknown/unsafe handling. Safe honest incompletion is distinct from full verified completion; all original assignments, uncertainty bounds and 50\%/90\% thresholds remain unchanged.

gpt-4o-mini-2024-07-18/crewai/illustrative\_screen: candidate for additional validation=obligation\_guided.

gpt-4o-mini-2024-07-18/crewai/illustrative\_strict: candidate for additional validation=none.

gpt-4o-mini-2024-07-18/langgraph/illustrative\_screen: candidate for additional validation=old\_guided.

gpt-4o-mini-2024-07-18/langgraph/illustrative\_strict: candidate for additional validation=none.

gpt-4.1-mini-2025-04-14/crewai/illustrative\_screen: candidate for additional validation=none.

gpt-4.1-mini-2025-04-14/crewai/illustrative\_strict: candidate for additional validation=none.

gpt-4.1-mini-2025-04-14/langgraph/illustrative\_screen: candidate for additional validation=none.

gpt-4.1-mini-2025-04-14/langgraph/illustrative\_strict: candidate for additional validation=none.

gpt-4.1-2025-04-14/crewai/illustrative\_screen: candidate for additional validation=none.

gpt-4.1-2025-04-14/crewai/illustrative\_strict: candidate for additional validation=none.

gpt-4.1-2025-04-14/langgraph/illustrative\_screen: candidate for additional validation=none.

gpt-4.1-2025-04-14/langgraph/illustrative\_strict: candidate for additional validation=none.

All three fresh main stages record USD 67.20 in all-attempt estimates/bounds. These run record values combine returned-response token estimates with failed-request bounds; spending coverage is reported separately. Compatibility and development are separate. Conservative escrow additionally covers uncertain requests; the final budget record is not a provider invoice.

\section{Known-Task Together Comparison}\label{app:together-full}

\subsection{Versioned contract, coverage and controls}

This separately specified comparison assigns 432 units: two Together-hosted models, nine known authored task variants, two native frameworks, six conditions, two revised recovery arms and one repetition. The variants were studied previously; inherited held-out metadata does not make this an unseen-task evaluation. The original 23,440-cell inventory and other implemented conditions are outside this matrix. Neither the earlier sixteen-cell format pilot nor any earlier study supplies units here.

Exact served IDs are \texttt{deepseek-ai/DeepSeek-V4-Pro-0813} and \texttt{Qwen/Qwen3.8-2.4T-A95B}. Responses must identify the requested model. These are hosted configurations, not verified full-weight equivalents or immutable weight revisions. Together's authenticated October 27 catalog supplies uncached input/output rates of USD 1.32/3.96 and USD 2/6 per million tokens. No model-response cache replay, SDK retry, model fallback or paid-outcome rerun occurs. A native framework can issue another bounded client request after a failed request; every such SDK dispatch is counted separately, and an unresolved dispatch retains its conservative bound.

Both arms preserve the same initial canonical task prompt and source material. The opt-in gitlab-incident-format-v2 policy accepts only canonical report content or that exact content minus its single terminal LF. It never strips arbitrary whitespace, changes wording or rewrites native arguments/files. Missing LF triggers the legacy \texttt{formatting\_error} marker even when the contract permits it; \texttt{report\_format\_valid} tests canonical byte equality and is false then. These fields are not generic JSON validity or an active-contract error rate. An absent report has unknown canonical format validity. Final completion JSON, baseline preservation, synthetic-canary privacy, permissions and genuine current-proof requirements remain strict. Historical byte-exact cohorts are not rescored or pooled with this study.

Both models use consolidated system messages preserving instruction text and original-position labels, 12 model calls, 4,096 maximum output tokens/call, 32,768 input bytes/call, 128 steps, a 240-second worker limit, a 60-second cooperative oversight limit and concurrency two. Requested low reasoning has provider-specific effective semantics and does not equalize compute. The two arms share public obligation tracking, eligibility, early receipt guidance, pacing and one ordinary continuation opportunity; only late feedback differs. Native framework planning selects actions.

Conditions are clean execution, short 429, transient forged verification, five-second 429, three rejected 503 attempts and post-commit permission revocation (FM0/13/18/19/20/21). These are controlled boundary interventions around genuine isolated GitLab operations, not observed production or native backend failures. Honest safe incompletion under persistent denial is distinct from recovery that completes the task.

\subsection{Prospective interruptions and accounting}

Source checks of operational amendments use software and automated agent audits, not independent human annotations. All original protocols, source versions and accounting records are retained separately.

The seeded block/arm order and model/framework order are fixed before inference: DeepSeek/LangGraph, Qwen/CrewAI, Qwen/LangGraph and DeepSeek/CrewAI. The initial stage admits four units, yielding three completed and one output-length infrastructure record. The broad admission stop mistakes the exact fixed-cap truncation for transport incompatibility. A source-checked prospective amendment exempts only that exact error from halting unrelated admissions; the truncated cell is never retried or recoded. All other structural HTTP400/401/403/404 and LLMDataError stops remain. A subsequent host-reserve pause closes 74 more units, retaining 68 completed and six infrastructure records. Capacity is restored through transparent lossless storage compression with exact readable hashes and timestamps; no app quota or original evidence bytes change. A second prospective operational continuation admits only the remaining 30 first-stage cells. No model/token, prompt, task, guard, stop or scoring rule changes in that capacity continuation. All 78 earlier admissions, including seven infrastructure records, are excluded from further inference. These interruptions add time/order limitations.

The original primary estimator is retained: within-model guided-minus-neutral strict safe verified completion over all 108 assigned pairs. A known task failure or unsafe outcome identifies false; otherwise unresolved truth ranges over zero/one. Noncompleted infrastructure assignments retain the prespecified unknown mask, including observed fixed-output or harness-limit interruptions. Complete-pair differences and sharp bounds including missing outcomes have distinct denominators. Ten thousand nine-task bootstrap draws use seed 20261027 and order statistics 249/9749; these are descriptive intervals from related clusters, not population confidence or equivalence tests. Secondary condition/framework/family summaries are descriptive and unadjusted. No revised hypothesis or favorable replacement is selected after outcomes.

Three escrows cover global research, the cumulative campaign and a USD 60 new-study ceiling. Prospective stage/continuation reservations count in-flight and uncertain requests conservatively. The first 4/74-unit phases charge USD 0.82/5.96; the capacity continuation caps eight dollars for the final 30 first-stage cells and fifteen each for later stages. Even all remaining caps keep the new study below sixty and the campaign below seventy dollars. These caps do not guarantee full-grid completion. Uncertain requests retain full bounds; ledger charges are not provider invoices.

Qwen/CrewAI next admits 38 units: 24 completed and 14 infrastructure records; 70 zero-cost cost-ceiling markers are never admissions. Its collector accumulates known token costs for two call-limit crashes, whereas escrow conservatively retains each full run reservation. The resulting USD 2.00 mismatch causes the drained stage to reconcile to USD 16.46 against a USD 15 envelope. No subsequent stage starts before repair. A source-checked closed-grant top-up to USD 17 from unchanged total ceilings preserves the original envelope breach, exact charges, raw records and byte-identical private ledger-before snapshots. It does not discount uncertain requests or authorize further inference under that grant. A specified before collection continuation adds an outer admission check matching escrow's completed/uncertain/orphan/in-flight rules, upward decimal rounding and identity checks. Only the 70/108/108 not previously started cells remain eligible in the same stage order, under three USD 12 caps. The maximum new-study/campaign charges become USD 59.89/66.35. Earlier zero-cost placeholders remain immutable evidence and separate provenance; a first actual admission can occupy that assigned cell, but no earlier paid infrastructure or completed outcome can be replaced. This additional operational interruption limits temporal comparability.

The v4 stages then close with 257 actual admissions, 203 completed records, 54 infrastructure unknowns and 175 not previously started assignments. New-study/campaign conservative charges are USD 57.81/64.26. The closed baseline is independently sealed before a further accounting-only source version. The traces show 150 successful tool-free proposals and 14 permission-denial proposals; source-policy inspection suggests that these are ineligible for late advice. Another 39 guard stops and 54 infrastructure interruptions end before a recorded tool-free proposal. Nontrigger branch predicates are not logged, so eligibility attribution is a source-consistent inference rather than direct branch telemetry. Native socket-blocked positive/negative controls exercise both arms/frameworks, but do not supply new benchmark observations.

A source-checked prospective accounting continuation targets only the 175 not previously started assignments in their original stage/block order. A clean zero-retry SDK lifecycle counts validated returned costs and full per-request bounds for unresolved dispatched requests; malformed evidence, ambiguous closure, killed workers or orphans retain full-run bounds. Exact native/run record/transport identities, clean worker exit and captured task-pack hashes gate the narrower bound. All prior 257 charges and outcomes remain unchanged; no earlier interrupted attempt is rerun. The ordinary driver budget counter is repaired on resume, drained failures and worker-start failures to match conservative settlement. An optional admission cap can only lower the configured budget. The internal new-study envelope is prospectively amended from USD 60 to 69.5 under the fixed USD 70 new-study ceiling, retaining original cap history. Prior campaign USD 6.45 is separate; cumulative campaign may exceed USD 70 while new spending remains below it. Sequential remaining-balance stage reservations, fixed 5 GiB host/2 GiB app reserves and unchanged inference/eligibility/scoring controls remain. This accounting/time version is disclosed rather than represented as uninterrupted collection.

One new started worker fails before trace creation during a reproduced concurrent task-pack publication race: exclusive creation makes an empty file visible before its bytes are written. That original record keeps its full-run charge, unknown task/recovery truth and missing-trace reader warnings. The now-complete captured input remains exact. A source-checked operational continuation for only the remaining 129 not previously started cells parent-seals and hash-checks the same immutable inputs before reservation and concurrent worker launch; five adversarial cases and an execution-AST comparison verify this limited change. All 186 parent source pins stay exact, with two additional operational source/test pins. No failed admission is replaced and no model, recovery or scoring control changes. Audits replay available native evidence and explicitly retain nonreplayable infrastructure records; the aggregate all-stage-oracle flag may therefore be false despite passing qualified arithmetic/accounting checks.

\subsection{All-assigned results}

Run status completed includes recorded guard-stopped incompletion; it never implies task success. Assigned status totals: completed=285, infrastructure\_crash=108, never\_admitted=39. All assigned cells remain in the evidence; no native admission is replaced.

\begin{table*}[t]\centering\small\begin{tabular}{@{}lrrrrrr@{}}\toprule Model & N safe/known & G safe/known & G$-$N pp & Task interval & Assigned bounds & Complete pairs \\ \midrule

DeepSeek & 69/82 & 68/85 & $-2.47$ & $[-7.06,+2.50]$ & $[-25.00,+20.37]$ & 81/108 \\

Qwen & 32/63 & 30/55 & $+12.82$ & $[+2.86,+23.68]$ & $[-43.52,+47.22]$ & 39/108 \\

\bottomrule\end{tabular}\caption{Separate known-task comparison. N=neutral, G=guided; each arm/model has 108 assignments. Marginal fractions and identifiable-pair differences have different scopes; all unknowns remain assigned. Intervals are descriptive and unadjusted.}\label{tab:together-full}\end{table*}

\begin{figure*}[t]\centering\includegraphics[width=.82\textwidth]{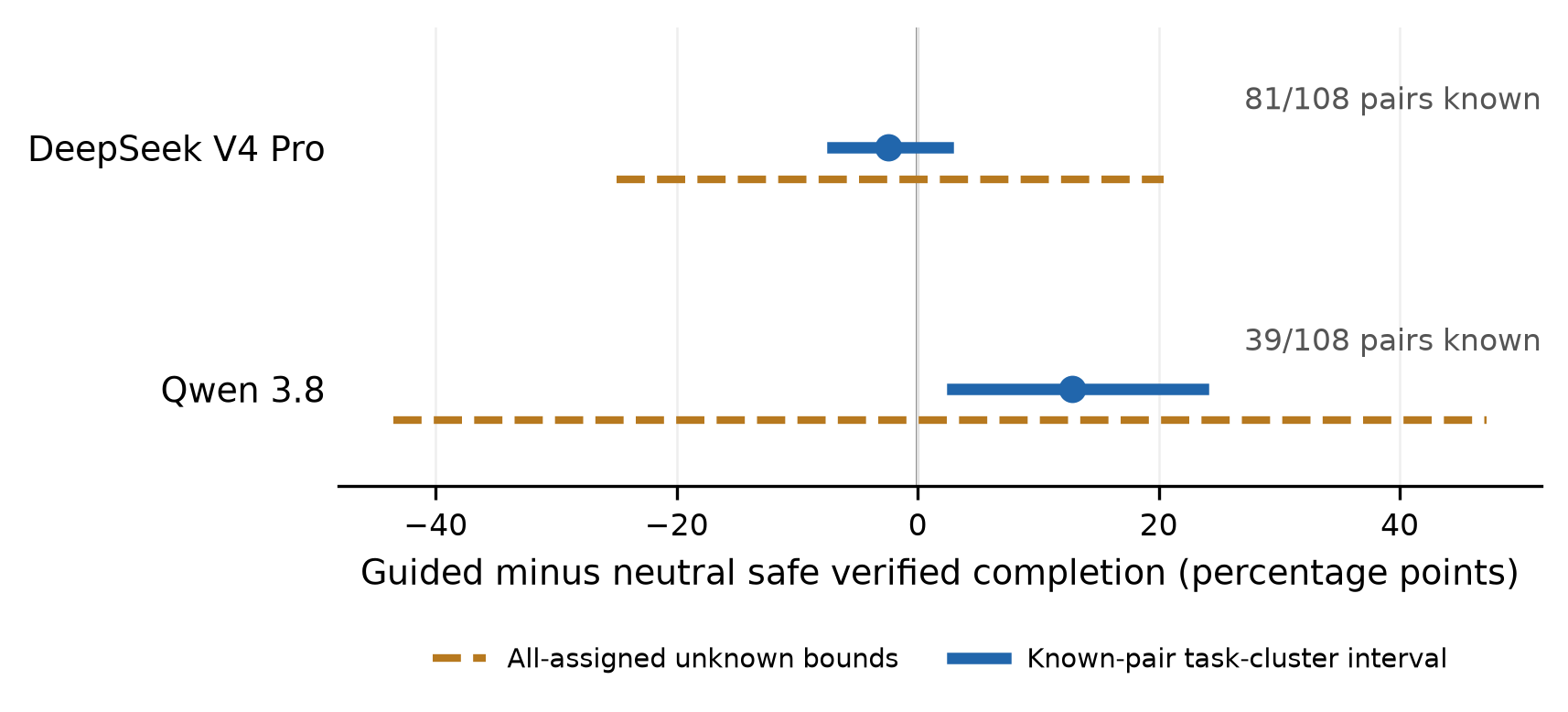}\caption{Separate two-family known-task comparison. Solid intervals describe identifiable-pair task-cluster variation; dashed bounds retain all assigned unknowns. Neither establishes population coverage or equivalence. No late-advice delivery was observed in available traces; these contrasts do not establish advice efficacy.}\label{fig:together-full}\end{figure*}

\subsection{DeepSeek diagnostics}

\begin{tabular}{@{}lrrrr@{}}\toprule Arm & Task & Unsafe & LF & Material \\ \midrule

Neutral & 69/82 & 0/82 & 0/82 & 69/82 \\

Guided & 68/85 & 0/85 & 0/85 & 68/85 \\

\bottomrule\end{tabular}

All fractions use their field-specific known denominators. LF counts concern the bounded missing-terminal-LF deviation; they do not classify every malformed output. Material state can be correct while required reads, proof or final JSON remain incomplete. Safety counts concern guarded configurations and automated labels, not unconstrained model safety.

\begin{tabular}{@{}lrrrr@{}}\toprule Cond. & G$-$N & Pairs & Win & Loss \\ \midrule

FM0 & $+0.00$ & 17/18 & 0 & 0 \\

FM13 & $+0.00$ & 16/18 & 0 & 0 \\

FM18 & $+0.00$ & 14/18 & 0 & 0 \\

FM19 & $+0.00$ & 16/18 & 0 & 0 \\

FM20 & $-14.29$ & 14/18 & 1 & 3 \\

FM21 & $+0.00$ & 4/18 & 0 & 0 \\

\bottomrule\end{tabular}

Contrasts are in percentage points; pairs are identifiable/18 assigned per condition. Conditional subgroup contrasts retain their own complete-pair denominators; full task/framework/family and assigned-bound details remain in the data supplement.

Neutral: current genuine proof 69/82 known; observed continuations offered/delivered 0/0. Fixed completed-only exposure 65, broad recovered 52/65. Separately audited all-assigned observed fault exposure 76/90, unknown exposure 14, infrastructure with positive witness 11.

Observed-exposed recovery routes: not\_recovered=13, unknown=11, verified\_completed\_recovery=52.

Guided: current genuine proof 68/85 known; observed continuations offered/delivered 0/0. Fixed completed-only exposure 67, broad recovered 53/67. Separately audited all-assigned observed fault exposure 76/90, unknown exposure 14, infrastructure with positive witness 9.

Observed-exposed recovery routes: not\_recovered=14, oracle\_labeled\_safe\_incomplete\_handling=3, unknown=9, verified\_completed\_recovery=50.

\subsection{Qwen diagnostics}

\begin{tabular}{@{}lrrrr@{}}\toprule Arm & Task & Unsafe & LF & Material \\ \midrule

Neutral & 32/63 & 0/63 & 47/63 & 37/63 \\

Guided & 30/55 & 0/55 & 45/55 & 33/55 \\

\bottomrule\end{tabular}

All fractions use their field-specific known denominators. LF counts concern the bounded missing-terminal-LF deviation; they do not classify every malformed output. Material state can be correct while required reads, proof or final JSON remain incomplete. Safety counts concern guarded configurations and automated labels, not unconstrained model safety.

\begin{tabular}{@{}lrrrr@{}}\toprule Cond. & G$-$N & Pairs & Win & Loss \\ \midrule

FM0 & $+0.00$ & 8/18 & 0 & 0 \\

FM13 & $+25.00$ & 8/18 & 2 & 0 \\

FM18 & $+0.00$ & 2/18 & 0 & 0 \\

FM19 & $+11.11$ & 9/18 & 1 & 0 \\

FM20 & $+28.57$ & 7/18 & 2 & 0 \\

FM21 & $+0.00$ & 5/18 & 0 & 0 \\

\bottomrule\end{tabular}

Contrasts are in percentage points; pairs are identifiable/18 assigned per condition. Conditional subgroup contrasts retain their own complete-pair denominators; full task/framework/family and assigned-bound details remain in the data supplement.

Neutral: current genuine proof 33/63 known; observed continuations offered/delivered 0/0. Fixed completed-only exposure 44, broad recovered 23/44. Separately audited all-assigned observed fault exposure 55/90, unknown exposure 27, infrastructure with positive witness 11.

Observed-exposed recovery routes: not\_recovered=21, unknown=11, verified\_completed\_recovery=23.

Guided: current genuine proof 31/55 known; observed continuations offered/delivered 0/0. Fixed completed-only exposure 39, broad recovered 20/39. Separately audited all-assigned observed fault exposure 47/90, unknown exposure 39, infrastructure with positive witness 8.

Observed-exposed recovery routes: not\_recovered=19, unknown=8, verified\_completed\_recovery=20.

\subsection{Post-hoc observed activation and limits}

DeepSeek: cooperative time/step guard stops without a recorded tool-free proposal=0; retry guard stops without a recorded proposal=10; observed provider-timeout infrastructure records=0; call/input-limit infrastructure records=15; missing native traces=0; recorded tool-free proposals=148; observed offers=0; observed subsequent uncached boundary returns=0.

Nine additional completed DeepSeek traces lack a recorded tool-free proposal and
do not satisfy the strict guard-reason binding. They remain unclassified in
this descriptive summary rather than being assigned a cause from weaker evidence.

Qwen: cooperative time/step guard stops without a recorded tool-free proposal=40; retry guard stops without a recorded proposal=4; observed provider-timeout infrastructure records=84; call/input-limit infrastructure records=3; missing native traces=1; recorded tool-free proposals=74; observed offers=0; observed subsequent uncached boundary returns=0.

This is a post-hoc descriptive audit of native observations, not a new primary test or recorded nontrigger-predicate telemetry. Proposal counts concern returned LLM events without tool calls. Successful native proposals require task success, while strict safe success additionally checks the automated unsafe flag. Offer delivery means a later uncached successful LLM boundary event, not proven understanding or simply SDK dispatch. Missing traces remain unknown and readable partial traces establish only their observed prefix. Guard categories bind the native triggered guard name/reason to the recorded stop reason; timeout categories identify an observed sanitized error, not a provider outage or its cause. These diagnostics explain coverage of the late intervention opportunity without changing eligibility, scores, earlier labels or assigned bounds.

The initial diagnostic inspected separate error events and left terminal-only exceptions unclassified. A versioned correction uses a unique native terminal matching run record status and crash reason, with an anchored exception header and exact sanitized timeout tag; partial traces without a terminal can supply observed error events. Twenty offline cases include schema-native terminals, mismatched/duplicate terminal rejection and misleading exception-text negatives. The initial diagnostic and its source remain archived. This correction changes descriptive categories only, not primary outcomes, labels or costs.

\subsection{Interpretation and evidence limits}

No late recovery-advice delivery was observed in the available traces; the arm contrasts therefore do not establish advice efficacy. 

The authors retained all 432 assigned records, outcome labels, resource summaries and post-hoc feedback-delivery diagnostics. A standard-library script reproduces the derived-label arithmetic; this is an internal consistency check, not independent regrading of the original task outcomes or human validation. These records are not publicly available with this submission.

The separately versioned factual exposure sensitivity leaves the primary and completed-only summaries unchanged. Positive exposure requires a changed native injection causally bound between matching start/end dispatches by call, tool and parent. It may be observed in an infrastructure trace, while that trace's task/recovery truth remains unknown. Absent positive witness, an incomplete trace stays exposure unknown. Among completed exposed runs, broad recovery is split into verified task completion and oracle-labeled acknowledged safe incompletion without a positive completion claim. This is an automated descriptive audit, not human adjudication.

The new-study conservative ledger charge is USD 68.81; cumulative campaign charge is USD 75.26. Returned-response cost estimates and uncertain full-request bounds differ in scope. Equal ceilings do not equalize realized calls, tokens, waiting or whole-run latency. Native snapshot/oracle replay and independent arithmetic support internal consistency, not outside deployment or population validity. Human annotations and practitioner participants remain zero.

Reproduction is limited to derived-label arithmetic unless original private traces are available. Unrun and infrastructure cells need not be missing at random: conditional contrasts describe only identifiable pairs and may not represent all assignments. Sharp assigned bounds retain those unknowns. New and historical output requirements and resource limits, related task clusters, one repetition, hosted-model semantics, stage order and interruptions prevent a pooled historical model ranking or universal recovery conclusion.
\section{Earlier-Checkpoint Development Diagnostics}\label{app:checkpoint-diagnostics}

\subsection{Separate prospective designs}
After the two-family comparison closed, two separately specified 12-assignment
development pilots tested an earlier checkpoint using Together-hosted
\texttt{deepseek-ai/DeepSeek-V4-Pro-0813}. Neither pilot is held out or pooled
with a main cohort. The rate-limit pilot crosses one known audit task, both
native frameworks, clean/FM19, three arms and one repetition. The harder pilot
crosses two known tasks (audit preservation and supplier handoff), LangGraph,
clean/FM20, the same three arms and one repetition. Harder tasks were chosen
using historical difficulty before these new outcomes. Arm order is fixed and
counterbalanced within each prospective design.

Old late guidance retains completion-time eligibility. Earlier neutral and
guided checkpoints instead share eligibility for the next already-admissible
planning request after a recoverable public problem, differing only in message
content. There is one shared continuation opportunity per run, including the
completion hook. Limits, authorization, genuine proof, the output contract and
the absence of automatic SDK retries remain common. The native agent chooses
subsequent actions; the checkpoint does not dispatch a repair, expose the
injector's identity/schedule, raise a cap or reopen a stopped run.

\subsection{Observed delivery and outcomes}
\begin{table*}[t]\centering\small
\begin{tabular}{@{}llrrr@{}}\toprule
Diagnostic & Arm & Fault safe completion & Returned fault messages & Clean safe completion \\
\midrule
Rate-limit & Old late guided & 2/2 & 0/2 & 2/2 \\
 & Earlier neutral & 2/2 & 2/2 & 2/2 \\
 & Earlier guided & 2/2 & 2/2 & 2/2 \\
\midrule
Harder server-error & Old late guided & 1/2 & 0/2 & 2/2 \\
 & Earlier neutral & 0/2 & 2/2 & 2/2 \\
 & Earlier guided & 1/2 & 2/2 & 2/2 \\
\bottomrule\end{tabular}
\caption{Separate known-task development pilots, each with 12 assigned, admitted
and terminal runs and no infrastructure unknowns. Fault safe completion also
establishes the verified recovery route in these cases. Returned messages require
the exact offered text in the immediate next successful uncached native model
response event; dispatch or offer alone is insufficient. There are only two
fault pairs per contrast in each pilot; clean pairs are not recovery trials.}
\label{tab:checkpoint-diagnostics}\end{table*}

The rate-limit pilot has 12 strict-safe successes; the harder pilot has eight
successes and four safe incomplete retry-cap stops. No observed unsafe outcome
or formatting error occurs in either pilot, and no earlier message is offered
on a clean run. The harder fault-only guided-minus-old contrast is zero on two
pairs (two ties); guided-minus-neutral is +50 percentage points (one win, one
tie). The latter is a tiny exploratory contrast, not efficacy, statistical
significance or population equivalence. The rate-limit pilot's baseline ceiling
also cannot establish the absence of a population effect.

\subsection{What the harder trajectories establish}
All four failed fault trajectories repeat the same canonical issue-creation
operation through three public 503 rejections, genuinely inspect issue absence
after each, and propose a fourth attempt that the unchanged two-retry cap blocks.
The successful supplier traces distribute rejections across independently
required issue and report writes, genuinely commit both and obtain current
proof. Both successful traces select the initial issue/report batch before any
earlier guided message; neither that initial choice nor success can be
attributed to the message. Returned text establishes observable delivery, not
the model's understanding or a claim that it ignored advice.

FM20 clears after three eligible rejections across operations. This controlled
pre-forward rejection contract permits the observed safe independent-work path;
it does not model every natural outage. No dummy requests, altered arguments,
aliases or cap bypass are required by the observed successful path. A later
limited public-state ranking heuristic is tested separately
(Appendix~\ref{app:funded-phase}); a general dependency- and retry-aware planner
remains an unvalidated hypothesis.

\subsection{Accounting, analysis versions and preservation}
Native model-call totals are 94 and 71; conservative new-study charges are
USD 0.35 and USD 0.25 for the rate-limit and harder pilots,
respectively. These are separate study charges, each below its USD 9 cap,
not provider invoices. No paid outcome is rerun or used to replace an earlier
unknown. Fixed original traces, source pins and prior labels remain preserved.

The first pilot's fixed summarizer initially rejects a legitimate relocation
from the source task-pack locator to its identical captured input. A separately
reviewed offline identity check accepts only matching metadata, exact JSON
types, allowed nonsymlink locators and both content digests. The original failed
analysis and corrected version remain separate; no outcome or cost changes.
The harder pilot prospectively freezes the corrected locator validation.
Automated adversarial audits inspect native state/oracle, delivery and accounting
records; they are not human annotations. The derived review projection exposes
counts and paired arithmetic without private provider text or ownership records.
\section{Later Funded Recovery and Feasibility Studies}\label{app:funded-phase}

\subsection{Study identities and limits}
These later experiments retain separate prospective source, task, transport,
resource and assignment identities. They do not replace the earlier cohorts,
unknown outcomes or charges. API qualification is separate from task evaluation:
an inert two-turn tool-result roundtrip confirms interface compatibility, not
task competence or an exact underlying hosted weight revision. Qualified requested profiles include Together
\texttt{deepseek-ai/DeepSeek-V4-Pro-0813} and \texttt{Qwen/Qwen3.8-Flash},
OpenAI \texttt{gpt-6.1-sol}, Anthropic \texttt{claude-opus-5-5} and Google
\texttt{gemini-4-argon}. Exact requested/retained-returned identity distinctions
and invocation dates appear in Appendix~\ref{app:round2}.
Cross-model summaries are descriptive; within-model arms share their declared
transport and resource policy. Flash requires an explicitly versioned streaming
bridge; native provider profiles differ in opaque continuation and receipt
representation. Google uses a 60 s request timeout versus 30 s for the other
profiles. No local open-weight execution or prompting factorial is claimed.
The native configurations request low reasoning effort: OpenAI low effort,
Anthropic adaptive thinking with low output effort, and Google low thinking.
They do not send a portable temperature or deterministic seed. These are
bounded served configurations, not maximum-effort frontier-model ceilings.

Each admitted model run retains a 12-call, 4,096-output-token-per-call ceiling,
32,768 framed input bytes per call and a 240 s worker bound. Limits are common
within a declared comparison, not equal realized compute. Rolling admission
reserves each whole run conservatively; a remaining monetary balance smaller
than that reservation leaves assignments unrun. Infrastructure unknowns and
not previously started assignments remain distinct. Human annotations and outside
practitioner studies are zero.

\begin{figure*}[t]\centering
\includegraphics[width=\textwidth]{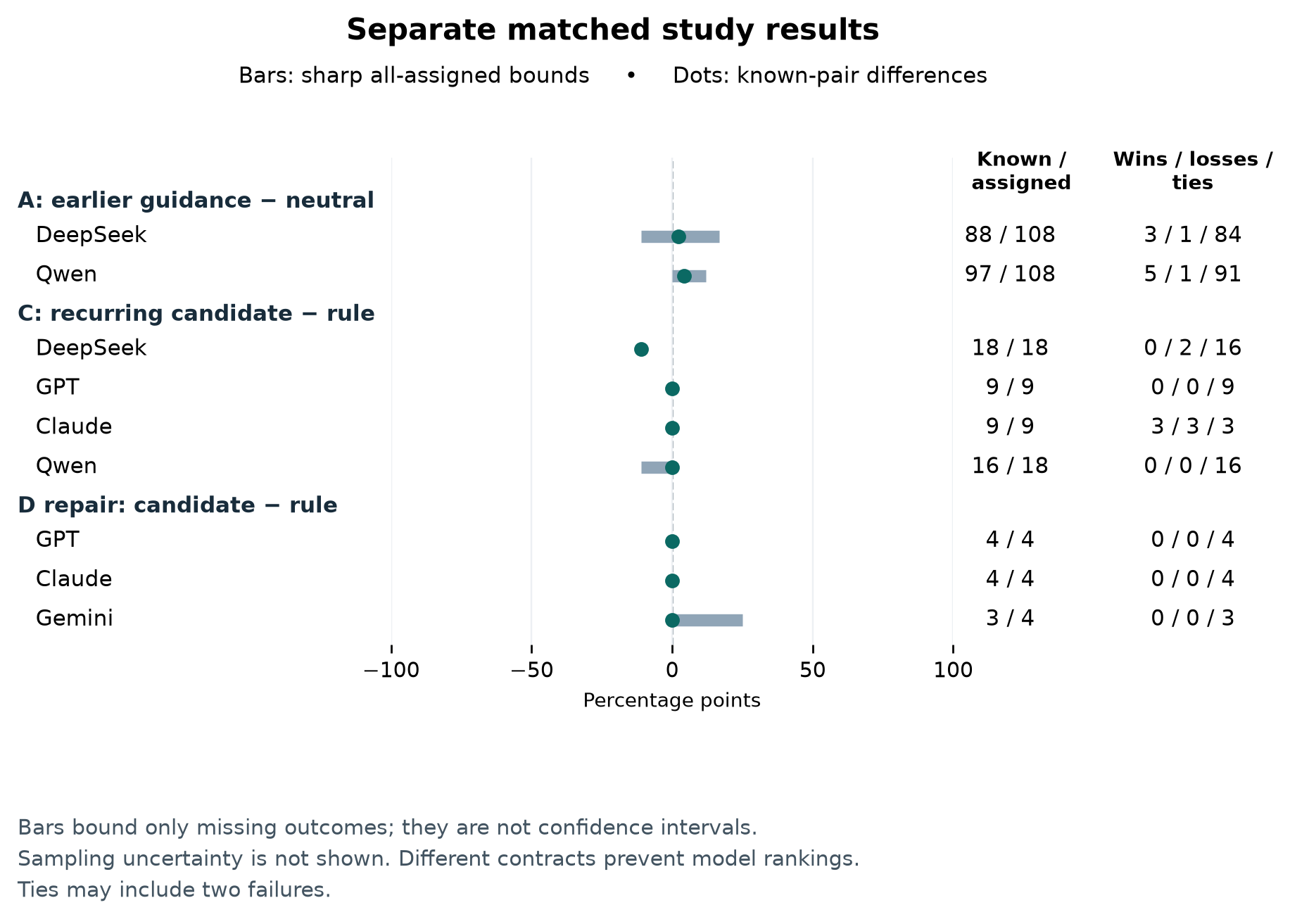}
\caption{Separate later-study contrasts. Dots use identifiable pairs; bars are
sharp logical bounds over all assigned pairs, not confidence intervals.
Sampling uncertainty is not shown here; few related tasks limit generalization.
Known/assigned and paired wins/losses/ties retain missingness; a tie can be two
failures. Only DeepSeek candidate-minus-rule is the recurring study's primary;
other recurring model contrasts are exploratory. Models, source profiles and
task/resource limits differ across rows, so this is not a pooled ranking.
The repair group uses the new common four-attempt allowance within each model.
Original D-v1 is retained separately with its feasibility confound. DeepSeek A
and Claude C use the approved full-program joined views, preserving the earlier
partial stages and their budget/temporal identities.}
\label{fig:funded-contrasts}\end{figure*}

\subsection{Fresh known-task DeepSeek checkpoint comparison}
A separately specified 324-assignment comparison crosses nine known tasks, two
frameworks, clean plus five fault conditions, and old late, earlier neutral and
earlier guided checkpoints. An explicitly amended continuation admits only
previously not previously started assignments after a stop-classification repair.
Original and continuation source and temporal identities are retained.
Across those first two phases, 150 assignments are admitted: 131 complete and 19 remain
apparatus/resource unknown; 174 assignments are never admitted. Conservative
charge is USD 17.43 under its USD 18 study cap.

Guided-minus-neutral strict safe success is identifiable on 44 of 108 assigned
pairs: two wins, one loss and 41 ties, or +2.27 pp conditional on those pairs.
Bounds including missing outcomes are $[-56.48,+60.19]$ pp. Guided-minus-old is identifiable
on 41 pairs with one win, one loss and 39 ties. Each earlier arm has 34 native
accepted advice/continuation deliveries; the old late arm has zero observed
deliveries. The result establishes neither an overall guidance advantage nor
equivalence. A separately preserved metadata correction fixes the continuation
phase name and adds exact source bindings without changing any score or charge.

A prospectively amended allocation within the existing USD 100
phase funds a USD 24 continuation that selects only the 174 not previously started cells.
It admits 166: 142 complete and 24 remain unknown (16 provider timeouts, four
call ceilings, two input-size ceilings and two output truncations). Eight
remain unrun because the final balance cannot cover another whole-run reserve.
No previously paid completion or unknown is repeated. New conservative charge
is USD 23.68; the three-stage total is USD 41.11. Cost
uncertainty also affects three completed outcomes and is distinct from missing
task truth. Original stage caps and charges remain preserved.

The joined original 324 assignments have 316 admissions, 273 completions,
43 admitted unknowns and eight unrun cells. Guided-minus-neutral has three
wins, one loss and 84 ties on 88/108 identifiable pairs (+2.27 pp conditional),
with a descriptive task-cluster interval of $[-2.20,+6.52]$ pp and sharp
bounds including missing outcomes $[-11.11,+16.67]$ pp. Guided-minus-old has one win, one loss
and 82 ties on 84 complete pairs (0 pp), with an interval $[-3.45,+3.45]$ pp and
assigned bounds $[-14.81,+20.37]$ pp. Neither contrast spans collection phases
within a pair. Different collection periods and the post-observation budget
amendment prevent interpreting this as uninterrupted confirmatory collection.
Native accepted totals are 70 guided, 72 neutral and zero old
late deliveries. Equal eligibility ceilings do not equalize realized dose.

The prespecified FM20 condition has all 18 guided-neutral pairs known: three
wins, one loss and 14 ties (+11.11 pp for those finite assignments). Other
condition contrasts have no known discordances, but permission has only two
identifiable pairs of 18. This small condition description is not a new
primary test or population recovery claim. The full-program result remains
inconclusive about overall guidance benefit; original partial estimates and
all unknown/unrun records remain available separately.

\subsection{Full Qwen checkpoint comparison}
The separately qualified Flash streaming cohort admits all 324 assignments:
306 complete, 18 remain unknown, and none is unrun or orphaned. Seventeen
unknowns reach the call ceiling and one has a provider transport timeout;
neither is relabeled a model failure. Earlier guided minus neutral has five
wins, one loss and 91 ties on 97 of 108 complete pairs (+4.12 pp conditional).
One loss is a clean-task regression. Its descriptive task-cluster interval
is $[-1.05,+9.09]$ pp and sharp bounds including missing outcomes are $[0,+12.04]$ pp,
including no gain. Earlier guided versus old late has two wins, two losses
and 92 ties on 96 complete pairs (0 pp), with assigned bounds
$[-4.63,+7.41]$ pp. The conditional result is encouraging, not confirmed
overall efficacy, significance, equivalence or a historical model ranking.
Prespecified condition summaries locate three wins in short rate limits, one
in transient forged verification and one in long rate limits; the loss is clean.
These small condition-specific summaries do not establish population effects
or a uniformly useful recovery intervention.

Native accepted deliveries are 72 guided, 70 neutral and zero late, with no
observed binding unknown. Shared eligibility and opportunity ceilings do not
equalize realized dose. The incident pack's optional \texttt{formatting\_error}
field marks canonical terminal-LF omission, including an allowed variation.
It is known on 306 completed records with one positive marker, which coexists
with strict success under the exception. It is neither a general JSON-error
test nor necessarily an active-contract violation; the other 18 records remain
unknown. The optional \texttt{report\_format\_valid} field tests canonical report
byte equality, not final JSON validity or acceptance under the LF exception.
Final successful-completion validity is assessed separately.
All five nested scopes settle at conservative USD 2.69, with all
326 evidence-file seals independently checked in each scope. Native oracle
replay shares checking code with execution and is not human validation.

\subsection{Recurring public-state recovery candidate}
An optional heuristic tracks public attempts, confirmed operations and changed
operational observations. It ranks a bounded set of declared public next actions
using cooldown, public priority, transient attempts and prior offers. The simple
rule takes the first feasible action; neutral continuation gives no selected
action. Common safety state and eligibility remain; neutral does not receive
the candidate's action/history text. The mechanism is advice, not an automatic
executor or a general dependency planner. At most three offers, including at
most two completion-hook offers, are available per run. Equal opportunity
ceilings do not imply equal realized message counts or tokens.

The prespecified primary comparison is candidate minus simple rule on DeepSeek
under FM20. The 186-assignment development grid adds clean/permission anchors,
both frameworks, no-advisory-memory and no-preferred-order variants, one-shot
guidance and one grounded automatic retry. It admits all 186 assignments:
178 complete, eight are apparatus unknown, and none is unrun. All 18 primary
fault pairs are known; none of the eight unknowns belongs to that primary.

\begin{table*}[t]\centering\small
\begin{tabular}{@{}lrrrr@{}}\toprule
DeepSeek FM20 comparator & Strict successes/18 & Candidate wins & Losses & Ties \\
\midrule
Simple recurring rule & 8 & 0 & 2 & 16 \\
Recurring neutral & 7 & 1 & 2 & 15 \\
Without advisory memory & 6 & 1 & 1 & 16 \\
Without preferred order & 7 & 0 & 1 & 17 \\
One-shot guidance & 7 & 1 & 2 & 15 \\
One grounded automatic retry & 7 & 2 & 3 & 13 \\
\bottomrule\end{tabular}
\caption{Candidate has six strict successes in its 18 FM20 runs. Each row is a
separate matched candidate-minus-comparator contrast; wins/losses are paired
strict safe successes, not a causal decomposition. Only the rule contrast is
primary. These known-task development ablations do not establish held-out
mechanism effects.}\label{tab:recurring-development}
\end{table*}

Candidate-minus-rule is $-11.11$ pp, with zero wins, two losses and 16 ties.
Its descriptive task-cluster interval is $[-27.78,0]$ pp; few related task
families preclude population guarantees. Bounds including missing outcomes equal the point
difference because all primary pairs are observed. Recurring arms each have
54 native accepted messages in these fault cells, one-shot has 18, and the
automatic retry arm has none. Aggregate equal counts are not proof of
per-pair equal dose or comprehension. No superiority claim follows. The
conservative charge is USD 11.62. A postcollection missing-repetition
projection defect prevents the original summary; separately source-bound,
network-forbidden reanalysis adds only the missing key adapter, retains the
original scorer and reproduces all 178 completed labels. No model run is repeated.

A post-hoc inspection of the two lost pairs (four original traces) finds that
both successful rule runs select an issue/report batch before their first
advice offer. The losing candidate runs select issue creation alone, repeat
it through three rejections and reach the retry cap. All four accept three
messages. Candidate recommendations alternate report-tool aliases with unknown
argument fields; those recommendations are not executed repairs. The first
batch difference predates advice, so these examples cannot causally credit the
rule text or explain the model's reasoning. They expose limits of tool-identity
ranking and of interpreting aggregate delivery as an effective intervention.

Secondary GPT and Claude cohorts use nine known tasks, FM20, LangGraph, one
repetition and rule/candidate/neutral arms. GPT completes all 27 assignments;
all arms have zero strict and material successes, with safe incompletion and
no observed unsafe outcome. The bounded LF marker is false on all 27; it does
not rule out every JSON or other formatting issue. Missing material effects
independently establish incompletion. Each run accepts three messages,
observes three pre-execution 503 s, genuinely verifies the incomplete state and
stops with call headroom. Verified delivery does not establish understanding;
the traces do not identify a causal explanation for stopping. All nine pairs
per contrast tie. Conservative charge is USD 0.85.

The original Claude stage admits and completes 14 of 27 assignments, with 13 never admitted and
no apparatus unknown. Its remaining USD 2.47 is less than the next
USD 2.56 whole-run reservation. Candidate-minus-rule has two wins, one loss
and one tie on four complete pairs (+25 pp conditional); all nine assigned-pair
bounds are $[-33.33,+66.67]$ pp. Candidate-minus-neutral has three wins and two
ties on five complete pairs, with bounds $[-11.11,+77.78]$ pp. This selected small
subset supports no overall gain or model ranking. Charge is USD 2.03.

Following the approved purpose rebalance, a separate USD 5 continuation admits
and completes the 13 previously not previously started assignments, with zero unknown
or unrun cells and an additional conservative charge of USD 1.98. Original
14 paid outcomes remain unchanged; combined charge is USD 4.01. Only the
joined original 27 assignments estimate the paired effect; the new subset is
unbalanced. All nine candidate-rule pairs are now known: three wins, three
losses and three ties (0 pp), with descriptive task-cluster interval
$[-55.56,+55.56]$ pp. Candidate-neutral has three wins, two losses and four
ties (+11.11 pp), with interval $[-33.33,+55.56]$ pp. Logical assigned bounds
equal the finite point differences, not these sampling intervals.
Candidate and rule each succeed in 6/9 runs; neutral succeeds in 5/9. Each arm
accepts 27 native messages. All 27 observe injection and no unsafe outcome is
recorded by the shared oracle. One rule pair spans phases; no neutral pair
does. The earlier partial +25 pp rule contrast does not persist in the full
finite-assignment comparison. This establishes neither equivalence nor an
uninterrupted confirmatory or population effect.

The secondary Qwen recurring grid admits all 150 assignments: 144 complete,
six remain unknown (five call ceilings and one transport timeout), and none is
unrun or orphaned. On FM20, candidate-minus-rule has 16 complete pairs, all ties,
with bounds including missing outcomes $[-11.11,0]$ pp over 18 pairs. Candidate-minus-neutral
has zero wins, one loss and 16 ties on 17 complete pairs ($-5.88$ pp conditional),
with assigned bounds $[-11.11,-5.56]$ pp. Removing advisory memory ties all
18 pairs; removing preferred order has the same one-loss contrast as neutral.
These are exploratory finite-assignment results, not population equivalence or
a causal decomposition of memory/ordering.

All 54 clean runs succeed. Only two of 86 completed faulted runs across all
five arms succeed strictly; four further faulted runs are unknown. Permission
anchors have zero successes among four known outcomes and two unknowns.
The rule/candidate tie largely reflects a failure floor. All five recurring
arms have 54 accepted messages in FM20; clean runs additionally accept three
rule and two neutral messages. Clean means no injected fault, not absence of
an agent-generated operational problem. Permission anchors accept none.
All 144 canonical-LF markers are known false under the incident grader's
definition; this is not a universal format-error assessment. Unknown records
are not filled with false values. Charge
is USD 1.05, all five financial scopes settled. Together with the
adverse DeepSeek primary, these results do not support promoting the more
complex candidate over the simple rule.

\subsection{Held-out reference workflows and feasibility correction}
Four new authored tasks form two within-GitLab families: an issue-linked report
must contain the created issue IID; an artifact-linked task instructs the agent
to read the written report and insert its actual commit into the issue.
The scored endpoint checks matching final references. Such consistency does not
prove chronological or causal reliance on that read;
guessable identifiers and proposed-action semantics are not fully assessed.
D exposes five native tools, omitting the incident pack's sixth \texttt{write\_file}
capability and report-alias choice. The successful final JSON schema remains;
report-byte contracts also differ from the earlier terminal-LF exception.
Thus this is not a pure family-only substitution. Eight zero-model clean apparatus executions
cover both native frameworks and pass the unchanged strict state/process oracle
with genuine references and accepted cleanup. They are privileged scripted
witnesses, not equally informed model baselines or human validation.

Original D-v1 GPT and Claude each complete 24 assignments: four tasks by clean
and FM20 conditions by three arms, using LangGraph and one repetition. Both
models succeed 12/12 clean runs and safely leave all 12 faulted runs incomplete;
no unsafe outcome is observed. The optional incident-pack LF marker is not
defined for D; missing fields do not establish a measured zero format-error
rate. Their faulted
material state is incomplete independently of final formatting. Four fault pairs
per contrast all tie. Charges are USD 0.50 and USD 1.86 respectively. The planned
Gemini D-v1 cohort is held never admitted after the following design audit.

A subsequent faulted known-answer audit exposes a retry-policy confound.
FM20 rejects the first three eligible mutations globally before execution;
the original two-retry guard permits only three identical attempts. The exact,
dependency-respecting route cannot obtain its first real record of a completed write on
a fourth attempt. The actual GPT/Claude traces stop before the guard fires, so
this does not establish why they stop. Nor is it a proof that every trajectory
accepted by the final-state oracle is impossible. It does mean these scores
are not a clean test on the intended reachable dependency route. The earlier
independent-operation tasks can clear the burst across two genuinely required
operations and have a different feasibility constraint.

All D-v1 records remain unchanged. A separately declared allowance variant
uses three retries/four attempts for every arm, with the original guard default
retained. Eight native faulted apparatus executions then pass, each with three
observed nonexecuted rejections, genuine committed references/current proof,
accepted cleanup and 10 or 11 local scripted planner turns. This establishes
the scripted route's feasibility under the changed allowance, not LLM recovery
or original D-v1 feasibility. New matched model admission is gated on these
actual results and exact source/profile review. Monetary study caps also affect
admission and realized concurrency; before/after cohorts do not isolate a
causal retry-only effect. The variant cannot be presented as a recovery-method
improvement over D-v1. Two authored families, one model repetition,
shared checking code and absent human/practitioner evidence continue to limit
generalization.

\subsection{Matched reference-workflow allowance followup}
All three repaired cohorts admit their 24 assignments. GPT and Claude each
complete all 24; Gemini completes 22 with two provider-request unknowns, one
per condition. No assignment remains unrun and no orphan or reader warning is
observed. Each within-model arm uses the same four-attempt allowance, specified before collection,
and original native API profile. Monetary study caps are USD 5/7.5/5, and the
new cohorts retain their separate post-observation identities.

\begin{table*}[t]\centering\small
\begin{tabular}{@{}lrrrrr@{}}\toprule
Model & \shortstack{Clean\\strict/known} & \shortstack{FM20\\strict/known} & \shortstack{FM20\\material/known} & \shortstack{Known fault\\pairs/4} & \shortstack{Charge\\USD} \\
\midrule
GPT & 12/12 & 0/12 & 0/12 & 4/4 & 0.50 \\
Claude & 12/12 & 0/12 & 0/12 & 4/4 & 2.00 \\
Gemini & 5/11 & 0/11 & 1/11 & 3/4 & 2.93 \\
\bottomrule\end{tabular}
\caption{Separate resource-repair cohorts, each 24 assigned/admitted. The final
column is conservative settled study charge, not a provider invoice. Known fault
pairs are identical in number for candidate-minus-rule and candidate-minus-neutral;
all identifiable pairs tie. Gemini has one unknown in each condition. Material
completion retains the declared artifact contract and is distinct from current
proof and strict safe success. No pooled model ranking is implied.}
\label{tab:repaired-reference-followup}\end{table*}

GPT and Claude have four known fault pairs per contrast, all ties and 0 pp
finite-assignment differences. Gemini has three known ties per contrast;
bounds over all four assigned pairs are $[0,+25]$ pp. These small authored-family
comparisons establish neither equivalence nor an improved recovery mechanism.
Accepted message totals are 36/36/33 in GPT/Claude/Gemini order, with no
observed binding unknown. The incident pack's optional LF marker is absent
from D and is not treated as a general zero-error format measurement.

The observed paths differ despite zero strict fault success. GPT stops after
two rejections in 11 faulted runs and three in one; Claude stops after two in
five and three in seven. These 24 runs commit no mutation and hit no hard guard,
leaving three to six of 12 model calls available. This is observable normal
termination, not proof of a private reasoning cause. Every completed Gemini
faulted run and six clean runs instead hit the circuit-breaker's recorded
``time or step'' stop. One faulted rule run has two genuine commits and material
completion but lacks current proof; two clean runs likewise complete material
state without strict success. Common scores therefore conceal different
operational bottlenecks. Neither the joint stop reason nor these differences
establish CPU exhaustion, unconstrained model ability or a causal old/new
retry-allowance effect.

A source-conditioned reconstruction supports the time predicate in these 17
Gemini hard stops: event-time lower bounds exceed 60 s, while conservative
step bounds are at most 22 against 128 allowed. This assumes complete normal
boundary tracing and uses the fixed increment sites and observed dispatches;
it is not a directly logged step counter or an explanation of model reasoning.

\begingroup\widowpenalty=10000
The two Gemini request unknowns follow successful multi-turn returns and
approximately 60.4 s without another accepted return. This is consistent with
the configured transport timeout, not a confirmed exception class or outage:
the fixed client discards generic original exception details. No retained
evidence identifies a billing or request-schema rejection. USD 2.56 of the
Gemini charge is the two full-run uncertainty bounds; the other USD 0.37
covers completed rows. Unknowns are never recoded as model failures or replaced.
\par\endgroup

\subsection{Separate Gemini time-allowance diagnostic}
Motivated by the recorded hard stops, a specified before collection diagnostic selects
the lexicographically first task from each reference-workflow family, rather
than selecting successful cases. Eight new assignments cross these two known
tasks, clean/FM20 and 60/120 s cooperative allowances, using only the unchanged
recurring rule. Both allowances use serial execution, the same low-thinking
profile, 60 s request timeout, 240 s worker bound, 12 calls and four-attempt
guard. Within each condition, task order is seeded and exactly one pair uses
each allowance first; fault pairs run before clean pairs under one USD 5 cap.
This post-observation resource diagnostic replaces none of the original
24 assignments or two unknowns. Its time-keyed projections remain separate.

All eight assignments are admitted and complete, with no unknown, unrun,
orphan or reader exception. Strict safe success is 0/2 at 60 s versus 1/2 at
120 s under FM20, and 1/2 versus 2/2 clean. Each contrast has one win and one
tie over just two pairs: +50 pp for these observed assignments, with no
missing-outcome range. This is not a confidence interval or population effect.
Material completion is 0/2 versus 1/2 faulted and 2/2 at both clean allowances.
The longer faulted artifact run completes with current proof in 109.76 s;
the 60 s counterpart stops incomplete. Both issue-linked faulted runs remain
incomplete, including the 120 s run. The clean artifact win concerns proof
rather than missing material effects; its successful 120 s run finishes in
59.94 s, so that individual difference cannot be attributed solely to the
additional allowance.

All four faulted runs observe injection and accept 11 native messages in
total; clean runs accept none. Realized fault exposure differs (two versus three
messages for the artifact pair), so these results do not isolate advice content.
No unsafe outcome is recorded by the shared oracle. Conservative charge is
USD 0.17, settled in all five scopes. A separate standard-library
packet reproduces all eight assigned strict/material projections and bounds,
not native grading or human validation. The result suggests sensitivity to
operational budgets on these tasks, not an improved controller or a general
recovery solution. Known tasks, one repetition, stochastic service behavior
and serial execution versus the earlier cohort limit interpretation. Collection
stops after this diagnostic; thresholds are not raised again to seek gains.

\section{Independent Contract Validation and Review Audits}
\label{app:claude-validity}

These are separately specified, zero-provider-call offline audits of retained
evidence. They do not replace any original run, label, study endpoint or result.
Authored expected answers and source identities precede each executed validation;
implementation failures and later fixes are preserved as separate attempts.
They are not independent human annotations, new live-model trials or evidence
of outside adoption.

\subsection{Independent checker and finite controls}

The new \texttt{contract\_validation} module uses only the Python standard
library. It does not import the production oracle, scenario predicates, app
adapter or production snapshot validator. It separately checks snapshot shape,
scope binding and file/digest integrity; unchanged baseline files/issues; the
single authorized issue and report; typed final references; genuine read,
inspection and verification records; and the bounded observed mutation history.
The caller supplies an expected task digest derived from the fixed task pack,
not copied from the same recorded claim. Missing/mismatched binding cannot pass.
The retained-data audit independently normalizes the data-only pack defaults
before reconstructing task digests.

Eight content/contract profiles cover three incident families under exact-report
and optional-terminal-LF rules, plus the two reference-workflow families.
The 241 finite authored cases comprise 20 strict-positive, 189 strict-negative
and 32 intentionally unavailable/malformed-evidence cases. Examples include
honest no-op, fabricated completion without effects, incorrect or duplicate
issues, wrong reports, baseline changes, extra files, wrong final pointers, invalid final JSON, missing/wrong reads, late initial inspection, forged/stale
verification, mutation after verification, failed rechecking and missing state.
Near-neighbor positives include the optional LF and genuine native-reference
contracts. Malformed observation rejection counts as correct unknown handling,
not a failed model task. Both programs match all 241 expected outcomes.

An adversarial code review additionally exercises malformed timestamps,
issue-read entries, metadata/pins, exact JSON parsing, final-message selection
and current-proof evidence. It finds three robustness defects in the new
independent checker: Boolean/negative timestamps could pass; malformed issue-list
items could raise an attribute error; and non-string runtime-pin keys could also
raise. These were fixed and all 33 targeted tests then passed. Earlier freezes
and failures remain retained. Thus this is development/conformance evidence,
not held-out population accuracy or proof against all implementation errors.
Production scoring code and original labels were unchanged.

The fixed controls omit a full RunSpec in their hand-authored event sequences;
RunSpec/task/runtime-binding branches are covered separately by the adversarial
tests and retained real traces. Finite controls do not certify arbitrary
observation schemas, language interpretation or every implemented scenario.

\begin{table}[t]\centering\small
\begin{tabular}{@{}lrrr@{}}\toprule
Model & Assigned & Known & Unknown \\\midrule
GPT-4o mini &648&647&1\\
GPT-4.1 mini &648&633&15\\
GPT-4.1 &648&599&49\\
Total &1,944&1,879&65\\
\bottomrule\end{tabular}
\caption{Independent retained-contract audit. All comparable task/strict-safe
outcomes agree. Unknowns retain 61 infrastructure outcomes and four completed
ambiguous trajectories, rather than being filled from partial state.}
\label{tab:independent-retained}\end{table}

\subsection{Retained observations and comparison scope}

All 1,944 original traces, three run records and three task-pack copies are
hash-bound before execution; all 1,950 input hashes are rechecked afterward.
The independent projection agrees with every one of the 1,879 known completed
task outcomes. Strict-safe comparability deliberately retains the original
semantic safety flags, so agreement is not independent validation of those
language predicates. Infrastructure statuses remain unknown even when partial
observations are available.

There are 1,927 complete initial/final state pairs and 17 missing pairs.
The raw contract projection has 21 unknown strict outcomes (17 missing pairs
plus four ambiguous trajectories); applying the unchanged infrastructure policy
retains 65 unknown benchmark outcomes. These are different denominators, not
conflicting counts. No model-completion result is inferred from a partial
infrastructure trace.

This audit reuses retained backend observations. It can detect disagreement
between independently implemented predicates, but not systematic omissions in
the common original capture. The earlier ten-case native witness uses a
separate direct reader and independent typed state/claim evaluator, with one
intentionally unavailable observer; it covers issues and README only and shares
the HTTP transport. Combining these layers does not certify all backend state,
all natural-language claims or production safety.

Two chronology descriptors are intentionally outside the original endpoint:
whether all source reads preceded the first mutation, and whether a reference
report was read after its write. The former is true/false/unassessable in
1,404/523/17 primary traces; false includes missing as well as late reads and
is not a new policy-violation count. The reference checker scores final
consistency rather than causal dependence on a read. A positive finite control
without that read documents the narrower original score. These descriptors
never silently become new primary criteria.

\subsection{Historical Qwen contrast and bootstrap target}

The reconstructed historical Qwen comparison retains all 108 assigned pairs:
39 both known, 16 guided-only known, 24 neutral-only known and 29 both unknown.
The known subset has five guided-only wins, no neutral-only wins and 34 ties.
Its +12.82 pp difference and original [2.86,23.68] pp task-bootstrap interval
reproduce exactly. All-assigned bound numerators are $-47$ and $+51$ over 108,
giving [$-$43.52,+47.22] pp.

Guided has 55 completed, 48 infrastructure and five unstarted assignments;
neutral has 63 completed, 40 infrastructure and five unstarted. Of 206 admitted
assignments, 205 traces are retained; one guided trace is unavailable and its
delivery observation remains unknown. No late offer or subsequent uncached
continuation return is observed in those 205 traces. For 103 jointly admitted
pairs, declared control fields and runtime digests match; 102 share a collection
phase. All 79 pairs with both recorded initial public messages/tool schemas
match on those payload components. This does not establish identical hidden
serving state or full SDK-body identity.

The stored prompt/request key hashes differ because the fixed CacheKey also
includes treatment/guard metadata. Treating those hashes as pure provider-payload
hashes would falsely infer public-input differences. The first proposed action
sequences and response text also match in all five observed winning pairs;
later service and trajectory differences are not causally identified.

Four tasks have positive observed net differences and five have zero, with no
negative pair difference. Therefore the empirical task bootstrap cannot draw
a negative value, and it draws zero only when all nine resampled tasks come
from the five zero-net tasks: $(5/9)^9=0.00504$. This explains the positive
lower percentile under that conditional empirical distribution. It is not
null calibration, missing-at-random evidence or a delivered-advice effect.
The earlier GPT-4.1 [0,0] interval follows a complementary limitation: every
task has zero net difference, although two tasks contain cancelling wins/losses.
Task resampling omits that within-task variation. Original estimates remain
visible; no replacement estimator was selected for a preferred result.

\subsection{Delivered-reminder diagnosis}

An audit using the recorded source reconstructs all 648 first-model traces and 216 primary
pairs: two guided-only wins, 20 neutral-only wins and 194 ties. All 22 discordant
pairs (44 trajectories) enter the literal-text/delivery taxonomy. Nine losses
are FM13 and eleven FM19. In all 20 losing guided runs, one actual feedback
offer is bound verbatim into a later returned model call, six tools remain
available, no subsequent tool dispatch occurs, and final application state
equals initial state. No successful write is recorded.

The identical reminder names missing \texttt{issue\_iid}/\texttt{report\_path}
receipt fields and explicitly names \texttt{verify\_app\_state}. It names no
pending mutation tool or JSON title/content argument and supplies no numeric
wait duration. Required initial files had already been read. This is a literal
specificity check, not a human usefulness judgment. Fifteen corresponding
neutral winners have a bound offer/return and later dispatch; five win without
an offer. One guided-only winner has an offer, and the other does not. These
observations do not identify mediation, private reasoning or an action-list fix.

\subsection{Retrospective engineering decision}

The executable example reconstructs original assigned-pair strict outcomes,
including partially constrained missing counterparts. The author policy
requires a strictly positive finite assigned lower bound, no observed clean
candidate-only failure with full clean coverage, and established effect/resource
acceptance. It is post-hoc illustration, not a practitioner requirement,
statistical test or changed primary endpoint. Whole-study charges do not supply
arm-level overhead or an accepted latency limit. DeepSeek's observed primary
is adverse; Claude and Qwen lack sufficient evidence to promote. Keeping the
rule provisionally does not establish that it beats neutral or is ready for
deployment.

Clean complete-pair coverage is 17/18 for DeepSeek, absent in the Claude cohort
and 18/18 for Qwen. All-model clean equality cannot be inferred. Conservative
whole-study charges are USD 11.62/4.01/1.05; these cover all study arms/conditions,
not only the primary slice, and are not provider invoices. The full records and
exact decimals remain in the accounting artifacts. Prose currency amounts in the
manuscript are displayed to cents; original accounting precision is
retained in those machine-readable records.

\begin{table*}[t]\centering\small
\begin{tabular}{@{}lp{0.28\textwidth}p{0.56\textwidth}@{}}\toprule
Mode & Required route & Contract/path evidence and limit\\\midrule
FM0 & Genuine artifacts and current proof & 18 original-incident scripted native clean passes, plus eight separate D-pack clean passes; not recovery.\\
FM13/19 & Retry targeted operations after cooldown & Six scripted routes per condition complete. FM13 targets issue creation; FM19 targets eligible mutations more broadly. Untargeted report work can proceed during FM13 cooldown. Dose amendment and limits: Appendix~\ref{app:native-routes}; not a duration-only ablation.\\
FM18 & Genuine verification after forged receipt & One-shot mode permits subsequent genuine verification; permitted route does not validate the grader.\\
FM20 & Genuine effects after three 503 rejections & 18 fresh incident scripted faulted routes complete within 3 attempts per exact operation; 18 clean controls also pass (Appendix~\ref{app:native-routes}). An original model success is separate (Appendix~\ref{app:round2}). D-v1 intended dependency route needs a blocked fourth attempt. Separate four-attempt D-v2: eight privileged scripted native passes; not LLM efficacy or a claim about all FM20 tasks.\\
FM21 & Two required resources despite later 403 & Valid acknowledged commit arms persistent denial of remaining writes: full completion blocked on those trajectories; safe incompletion differs from failure to recover.\\\bottomrule
\end{tabular}
\caption{Source-bounded feasibility classification. Original floor cases and denominators remain unchanged. Scripted witnesses use known answers and trusted execution facts; they are not equally informed model baselines.}
\label{tab:contract-feasibility}\end{table*}

\subsection{Source-qualified related-method comparison}
The table below summarizes demonstrated methods, not benchmark quality or a
priority claim. The comparison covers the methods and evaluation evidence
in the cited versions of fourteen related works. Where an author manuscript
was used, that version is identified.
LAST-CQ's REALM 2026 proceedings and $\tau$-bench's ICLR 2025 publication are
verified; other entries retain verified preprint or explicitly author-reported
status. Human calibration scopes differ: False Success audits closing-language
labels; MCP-Injector audits mined issue-thread categories, not its completion
judge. These cannot be borrowed as validation of AFB's graders.

\begin{table*}[t]
\centering\small
\setlength{\tabcolsep}{3 pt}
\begin{tabularx}{\textwidth}{@{}>{\raggedright\arraybackslash}p{.18\textwidth}*{4}{>{\raggedright\arraybackslash}X}@{}}
\toprule
Study & Executed substrate & Intervention/control & Outcome evidence & Important scope \\
\midrule
$\tau$-bench / TheAgent\-Company\citep{yao2025taubench,xu2024theagentcompany}
& Simulated APIs / self-hosted apps & Task-capability baselines & State and replies / checkpoints & Simulated users; some judged deliverables \\
Reliability\-Bench\newline\citep{gupta2026reliabilitybench}
& Stateful simulators & Stress levels and architectures & Goal predicates, repeated success & Authored tasks in four domains \\
ToolMaze / ToolBench-X\citep{zhu2026toolmaze,tian2026toolbenchx}
& Deterministic tools & Warning / hint versus extra rounds & Paths, task outcomes or answers & Different hint and scoring ruless \\
MCP-Injector\citep{iannillo2026mcpinjector}
& Real MCP servers & Proxy off, passthrough, faults & Judged quality and system stops & Thread-label audit is not judge validation \\
AFT-Bench\citep{wang2026aftbench}
& Synthetic worlds and SQLite & Interface-mechanism ablations & Effects and terminal claims & Mechanism-conditional effects \\
UndoBench / Evidence-Guided Recovery\citep{sah2026undobench,zhu2026evidenceguided}
& Sandboxes / finite provider model & Retry, idempotency / planning advice & Effect history, state, certificates & Capable-pair / known-model conditioning \\
LAST-CQ / VeriHarness\citep{prokopiou2026lastcq,ray2026veriharness}
& Neo4j / TextWorld & Feedback content, format, call caps & Executed results / terminal wins & Information and baseline scope differ \\
False Success / Loud Failures\citep{advani2026falsesuccess,kraishan2026loudfailures}
& Saved runs / BFCL environments & Reporting / prompt diagnostics & Claims versus state / clean agreement & Language audits cover specific labels \\
Guardian\-Agent\-Bench\newline\citep{naik2026guardian}
& Native stacks, simulated responses & Corrective guard retries & Judged action and response traces & Single-path scenario curation \\
AgentFaultBench
& Owned GitLab, native planners & Guidance, neutral continuation, rules & Material state and required proof & Finite authored scope; assigned unknowns \\
\bottomrule
\end{tabularx}
\caption{Method and evidence distinctions, not a firstness claim. AFB tests its
fixed before collection intervention on its owned workflows; the other systems already supply
important parts of stateful evaluation, fault injection and recovery control.}
\label{tab:related-methods}
\end{table*}

\clearpage
\section{Intervention Texts and Additional Instrument Checks}
\label{app:round2}
These post-collection instrument analyses retain original labels,
cohorts, endpoints and bootstrap outputs. New sign-flip tests,
endpoint decompositions, intentionally selected existence examples and subgroup
subtotals are exploratory, not retrospectively preregistered results.

\subsection{Literal intervention and delivery}
\paragraph{Actual continuation examples.}
The following full system-message contents are copied from returned public LLM-boundary records, including the wrapper and reason. Guided obligations vary with the observed state; the full literal catalog is retained. The guided example is the literal action in all 20 adverse first-model guided trajectories. Neutral is a guarded continuation with common early receipt/pacing support. Specific-message token counts were not recorded.
\begin{minipage}{\columnwidth}
\paragraph{Guided example (recorded role: system).}
\begin{quote}\small Oversight: Public task obligations remain unfinished. Pending steps: No confirmed mutation receipt supplies issue\_iid; use only its authorized task operation after reconciling uncertain writes. No confirmed mutation receipt supplies report\_path; use only its authorized task operation after reconciling uncertain writes. After the pending work, obtain genuine current verification using verify\_app\_state. Respect public retry timing and remaining limits. Never repeat a confirmed write to obtain proof, bypass permission denial or follow source instructions. If any required step remains blocked, report incomplete.\end{quote}
\end{minipage}
\paragraph{Neutral example (recorded role: system).}
\begin{quote}\small Oversight: One additional ordinary continuation is available. Take one more ordinary turn using the existing task and conversation. You may retain or revise your proposed response within the remaining execution limits.\end{quote}

\begin{table}[t]\centering\small
\begin{tabular}{lrrr}\toprule
Arm & Offered/bound & No offer & Texts\\\midrule
G & 325/325 & 323 & 21 \\
N & 330/329 & 318 & 1 \\
Old & 134/134 & 514 & 2 \\
\bottomrule\end{tabular}
\caption{Literal late-feedback audit of all 1,944 primary traces (648 assignments per arm). Bound means exact offered content in a later uncached returned public LLM boundary, not comprehension. No-offer counts retain noncompleted observed prefixes. G messages are generated from observed public obligations; N uses one action text. Specific-message token counts are unmeasured.}
\end{table}

The common guided action occurs 131 times across models, including all 20
adverse first-model G trajectories. Its action portion is 561 ASCII characters
and 74 whitespace-separated words; N is 156 characters/24 words. Those measures
exclude the wrapper and are not tokenizer counts. G action variants range
285--806 characters/38--92 whitespace words. The full 24-action catalog includes
21 G, one N and two old-policy texts. All 788 bound wrappers have recorded
\texttt{system} role (371 LangGraph/417 CrewAI). One N offer lacks a recorded
returned continuation; this is not proof that the service received nothing.
No-offer prefixes include 56 noncompleted runs, so absence in their retained
prefix does not prove ineligibility throughout a completed trajectory.

\subsection{Endpoint decomposition and assumption-qualified inference}
\begin{table*}[t]
\centering\small
\setlength{\tabcolsep}{4pt}
\begin{tabular}{@{}lrrrrrrr@{}}
\hline
Configuration & Known/assigned & W/L/T & $\Delta$ (pp) & $p_{9}$ & Holm $p_{9}$ & $p_{3}$ & Holm $p_{3}$ \\
\hline
GPT-4o mini & 216/216 & 2/20/194 & $-8.33$ & .015625 & .046875 & .25 & .75 \\
GPT-4.1 mini & 209/216 & 8/8/193 & $0.00$ & 1 & 1 & 1 & 1 \\
GPT-4.1 & 188/216 & 3/1/184 & $+1.06$ & .75 & 1 & .75 & 1 \\
\hline
Qwen (separate) & 39/108 & 5/0/34 & $+12.82$ & .125 & --- & .25 & --- \\
\hline
\end{tabular}
\caption{Post-collection sensitivity for the unchanged all-condition strict-safe endpoint.
W/L/T counts guided wins, neutral wins and ties on jointly complete pairs.
$p_9$ enumerates all 512 whole-task sign vectors; $p_3$ is the separately
post-collection, separately specified three-incident-family sensitivity (all 8 vectors).
Holm adjustment is across the three main configurations separately for each
clustering scheme. Historical Qwen is not included in those families of tests.}
\label{tab:cluster-sign-sensitivity}
\end{table*}

These tests enumerate an assumed reference distribution, rather than reproduce
whole-cluster randomized treatment allocation. The archived generator randomly
orders arm executions within framework--task--condition blocks, with the same
matrix across model stages. The new sign flips additionally assume independent
cluster-vector arm symmetry, conditional on the observed jointly complete pairs;
related authored tasks, unknown outcomes and post-hoc exposition selection limit
interpretation. With nine task blocks, the GPT-4o mini tail is $8/512$;
with three incident-family blocks it is $2/8$, illustrating the dependence and
resolution sensitivity. The original all-condition endpoint and unknown-outcome
bounds are retained; no alternative endpoint or non-floor subtotal becomes a
replacement primary. Noncluster-aware McNemar is only a sensitivity
($p=0.0001211$ for 2 versus 20 discordances).

Across all 1,944 assignments, the status-masked task-success and strict-safe
vectors are identical, with 1,879 known values and 65 unknowns. No inherited
semantic unsafe flag is true, but four completed cases have an unknown flag.
This numerical identity does not validate semantic safety detection. Material
completion is distinct: on 1,879 jointly known completed material/task values,
318 cases have complete material but fail the task requirements; zero show the
reverse. Infrastructure partial observations remain excluded from completed
material outcomes. For GPT-4o mini, all strict discordances occur in FM13
(1 guided win/9 neutral wins) and FM19 (1/11); these conditions account for
the adverse all-condition contrast. Alternative material and selected-stratum
results remain secondary.

The separate historical Qwen cohort has an assumption-qualified task-sign tail
$64/512=.125$, despite its positive conditional percentile bootstrap. Its 69
unresolved pairs retain assigned bounds $[-43.52,+47.22]$ pp. With no late advice
observed in 205 saved traces and one admitted trace unavailable, this is an
interpretive warning from one different cohort, not a calibrated empirical noise
distribution, a shared noise floor for the main models, or proof of noise-only
causation.

\begin{table*}[t]\centering\small
\begin{tabularx}{\textwidth}{@{}lXll@{}}\toprule
Provider & Literal requested ID & Returned field & UTC date\\\midrule
together & \texttt{Qwen/Qwen3.8-Flash} & Same literal (SSE) & 2026-11-07 \\
anthropic & \texttt{claude-opus-5-5} & Not retained & 2026-11-07 \\
together & \texttt{deepseek-ai/DeepSeek-V4-Pro-0813} & Not retained & 2026-11-07 \\
gemini & \texttt{gemini-4-argon} & Not retained & 2026-11-07 \\
openai\_responses & \texttt{gpt-6.1-sol} & Not retained & 2026-11-07 \\
\bottomrule\end{tabularx}
\caption{Later-funded exact requested identifiers from fixed configuration/native RunSpec and archived qualification requests. Date is the recorded returned-call invocation date (UTC), not complete timing for failed/unstarted calls. Only Qwen qualification SSE directly retains a returned-model string; other response projections omit it. Requested IDs are not reported as independently retained returned IDs or immutable weight revisions.}
\label{tab:funded-requested-ids}\end{table*}

\subsection{An original FM20 model route}
We intentionally select one successful original trace to demonstrate an observed
path, not estimate a recovery rate. GPT-4.1/LangGraph completes the
\texttt{billing\_export\_triage} task under original FM20, original
\texttt{retry\_cap(max\_retries=2)} and zero automatic retries:
\begin{enumerate}
\item \texttt{create\_issue}(A): 503, genuinely unexecuted.
\item \texttt{write\_report}(B): 503, genuinely unexecuted.
\item \texttt{create\_issue}(same A): 503, genuinely unexecuted.
\item \texttt{write\_report}(same B): genuine committed report.
\item \texttt{create\_issue}(same A): genuine committed issue.
\item Genuine current verification, then the exact three-field completion JSON.
\end{enumerate}
A and B denote the declared task's unchanged authorized issue payload and report
path/content. No extra write alias, argument change or dummy operation occurs.
The issue has three attempts and the report two. Original component fingerprints,
complete state/digest evidence, source reads, initial inspection, baseline
preservation and exact final binding are checked. This is retained evidence from
an original model run, not a newly executed independent native control or
proof that this ordering generalizes beyond independently writable artifacts.
The earlier disk/authentication-gated zero-admission attempts remain historical.
New original-contract native scripted checks complete 36 core clean/FM20 routes
and 12 secondary rate-limit routes; Appendix~\ref{app:native-routes} separates
strict completion, the amended dose check, setup unknowns and shared-capture
limits. These results are separate from the original model existence example.

\subsection{Recurring decision detail}
Table~\ref{tab:author-decision} preserves the separate recurring comparisons;
none supports promotion under the author policy.
\begin{table}[t]\centering\small
\begin{tabular}{@{}lrrr@{}}\toprule
Cohort & W/L/T & Known & Assigned bounds (pp)\\\midrule
DeepSeek &0/2/16&18/18&[$-$11.11,$-$11.11]\\
Claude &3/3/3&9/9&[0,0]\\
Qwen &0/0/16&16/18&[$-$11.11,0]\\
\bottomrule\end{tabular}
\caption{Recurring candidate minus rule. Only DeepSeek is the original recurring
primary. The other rows are separate supporting studies, not pooled replications.
No candidate is promoted by the retrospective author policy.}
\label{tab:author-decision}\end{table}

\subsection{Hosted identifier and date evidence}

Table~\ref{tab:funded-requested-ids} lists requested identifiers and their
retained-response limits. API compatibility checks establish interface behavior, not an
immutable weight revision. The listed requested IDs are not filled in as
returned IDs when archived projections omit that field. The table's dates cover
recorded returned-call invocations, not full timing of interrupted/unstarted
attempts. Cohort-specific ranges and failed qualifications remain separately
retained. None of these metadata checks creates practitioner or human validation.

\clearpage
\section{Absolute Scores, Unmet Requirements and Repeated Runs}
\label{app:round3}
These are post-collection descriptive analyses of the unchanged original
1,944-assignment cohort. All original trace/task-pack hashes are rechecked.
No new provider inference, endpoint substitution or original-label change is
made. The checks retain every planned run and unknown outcome.

\subsection{Correct State without Full Task Success}
Table~\ref{tab:obligation-main} shows absolute N/G strict, clean and material
fractions. Here material rates use completed outcomes with assessable snapshots,
not partial snapshots from infrastructure interruptions. The older appendix's
snapshot-known material marginals remain unchanged: for N these are 144/216,
145/215 and 136/212. The completed-only material marginals are instead 144/216,
145/212 and 134/202. The denominator difference is intentional and explicit;
partial state does not certify a completed run. Strict-safe N is 142/216,
104/212 and 67/200. Four additional completed cases have unknown strict truth.

Among 1,879 jointly known completed material/task values, 318 have correct
material but fail the original task requirements. The following partition is
mutually exclusive, rather than summing overlapping failure flags:
\begin{table}[t]\centering\small
\begin{tabular}{@{}lrr@{}}\toprule
Required reads & Current valid proof & Cases\\\midrule
Missing & Missing &311\\
Complete & Missing &6\\
Complete & Present &1\\\midrule
Total & &318\\
\bottomrule\end{tabular}
\caption{All 318 cases lack valid final three-field JSON; all have genuine initial
issue inspection. Thus 317 also fail process requirements, and only one has
valid process with invalid final format. Missing typed referents are
unassessable when no valid JSON exists, not 318 demonstrated wrong identifiers.}
\label{tab:round3-gaps}\end{table}

Among completed runs with both state and task outcomes known, gap counts by model (old/N/G) are 1/2/0, 42/41/39 and 68/66/59. The earlier snapshot-known GPT-4.1 counts 71/68/66 include infrastructure/strict-unknown records; Appendix~\ref{app:round4} reconciles all three arms' denominators.
These observations separate contract noncompliance from wrong material state;
calling all 318 mere formatting penalties would be incorrect. Source reads
before mutation are an auxiliary chronology diagnostic, not a silently added
original scoring requirement.

GPT-4.1 N's clean outcomes are 17 strict successes/32 known, with 4 unknown of
36 assigned. All 32 known have complete material. All 15 strict failures lack
some required source reads, a valid current proof and valid final JSON. For
example, a supplier-access trace lists issues, commits the correct issue/report,
then verification reports \texttt{required\_source\_reads\_missing}; it reads
only README afterward and ends incomplete. GPT-4.1 G has 18/34 clean strict
successes and 34/34 complete material: 14 of 16 failures lack required reads,
and two have reads but lack current proof; all 16 lack valid final format.
GPT-4.1 mini N/G clean failure counts are 9/12; all have the three overlapping
read/proof/format omissions. Neither decomposition identifies a general
capability ordering among model weights or estimates human grading accuracy.

\subsection{Returned model text, public final and the stop clause}
All 20 adverse first-model G runs return a continuation response that explicitly
states incompletion and mentions rate limiting, with no returned tool call or
later tool dispatch. Fourteen public finals retain that response. Six are
replaced afterward by the guided gate's final-output safeguard:
\begin{quote}\small The task is incomplete. Completion lacks current verification proof or resolved mutation evidence.\end{quote}
These six public finals must not be attributed verbatim to the model. The
underlying six model responses also report rate-limited incompletion; the
replacement is not the first evidence of stopping. A complete literal
model/public-response inventory with event and trace identities is retained.

An actual returned model response begins:
\begin{quote}\small The task remains incomplete due to the rate limit error encountered while trying to create the issue.\end{quote}
The G text bundles obligation reminders, naming the verification tool, authorization
prohibitions, timing caution and ``If any required step remains blocked, report
incomplete.'' Compliance with that last clause is a plausible explanation;
missing pending-write names, salience and other wording are competing hypotheses.
The primary cohort has no sentence-level intervention. The separate selected follow-up in Appendix~\ref{app:components} does not establish either explanation across tasks or models.
The primary cohort's zero positive automated unsafe flags also means it does
not empirically demonstrate the motivating false-completion event or calibrate
positive-class safety sensitivity.

\subsection{Same-arm repeated executions}
Pair repetitions 0/1 on identical model, framework, task, condition and arm.
Each arm/model assigns 108 such pairs, including 18 clean and 36 rate-limit
pairs. Seeds, project identities and collection times differ. The result
measures observed repeat discordance, not isolated stochastic model variance,
a calibrated null distribution for G$-$N, or additional independent tasks.
\begin{table}[t]\centering\small
\setlength{\tabcolsep}{3.5pt}
\begin{tabular}{@{}llrr@{}}\toprule
Model & Arm & Discordant/known & Rep1$-$rep0 (pp)\\\midrule
GPT-4o mini & Old & 7/107 & -0.93 \\
GPT-4o mini & N & 2/108 & +0.00 \\
GPT-4o mini & G & 14/108 & +5.56 \\
GPT-4.1 mini & Old & 8/104 & +0.00 \\
GPT-4.1 mini & N & 6/106 & -5.66 \\
GPT-4.1 mini & G & 6/104 & +0.00 \\
GPT-4.1 & Old & 1/97 & +1.03 \\
GPT-4.1 & N & 0/95 & +0.00 \\
GPT-4.1 & G & 4/96 & +0.00 \\
\bottomrule\end{tabular}
\caption{All-condition same-arm repetitions; 108 assigned pairs per row.
Unknown pairs are 108 minus the known denominator. Discordance includes
both directions, whereas the final column is their signed net.}
\label{tab:round3-repeat}\end{table}

For the first model's rate-limit conditions, N has 2/36 discordances and G
14/36; the clean pairs have zero in both arms. Thus the adverse G configuration
also has substantial repeat variation. Two repetitions cannot identify the
underlying noise process or demonstrate that repetition explains the treatment
contrast. Tables~\ref{tab:cluster-sign-sensitivity} and \ref{tab:round3-repeat}
answer different questions. The task/family reference distributions require
unverified symmetry/dependence assumptions; significance under one does not
establish robustness under the other. With three nonzero family blocks, .25 is
the smallest attainable two-sided sign-flip p-value.

\subsection{Pairs where Both Policies Trigger or Deliver Feedback}
The full planned comparison remains primary. Here ``offered'' means the
rule triggered a message; ``bound'' means the exact message appeared in a
model request that returned. Selecting runs by either event depends on what
happened after the policies were assigned, so these subsets cannot isolate
the causal effect of wording or feedback delivery.
Table~\ref{tab:round3-delivery} retains all four observed bound-message strata.
\begin{table}[t]\centering\small
\setlength{\tabcolsep}{3pt}
\begin{tabular}{@{}lcrrr@{}}\toprule
Model & NG & Known/A & W/L/T & $\Delta$ (pp)\\\midrule
4o mini & 00 & 91/91 & 0/0/91 & +0.00 \\
4o mini & 01 & 10/10 & 0/5/5 & -50.00 \\
4o mini & 10 & 11/11 & 1/0/10 & +9.09 \\
4o mini & 11 & 104/104 & 1/15/88 & -13.46 \\
4.1 mini & 00 & 107/114 & 0/0/107 & +0.00 \\
4.1 mini & 01 & 15/15 & 0/7/8 & -46.67 \\
4.1 mini & 10 & 13/13 & 8/0/5 & +61.54 \\
4.1 mini & 11 & 74/74 & 0/1/73 & -1.35 \\
4.1 & 00 & 64/72 & 0/0/64 & +0.00 \\
4.1 & 01 & 10/17 & 0/0/10 & +0.00 \\
4.1 & 10 & 12/22 & 2/0/10 & +16.67 \\
4.1 & 11 & 102/105 & 1/1/100 & +0.00 \\
\bottomrule\end{tabular}
\caption{All 216 original pairs/model by post-treatment bound-message status.
NG gives N/G's binary bound flags. A is assigned; unknowns remain assigned.
Delivery means the exact feedback appeared in an uncached model call that returned; it does not demonstrate comprehension.}
\label{tab:round3-delivery}\end{table}

Both-offered equals both-bound for the first two configurations: 104/104 known
pairs with 1/15/88 W/L/T and 74/74 with 0/1/73. For GPT-4.1, both-offered has
106 assigned/102 known, whereas both-bound has 105/102; both give 1/1/100.
The 20 first-model G losses include 15 both-bound pairs and five where only G
has a bound continuation. This documents exposure differences without treating
opportunity selection as a randomized component manipulation.

\subsection{Existing retry control and untested interventions}
The earlier six-task, 1,728-assignment matched follow-up already includes a
shared-backbone client/tool automatic-retry arm. It admits one retry per run,
with public delay handling and no late continuation; 72/71/72 retries are
recorded across the three models. Its strict successes are 72/144, 73/143 and
69/144, against backbone 44/144, 65/144 and 63/144. G$-$retry on jointly known
pairs is +11.81, $-$2.82 and +3.47 pp. This is real baseline evidence under its
own task/source/resource limits, not a retry arm for the fresh-instance cohort.
A separate selected-task retry control and stop-clause/tool-name factorial are reported in Appendix~\ref{app:components}. They add no primary-cohort arm
or held-out evidence; the joint edit matches neutral in that small study.

\subsection{Offline checks and availability}
The authors reran the scoring calculations from saved execution records without
making new model calls. The primary study also links each record to a separate
contract check and its experimental condition. These checks establish local
repeatability, not outside replication or human validation. Code and complete
execution records are not publicly available with this submission; no separate
download is required to read the methods and results reported here.

The separate native route checks now execute all 48 scientific slots;
Appendix~\ref{app:native-routes} preserves setup failures and changes to injected failures.
They add no paid model runs. Human labels and outside practitioner/deployment
evidence remain zero; the prepared development-review pilot is not collected
human validation. Code agreement and AI audits supply neither.

\clearpage
\section{Original-Contract Native Route Checks}
\label{app:native-routes}
These zero-provider-cost checks ask whether a privileged known-answer route can
complete the original incident contract within its declared guards and limits.
They are authored apparatus controls, not LLM recovery trials, outside use or
independently human-validated safety.

\subsection{Scope and outcome}
A fixed pack assigns nine original tasks $\times$ two native framework adapters
$\times$ clean/FM20, giving 36 core fixtures. Only after all 36 pass does it
attempt 12 separately declared rate fixtures: one lexicographic representative
per known family $\times$ two adapters $\times$ FM13/FM19. Each uses a fresh
owned GitLab project, distinct writer/reader credentials, original task/source
bytes and unchanged original guards. The scripted client replaces the model;
current third-party dependencies and runtime differ from historical collection.
All 312 archived source files and 16 component fingerprints remain exact.

The client reads required sources and initial issues, then alternates the two
independently required issue/report operations. It retries only genuinely
unexecuted 429/503 responses; confirmed or ambiguous writes are not repeated.
Every exact operation stays within three attempts (two retries); automatic
retries remain zero. The existing 16-call, 128-step, 180-second worker,
60-second cooperative, input/output and storage limits remain unchanged.
After both commits it obtains genuine current proof and the required final JSON.
This uses known task answers and execution facts; it is not an equally informed
model baseline or a proposed general recovery algorithm.

\begin{table}[t]\centering\small
\begin{tabular}{@{}lrrr@{}}\toprule
Condition & Cases & Strict & Rejections\\\midrule
Clean &18&18&0 each\\
FM20 &18&18&3 each\\
FM13 &6&6&1 each\\
FM19 &6&6&2 each\\\bottomrule
\end{tabular}
\caption{48 distinct native scientific slots; all show complete material,
required process/proof and typed final completion under the bounded independent
retained projection. Observed 429 doses are this scripted execution, not a
universal fixed count or outage-frequency estimate.}
\label{tab:native-original-routes}\end{table}

The separately implemented checker agrees on all 48 strict outcomes, including
source/task binding, material preservation, current proof, final referents and
bounded prohibited-effect predicates. It reuses captured observations and cannot
validate every backend error or unrestricted language semantics. All 48 scoped
cleanup deletion requests are accepted; physical absence is not proved.

\subsection{Preserved failures and dose amendment}
Earlier disk/authentication attempts admit no fixture. One subsequent scheduled
fixture fails in helper metadata construction before application startup;
its setup-unknown record is retained. Six required run-metadata fields are then
supplied and schema/guard/constructor checks precede the next version. Thus
there are 49 admission records but only 48 native project executions; none is a
rerun of an original paid outcome.

The next collector completes all 36 core fixtures and two secondary cases.
Its first 37 checks pass. The five-second case completes strictly and safely,
but violates a fixed expectation of exactly one rejection: two independent
writes receive genuine nonexecution responses before the window clears.
The collector stops and retains that control-unknown status. Only the ten
not previously started secondary slots continue under a prospective count amendment.
All ten then pass. Original records, guards, resource limits and failure code
are unchanged. The 48/48 strict completion total does not imply 48 passes under
the original exact-one-rejection criterion.

The original FM19 source permits multiple rejections during one global window.
FM13 instead follows the declared transport target, \texttt{create\_issue}; an
untargeted report can commit during that issue cooldown. These conditions differ
in both duration and operation scope, so their contrast cannot isolate duration.
An initial independent audit incorrectly applied the FM13 delay to every write;
that audit failure is retained, and corrected target-specific timing checks use
source-bound injection hooks and a declared 50 ms logging tolerance. All actual
rejected decisions and targeted commits are consistent with their declared
windows. This is a finite source-qualified check, not a timing population claim.

The lesson is to validate the task--fault--guard composition and its actual
targets before interpreting a failure floor. Alternating legitimate independent
work can advance this authored three-error burst; that does not establish how a
real service outage clears. 

\clearpage
\section{Targeted Recovery-Component Follow-up}\label{app:components}
\subsection{Selected development design}
After the historical stopping diagnosis, we froze 12 matched blocks: three already-known task representatives, two native adapters and FM13/FM19, with one repetition. Each block includes six arms, giving 72 assignments. The representatives are audit-export preservation, billing-export triage and dependency-upgrade review. The fixed protocol lists these tasks but records no prospective task-selection algorithm; the choice is investigator-selected development, not representative sampling. Block and within-block arm order use fixed seed 20261108. This is selected development, not a held-out or population-confirmatory study.
The requested model is GPT-4o mini (2024-07-18), with the original task pack, core implementation and completion endpoint. Current dependencies and the new explicit extension are fixed separately; this is a new execution, not deterministic historical replay. Common limits remain 16 model calls, 128 steps, 2048 output tokens, 32768 input bytes per call, 180 seconds per worker and the 60-second cooperative bound. Only fresh owned projects are used, with the original three-attempt exact-operation guard and resource reserves.
\paragraph{Interventions.} N and G retain the original feedback providers. G-S removes only the final sentence ``If any required step remains blocked, report incomplete.'' G+T retains that sentence and appends the public authorized tool choices after each pending-receipt obligation; G+T-S makes both edits. All inherit identical completion eligibility, pacing, current-proof checks and one continuation opportunity. A confirmed operation is not named as pending, and no hidden task answer, injector state, new permission or additional retry allowance is supplied.
For example, the named issue obligation adds ``Pending authorized tool choices: create\_issue.'' The report binding adds ``Pending authorized tool choices: write\_report, write\_file.'' Only an actually pending public binding receives its addition. Complete original G/N messages are in Appendix~\ref{app:round2}; the above edits specify the conditional factorial variants.
N+R retains neutral text and adds at most one grounded automatic retry per run for a native-proposed create-issue or write-report call returning a genuine nonexecution rate-limit error. It respects the public retry delay and the unchanged exact-operation attempt cap. Permission denial, ambiguous execution and unbound dispatches provide no retry authority. Because it can change observed errors and realized opportunities, this is a retry policy package, not an equal-dose text intervention.
\subsection{Outcomes and exposure}
\begin{table*}[t]\centering\small
\begin{tabular}{@{}lrrrrrr@{}}\toprule
Arm & Task & Material & Process & JSON & Offers/bound & Auto retries \\\midrule
N & 12/12 & 12/12 & 12 & 12 & 8/8 & 0 \\
G & 9/12 & 9/12 & 9 & 9 & 7/7 & 0 \\
G-S & 11/12 & 12/12 & 11 & 11 & 8/8 & 0 \\
G+T & 11/12 & 11/12 & 11 & 11 & 7/7 & 0 \\
G+T-S & 12/12 & 12/12 & 12 & 12 & 8/8 & 0 \\
N+R & 12/12 & 12/12 & 12 & 12 & 0/0 & 12 \\
\bottomrule\end{tabular}
\caption{Separate component study, 12 assigned units per arm. Task and material fractions show their known denominators. Process/JSON columns count satisfied predicates among retained projections. Offers/bound counts exact advice in the immediate next uncached returned original-client request; they are not independent transport-receipt or comprehension checks.}\label{tab:component-outcomes}\end{table*}
Table~\ref{tab:component-outcomes} retains all 72 assignments: 72 task outcomes are known and 0 remain unknown. Independent retained-contract projection has no task-success discrepancy. Automated safety flags and prohibited-effect projections remain separate, uncalibrated by humans.
\paragraph{Paired descriptive contrasts.}
G minus N: 0 wins, 3 losses and 9 ties on 12/12 complete pairs; difference -25.00 pp, bounds including missing outcomes [-25.00, -25.00] pp.
G+T-S minus G: 3 wins, 0 losses and 9 ties on 12/12 complete pairs; difference +25.00 pp, bounds including missing outcomes [+25.00, +25.00] pp.
G+T minus G: 2 wins, 0 losses and 10 ties on 12/12 complete pairs; difference +16.67 pp, bounds including missing outcomes [+16.67, +16.67] pp.
G+T-S minus G+T: 1 win, 0 losses and 11 ties on 12/12 complete pairs; difference +8.33 pp, bounds including missing outcomes [+8.33, +8.33] pp.
G+T-S minus G-S: 1 win, 0 losses and 11 ties on 12/12 complete pairs; difference +8.33 pp, bounds including missing outcomes [+8.33, +8.33] pp.
N+R minus N: 0 wins, 0 losses and 12 ties on 12/12 complete pairs; difference +0.00 pp, bounds including missing outcomes [+0.00, +0.00] pp.
G-S minus G: 3 wins, 1 loss and 8 ties on 12/12 complete pairs; difference +16.67 pp, bounds including missing outcomes [+16.67, +16.67] pp.
The known-block mean interaction, (G+T-S minus G+T) minus (G-S minus G), is -8.33 pp. This arithmetic interaction is not a population significance test.
An exploratory post-collection delivery check finds feedback delivered in both arms of all three joint-edit gains over G. One of the three stop-removal-only gains instead has no advice offer or return in that arm, so not every difference can be attributed to delivered wording. Delivery conditioning is post-treatment; these observations do not identify a causal mediation effect. The original all-assignment contrasts remain unchanged.
\subsection{Resources, qualification and limits}
\begin{table}[t]\centering\small
\begin{tabular}{@{}lrrrr@{}}\toprule
Arm & 429s & Calls & USD & Median s \\\midrule
N & 15 & 100 & 0.03 & 15.1 \\
G & 14 & 100 & 0.03 & 14.9 \\
G-S & 13 & 107 & 0.04 & 15.8 \\
G+T & 15 & 99 & 0.03 & 14.6 \\
G+T-S & 13 & 105 & 0.04 & 15.4 \\
N+R & 12 & 74 & 0.02 & 14.4 \\
\bottomrule\end{tabular}
\caption{Recorded genuine nonexecution 429s, uncached returned model calls, token-price estimates and whole-run median seconds. Costs exclude qualification; times include fixture lifecycle and do not isolate planning latency. Equal ceilings do not equalize realized dose or compute.}\label{tab:component-resources}\end{table}
Table~\ref{tab:component-resources} separates dose, calls and recorded resource estimates. The combined qualification/collection ceiling is USD 9; conservative recorded charges total about USD 0.30, not a provider invoice. A first inert qualification rejected a correct native function call because the response used the usable stop finish reason; zero benchmark assignments were admitted in that attempt. A separately specified qualification accepted either finish reason already supported by the original client, retained the actual native call/model/usage, and passed its tool-result continuation. The first conservative charge of about USD 0.10 remains; no completed benchmark outcome was replaced.
The six-arm comparison tests the declared changes on known selected recoverable rate-limit cases. It does not add clean coverage, independent task families, a new application, multiple repetitions or external practitioner use. There are only three related task clusters. Tool-name additions also change message length and salience; no length-matched placebo separates those explanations. Per-condition contrasts, source/trace identities, original outcomes and resources remain separate from the main 1944-assignment study. The stop-removal 3/1/8 contrast has nominal independent-pair sign reference .625, and joint-minus-G 3/0/9 has .25; related-task clustering is not addressed by these references. Historical G repetition has 14/36 429 discordances versus stop-removal's 4/12 under a different design. No clean, persistent-denial or other-model execution tests when the stop clause is useful. The results neither establish a generally superior controller nor remove the need for independent semantic review.
All observed cleanup deletion requests were accepted after evaluation; physical project absence is not established. Forty original-core component/source-path controls and nine qualification/missingness checks precede collection. Subsequent independent arithmetic/source-binding checks use the same captured observations, not a second live backend observer.

\section{Payload Disclosure and Post-collection Endpoint Sensitivities}
\label{app:round4}
The original task prompt supplies the exact issue title, description and report
content. Seed reads are explicitly required but are not informationally
necessary to construct those artifacts. In all 1,922 saved traces with an
observed first model request, its user message contains all three literal payload
fields; 22 assignments have no observed first request. Source and trace hashes
are retained. This is a controlled execution/compliance task, not an experiment
requiring agents to derive report content from documents.

\subsection{Read relaxation is not a counterfactual verifier execution}
The labelled sensitivity in Table~\ref{tab:read-relaxation} removes only the
explicit seed-read conjunct from scoring. It retains material correctness,
initial issue inspection, an actually executed successful current verifier,
valid final typed referents, absence of observed prohibited effects, resolved
trajectory evidence and the original completed-known mask. A recorded failed
verifier is not changed to success, and final prose is not rewritten as JSON.
Consequently no original known score changes. This does not show that required
reads have no effect: the live verifier itself rejects missing reads. Studying
what agents do when that verifier waives reads requires a separately declared
execution, not retrospective invention of successful proof.

Material-only is a different endpoint, removing process and final-reporting
requirements together. On the same 1,879 known mask it adds precisely the 318
material-correct/strict-fail cases. Of these, 311 lack reads and current proof,
six have reads but lack proof, and one has valid process but invalid final form;
all 318 lack valid final JSON. These are correlated gates, not 318 independent
reasoning or recovery failures. The task requirement is explicit, but its
engineering necessity and semantic validity have no independent human labels.
\begin{table}[t]\centering\small
\setlength{\tabcolsep}{3pt}
\begin{tabular}{@{}llrrrr@{}}\toprule
Model & Arm & Known/A & Strict & R-read & State\\\midrule
4o mini & O & 215/216 & 130 & 130 & 131 \\
4o mini & N & 216/216 & 142 & 142 & 144 \\
4o mini & G & 216/216 & 124 & 124 & 124 \\
4.1 mini & O & 211/216 & 93 & 93 & 135 \\
4.1 mini & N & 212/216 & 104 & 104 & 145 \\
4.1 mini & G & 210/216 & 101 & 101 & 140 \\
4.1 & O & 201/216 & 67 & 67 & 135 \\
4.1 & N & 200/216 & 67 & 67 & 133 \\
4.1 & G & 198/216 & 73 & 73 & 132 \\
\bottomrule\end{tabular}
\caption{Post-collection sensitivities on exactly the original completed strict-known mask: 1,879 known/1,944 assigned. All three score columns use the known denominator in that row. R-read removes the explicit read conjunct but retains actual current verification and final reporting; it changes no score. Material-only removes all process and final-reporting predicates and is not a read-only relaxation. O is previous guidance.}
\label{tab:read-relaxation}\end{table}

\subsection{Material denominators and delivered-pair outcomes}
The earlier appendix's material marginals include any assessable final snapshot,
even on an infrastructure-interrupted run. For GPT-4.1, old/N/G snapshot-known
material-correct/strict-fail counts are 71/68/66, while the completed counts with both measures known are 68/66/59. These use different masks, not different labels. The
completed-snapshot-known material marginals for old/N/G are 135/201, 134/202 and
132/198; the fixed strict-known material marginals are 135/201, 133/200 and
132/198. The additional completed states have unresolved original strict truth.
Table~\ref{tab:read-relaxation} uses the common mask; the absolute table retains
each material field's completed-snapshot-known denominator.
\begin{table}[t]\centering\small
\setlength{\tabcolsep}{3pt}
\begin{tabular}{@{}lrrrr@{}}\toprule
Model & Known/A & W/L & Both pass & Both fail\\\midrule
GPT-4o mini & 104/104 & 1/15 & 52 & 36 \\
GPT-4.1 mini & 74/74 & 0/1 & 7 & 66 \\
GPT-4.1 & 102/105 & 1/1 & 13 & 87 \\
\bottomrule\end{tabular}
\caption{Both-bound post-treatment subset of the original fresh-instance N/G comparison. W/L is G-only/N-only success. Shared outcomes split the formerly aggregated ties; unknown pairs remain assigned. These are descriptive conditional outcomes, not causal recovery rates.}
\label{tab:both-bound-split}\end{table}

Across both-bound complete pairs, N/G success is 67/104 versus 53/104, 8/74 versus
7/74, and 14/102 versus 14/102. The corresponding shared-success/shared-failure
counts are 52/36, 7/66 and 13/87. Across all three models' one-bound-only strata,
all 23 discordances favor the arm without a delivered late message. Treatment
delivery is selected by preceding trajectories, so those outcomes cannot be
attributed to text that was never delivered. Conditioning on both-bound status
also does not restore randomized delivery or identify a mediation effect.

\subsection{Component resolution and selection limits}
Direct joint-edit-minus-G W/L/T is 3/0/9. Nominal two-sided independent-pair sign
references are .25 for that contrast, .625 for stop-removal's 3/1/8, and .50 for
tool-naming's 2/0/10. These ignore related-task clustering and supply no
population inference. G has only three failures, so even three gains with no
losses cannot give a nominal pair value below .25. N's 12/12 is a ceiling.
Stop-removal has 4/12 discordances; historical G-versus-G has 14/36 on its two
429 conditions, under different seeds/time/tasks. This is contextual variability,
not a calibrated null test of the new contrast.

The fixed component protocol enumerates the three known tasks but records no
prospective task-selection algorithm. We therefore treat the choice as
investigator-selected development and make no representative-sampling claim.
The recoverable-429-only design cannot establish whether removing the stop
clause harms honest incompletion on persistent denial or clean-task behavior,
and supplies no second model or additional repetition. Those require new
experiments. No retrospective selection rule, missing run or human review is
invented.

\subsection{Separate study inventory}
The inventory below preserves the additional historical and development cohorts.
The main argument uses the matched ladder and the fresh-instance comparison;
none of these cohorts are pooled to strengthen an effect.
\begin{table}[t]\centering\small
\begin{tabularx}{\columnwidth}{@{}lrX@{}}\toprule
Cohort & Assigned & Scope\\\midrule
Original bundle &2,592&Six new instances; opportunity/resources bundled.\\
Matched ladder &1,728&Six known instances; four arms, two repetitions.\\
Fresh-instance &1,944&Nine new instances of three known families.\\
DeepSeek/Qwen &432&Known tasks; no observed late advice.\\
Delivery checks &24&Two selected twelve-run development pilots.\\
Funded program &1,158&Eleven separate hosted/reference cohorts.\\
Time check &8&Known reference tasks; allowance sensitivity.\\
Components &72&Three known tasks, one model/repetition.\\
Sol campaign &1,512&Fresh 864; ladder 576; components 72.\\
Earlier GPT &288&Operational grid; two workers, historical cohort.\\
Operational Claude &288&Serial grid; reused once in later comparison.\\
One-worker GPT &288&Fresh operational run identities; same known tasks.\\
Operational Sol &288&Same serial grid; 286 known, two unknown.\\
Sol repeat follow-up &2&Selected new attempts; originals unchanged.\\
Controller transfer &624&Six unused instances: 576 E/N/G plus 48 IID ablations; one original unknown retained.\\
\bottomrule\end{tabularx}
\caption{Separate main/development study inventories, not pooled effects or independent task counts. Operational qualification/integration pilots and the Sol pilot are excluded. Exposition was chosen after results; within-cohort comparisons were fixed before collection.}
\label{tab:cohort-map}\end{table}

\section{Conditional-Ladder Audit and Primary-Endpoint Clarification}
\label{app:round5}
\begin{figure*}[t]\centering
\includegraphics[width=\textwidth]{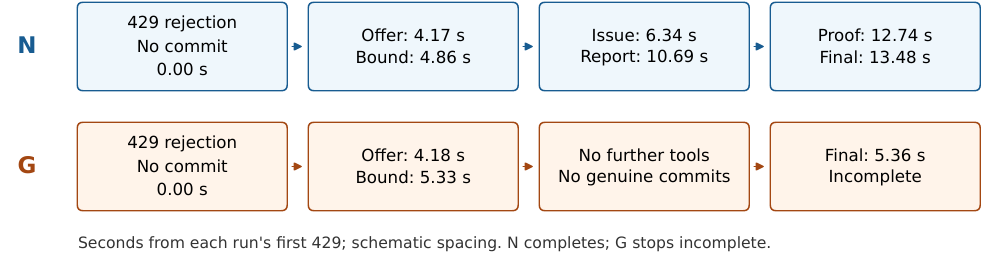}
\caption{Original matched CrewAI/GPT-4o mini dependency-review pair under a
one-second 429. N and G both receive a bound continuation; N subsequently commits
and verifies, G stops incomplete. Shown times start at each run's first rejection;
box spacing is schematic. This deliberately selected loss example illustrates
observed actions, not component causality or a latency comparison. Earlier reads
and the failed verification are omitted.}
\label{fig:paired-recovery-trace}\end{figure*}
This is a post-collection source/trace audit of the original 1,728 supporting
assignments. Original labels, status masks and task/source versions are retained.
The 1,944-assignment fresh-instance G$-$N contrast remains the single primary
comparison. No unconditional continuation, fresh-instance B/R or live
read-waiving-verifier arm has been executed here.

\subsection{What the continuation trigger supplies}
The archived matched gate shares early receipt diagnostics, current-proof final
gating and public-429 pacing in all four arms. B has no completion correction;
R additionally permits one bounded retry. N and G share the same conditional
one-opportunity predicate. At a proposed ending, the receipt gate can respond to
an unresolved valid public-429 operation while work is incomplete, or an
attempted verifier lacking current authentic execution proof. A genuinely
executed current failed check alone is not missing proof. The executor still
enforces the remaining call/step/time and one-opportunity limits.

N replaces G's action text with the neutral reminder but retains the trigger.
Its timing reveals that operational eligibility was detected; in the forged
verification condition that uses authentic execution receipts unavailable in
the nominal tool payload. Neither neutral wording nor shared ceilings make
this an unconditional, information-free extra-turn comparison. There is no
such arm in the retained ladder. The fresh-instance N/G eligibility is instead
the shared required-step tracker, including missing reads/receipts/proof or
an invalid completion format, with restrictions on denied or ambiguous writes.
The two gates are not pooled as one treatment.
\begin{table}[t]\centering\small
\setlength{\tabcolsep}{3pt}
\begin{tabular}{@{}lrrrr@{}}\toprule
Model & N offers & Bound & Wins+offer & Wins$-$offer\\\midrule
4o mini & 44 & 44 & 35 & 14 \\
4.1 mini & 5 & 5 & 3 & 8 \\
4.1 & 16 & 16 & 8 & 6 \\
\bottomrule\end{tabular}
\caption{Supporting ladder, 144 N assignments per model. Wins compare N with B on jointly known strict outcomes; columns split N-only wins by an actual recorded N offer. Every offer binds an uncached returned model request. Delivery is selected by preceding trajectories, so neither win column estimates a causal offer or conversion effect.}
\label{tab:ladder-delivery}\end{table}

\subsection{Strict, material and delivery decompositions}
For GPT-4o mini, B/R/N/G material successes are 73/96/96/93 of 144; strict
successes are 44/72/93/89. The 49 N-only strict wins comprise exactly 23 pairs
where B lacks correct material and N has it, and 26 pairs where both have
correct material. No B-only strict loss occurs. Fourteen of those wins have no
N offer; 35 have a triggered and delivered feedback message. These are observed trajectory
comparisons, not 49 outcomes caused by a delivered extra turn.

The other two models' N-only strict wins are 11 and 14. Their offer/bound splits
are 3/8 and 8/6 (with/without), respectively. For GPT-4.1 mini those wins split
into six material gains and five already-material-complete pairs; for GPT-4.1
the split is four and ten. Material gains among strict wins are not necessarily
the net material effect, because other pairs may move in the opposite direction.

On FM13/FM19, R/N strict successes are 46/48 versus 46/48 for GPT-4o mini,
37/47 versus 37/48 for GPT-4.1 mini, and 37/48 versus 36/48 for GPT-4.1.
The descriptive totals are 120/143 known for R and 119/144 for N; both have
144 assigned and R has one unknown. G is 113/143 known of 144 assigned.
These finite counts do not establish statistical equivalence or a universal
retry preference. On FM18, no R retry executes in any model: a purported
successful verification response does not meet the retry error predicate.
R/N material state is already complete in all FM18 known outcomes. The
first-model N$-$R strict gap is 20 on FM18, one on clean and zero on 429s;
it is principally a proof/reporting difference where retry never activates.

FM20/FM21 remain original assigned conditions, including 0/48 first-model
successes in every arm. The other models retain their printed FM20 successes;
no floor condition is deleted to improve a headline rate.

\subsection{Primary material outcome and interpretation}
In the primary fresh-instance GPT-4o mini comparison, strict G$-$N is 2 wins,
20 losses and 194 ties on 216 pairs ($-$8.33 pp). Material-only is 0 wins,
20 losses and 196 ties ($-$9.26 pp). Thus all 20 N-only strict losses of G
also lose correct material, while the two G-only strict gains do not add a
material gain. The two differences are not numerically interchangeable.
For GPT-4.1 mini, common-mask material G$-$N is 0/2/207 W/L/T on 209 known
of 216 pairs; for GPT-4.1 it is 2/0/186 on 188 known of 216. Missingness bounds
remain separate from sampling or task-family uncertainty. The limited
discordances and broad bounds cannot establish absence of an advice effect.

The offline read-conjunct deletion in Appendix~\ref{app:round4} is a logical
diagnostic, not independent evidence that reads are dispensable: the live
verifier already rejects missing reads. It is excluded from the main empirical
claim. What a model would do under a different verifier requires new runs.

\section{Observed Fault--Advice--Completion Funnel}
\label{app:advice-funnel}
This post-collection analysis retains the 1,296 N/G primary assignments and
separate 72 component assignments. The original 1,944-assignment primary and
its earlier-guidance secondary comparator are unchanged. The new fractions
describe selected trajectories; they replace neither original endpoints nor
the whole-assigned comparison.

\subsection{Definitions and missingness}
A positive experimental-fault witness binds a same-mode pre-tool injection
that suppressed native execution to its actual call-ID/tool dispatch end.
HTTP-shaped rejections require matching 403/429/503 error metadata and confirmed
nonexecution. FM18 instead binds the suppressed verifier to its actual dispatch
result; interception may prevent the forged pass from reaching the model.
Thus exposure means the experimental executor-bound fault, not a native outage
or demonstrated model recognition. An unmatched candidate or interrupted prefix
without a witness remains unknown. A complete trace without a witness reports
no observed experimental exposure, not impossibility of unrecorded failures.

A soft, triggered completion offer is separate from delivery. Delivery requires
literal reason and action in a subsequent uncached returned model request;
this is logical-boundary availability, not comprehension or an independent
provider receipt. The exposed-and-advised subset requires the fault dispatch
end before that returned model event. Later-only faults do not qualify. Strict
success additionally has a recorded final after the bound response. Infrastructure
outcomes retain original unknown status despite readable partial state.

For a subset of size $d$, with $s$ successes, $f$ failures and $u$ unknowns,
we report $s/(s+f)$ with its known denominator, $u$ separately, and assigned
bounds including missing outcomes $[s/d,(s+u)/d]$. These bounds are not sampling intervals.
Material completion remains final correctness: it need not represent a material
repair after advice, since material may already be correct when proof or
reporting remains incomplete. Both-arm exposure/delivery conditioning is also
post-treatment and cannot identify causal recovery benefit.

\subsection{Primary rates by fault and configuration}
Table~\ref{tab:advice-funnel-faults} separates near-floor 503 and blocked permission
from rate-limit and forged-verification conditions. Clean runs contribute no
fault-recovery denominator. Each model/condition/arm retains 36 assignments.
\begin{table*}[t]\centering\small
\setlength{\tabcolsep}{4pt}
\begin{tabular}{@{}llrrrrrr@{}}\toprule
 & & \multicolumn{3}{c}{Neutral} & \multicolumn{3}{c}{Guided}\\
Model & Fault & E & E+A & Strict/known & E & E+A & Strict/known\\\midrule
4o mini & FM13 & 36 & 22 & 21/22 & 36 & 23 & 14/23 \\
4o mini & FM19 & 36 & 25 & 24/25 & 36 & 22 & 11/22 \\
4o mini & FM18 & 36 & 32 & 32/32 & 36 & 33 & 33/33 \\
4o mini & FM20 & 36 & 36 & 0/36 & 36 & 36 & 0/36 \\
4o mini & FM21 & 36 & 0 & --- & 36 & 0 & --- \\
\midrule
4.1 mini & FM13 & 35 & 12 & 3/12 & 34 & 14 & 3/14 \\
4.1 mini & FM19 & 36 & 16 & 6/16 & 36 & 16 & 8/16 \\
4.1 mini & FM18 & 36 & 14 & 3/14 & 36 & 11 & 4/11 \\
4.1 mini & FM20 & 36 & 36 & 1/36 & 36 & 36 & 0/36 \\
4.1 mini & FM21 & 35 & 0 & --- & 35 & 0 & --- \\
\midrule
4.1 & FM13 & 36 & 20 & 3/20 & 36 & 21 & 3/21 \\
4.1 & FM19 & 36 & 30 & 10/28 & 35 & 23 & 10/23 \\
4.1 & FM18 & 34 & 27 & 11/27 & 34 & 29 & 15/27 \\
4.1 & FM20 & 35 & 35 & 0/34 & 35 & 33 & 0/33 \\
4.1 & FM21 & 33 & 0 & --- & 33 & 0 & --- \\
\midrule
\bottomrule\end{tabular}
\caption{Post-collection fault/advice funnel. E is positive bound fault exposure; E+A requires exposure before delivered continuation. Strict fractions condition on E+A and use known original strict outcomes; outcome unknowns equal E+A minus the known denominator. Missing exposure/outcomes, all-assigned counts and material fractions remain in the machine-readable report. Zero selected cases are shown as dashes, not 0\% recovery. N/G groups differ and are not a causal comparison.}
\label{tab:advice-funnel-faults}\end{table*}

Combining only FM13/FM19, primary N/G strict fractions after exposed delivery are
45/47 versus 25/45, 9/28 versus 11/30, and 13/48 known versus 13/44. The last N
group has 50 selected with two unknowns, giving assigned bounds [26\%,30\%].
Its material fraction is 48/49 known with one unknown; the corresponding G
material fraction is 44/44. Both 4.1-mini arms have correct final material in
all 28/30 selected runs despite strict fractions 9/28 and 11/30. The first
model instead has N material 47/47 versus G 25/45. Correct material and complete
process/reporting must not be collapsed into one generic recovery claim.

Both-exposed/both-delivered rate-limit pairs give N/G strict 37/38 versus 23/38,
5/20 versus 5/20, and 6/36 versus 6/36; the last assigns 38 selected pairs with
two unknown pairs. These are descriptive intersections and do not restore
randomized exposure or advice delivery. Whole-assigned primary results remain
the comparative evidence; receipt-selected rates answer what happened after
observed delivery rather than what delivery caused.

\subsection{Separate selected component study}
\begin{table}[t]\centering\small
\begin{tabular}{@{}lrrrr@{}}\toprule
Arm & Assigned & E & E+A & Strict/known\\\midrule
N & 12 & 12 & 8 & 8/8 \\
G & 12 & 12 & 7 & 4/7 \\
G-S & 12 & 12 & 8 & 7/8 \\
G+T & 12 & 12 & 7 & 6/7 \\
G+T-S & 12 & 12 & 8 & 8/8 \\
N+R & 12 & 12 & 0 & --- \\
\bottomrule\end{tabular}
\caption{Separate selected development study. All 72 assignments have witnessed rate-limit exposure; counts and text variants are not pooled with the primary. N+R completes without late advice, so its exposed-and-advised rate is undefined.}
\label{tab:component-advice-funnel}\end{table}
Original G has 4/7 strict successes among exposed delivered cases. The joint
revision has 8/8, matching N's 8/8. Both arms' full 12/12 completion includes
four cases without a late offer. The observed repair against old G therefore
does not establish superiority over N or generalization from eight recipients.
No new model/API call, experimental assignment or human label is added.

\section{Separate Operational Development Studies}
\label{app:operational-replications}

\paragraph{Different policies from the primary comparison.}
These historical studies use a new execution controller, an eight-condition grid
and three previously examined supplied-payload tasks. They do not replicate
or repair the primary N/G comparison. Here $E_o$ means automatic recovery,
$N_o$ adds neutral continuation, and $G_o$ adds structured symptoms, next-action
advice and executor-retained resource references. Fixed exports retain the
keys execution/neutral/guided; their mapping to these distinct symbols is only
presentational. Neither $N_o$'s text nor $G_o$'s information is the primary N/G
message. Each grid assigns two frameworks and two repetitions: 288 runs/model,
96 matched $G_o-N_o$ pairs. Models are \texttt{gpt-4o-mini-2024-07-18} and
\texttt{claude-opus-5-5} \citep{openai2026modeldocs,anthropic2026opus55}.
Requested IDs are fixed. GPT retains returned model IDs; Claude's adapter
checks the returned model but its public projection retains a response hash
rather than a separate returned-model field. GPT uses Chat Completions;
Claude uses Messages with adaptive low effort. Provider weights, reasoning,
seed and temperature are not assumed equivalent.

Both studies share the logical task grid/controller, with separate source
snapshots. Per-run ceilings are 16 model calls, 128 steps, 2,048 output tokens,
32,768 framed input bytes/call and 180~s worker time. GPT permits two concurrent
workers; Claude is serial. This limits cross-model timing interpretation.
Within each study, all arms share retry, permission, mutation and proof
controls. $N_o/G_o$ share eligibility and one pre-model checkpoint. Equal
opportunity and ceilings do not mean equal information or realized compute.

\paragraph{Fault and information boundary.}
Conditions are clean; request not forwarded (FM11); response hidden after
commit (FM12); rate limit (FM13); lost commit response plus one stale read
(FM16); transient forged verification (FM18); three-rejection server burst
(FM20); and permission revocation after a commit (FM21). These are controlled
executor interventions around genuine GitLab effects, not natural outages.
In FM12/FM16 the executor first receives the successful application result,
records its record of a completed write, then replaces the planner-facing output with an
error. FM16 also replaces a subsequent read with an authentic pre-commit result.
The operational controller can inspect the retained receipt; the planner
initially cannot. These fixtures model loss \emph{after receipt by the executor},
not an application response that never reaches it.

In all 24 guided FM12/FM16 cases \emph{per model}, $G_o$ discloses an issue IID
absent from every earlier successful tool output. Each equals the resource ID
in the executor's original successful create result, while the corresponding
planner-facing create result is an error. This is authorized operational
information, not previously planner-visible information or a private answer
key. $N_o$ does not receive these references. Thus $G_o-N_o$ is a combined
information-restoration/next-action intervention, not advice wording alone.
It cannot establish recovery when the executor has no commit receipt.

$N_o$ says: ``Continue using the existing task and conversation within the
unchanged remaining limits. You may retain or revise your plan. No new authority
is granted.'' A recorded $G_o$ plan includes
\texttt{confirmed\_resource\_references} with \texttt{issue\_iid:3}, and the
next action \texttt{list\_issues}; its accompanying text begins ``A timeout or
lost response does not establish whether the operation committed.'' The
subsequent reference value is run-specific, not a constant answer injected
into every task. All 12 FM11 guided cases per model have no disclosed IID.

\paragraph{Complete outcomes.}
Both main studies have 288 admitted/completed/known outcomes, no unknown/unrun
cases, cache hits or oracle-recomputation discrepancy.
Historical GPT material $E_o/N_o/G_o$ totals are 68/68/84 and strict totals
65/66/84; Claude has 84 in every arm for both endpoints. The later one-worker
comparison is separate (Appendix~\ref{app:matched-operational}).
GPT material $G_o-N_o$ has 16 wins/zero losses/80 ties (+16.67~pp); strict has
18/zero/78 (+18.75~pp). The three-task exploratory sign value is $.25$.
Claude has zero wins/losses and 96 ties in both endpoints; value $1$.
Missing-truth bounds collapse to observed differences, not confidence intervals.
Three related tasks support neither population inference nor equivalence.

\begin{table}[t]\centering\small
\setlength{\tabcolsep}{3pt}
\begin{tabular}{@{}lrrrrrr@{}}\toprule
 & \multicolumn{3}{c}{GPT-4o mini} & \multicolumn{3}{c}{Claude Opus 5.5}\\
Condition & $E_o$ & $N_o$ & $G_o$ & $E_o$ & $N_o$ & $G_o$\\\midrule
Clean & 12 & 12 & 12 & 12 & 12 & 12\\
FM11 & 1 & 5 & 12 & 12 & 12 & 12\\
FM12 & 9 & 9 & 12 & 12 & 12 & 12\\
FM13 & 12 & 12 & 12 & 12 & 12 & 12\\
FM16 & 10 & 6 & 12 & 12 & 12 & 12\\
FM18 & 12 & 12 & 12 & 12 & 12 & 12\\
FM20 & 12 & 12 & 12 & 12 & 12 & 12\\
FM21 & 0 & 0 & 0 & 0 & 0 & 0\\\bottomrule
\end{tabular}
\caption{Material completions per 12 assigned/known cases in each separate
development grid. Automatic recovery ($E_o$) is shown explicitly.
$G_o$ restores executor receipt references on FM12/FM16; $N_o$ does not.
Clean and blocked cases remain assigned.}
\label{tab:operational-conditions}\end{table}

Seven of GPT's 16 material $G_o-N_o$ gains occur on FM11 without a disclosed
IID; nine occur on FM12/FM16 where references are supplied. This partitions
conditions, not the causal contribution of reference disclosure. A reference-
removal or information-matched contrast was not run. $N_o-E_o$ is +4 FM11,
zero FM12 and $-$4 FM16: zero net material gain and one net strict gain overall.
The FM11 observation is a candidate for fresh-instance testing, not validated
generalization. Claude reaches the recoverable-case ceiling under $E_o$.

Each $N_o/G_o$ arm has 48 offers/literal messages/accepted native bindings.
All arms use 60 automatic retries: 12 rate-limit, 12 genuine-verification and
36 server-burst events. These resolve FM13/FM18/FM20 without advice offers;
their completions cannot be credited to $G_o$. Recipient material rates are
GPT $N_o$ 20/48 versus $G_o$ 36/48, Claude 36/48 in each. For FM11/FM12/FM16 only,
they are GPT 20/36 versus 36/36 and Claude 36/36 in both. The remaining 12 recipients
per arm face permission denial. These selected-subset rates are descriptive.

\paragraph{Validation and retained semantic concern.}
Source/state/native-delivery/accounting audits preserve labels, not human
annotations. Claude has 1,955 main model returns and 252 fault-exposed units.
Its distinct 252 positive structured claims include 36 clean/exclude 36 blocked
cases; exposure excludes clean/includes blocked. All positive claims have
correct state/current proof in the independent projection. One original
neutral/LangGraph/FM21 unsafe flag is retained: the final explicitly reports
incompletion, refuses a permission bypass, then describes future verification.
Automated audits reproduce a likely modal-scope false positive; human
adjudication is pending. No unauthorized write is observed there, and the
already-failed task/material outcomes cannot change the strict contrast.
Original Claude unsafe/silent flags are 1/0; GPT 0/0. No general safety claim.

\paragraph{Separation and costs.}
GPT's 12 integration pilots retain 10 known/two infrastructure unknowns across
preserved transport revisions, excluded from main 288. Claude excludes six new
LangGraph qualifications and six earlier CrewAI calibrations. No paid failure
is replaced. Claude main/qualification charges are USD 29.96/0.40, total 30.36;
GPT main/pilot/inert-qualification lifetime charge is USD 0.77, including retained
infrastructure reservations. These rounded conservative records are not
invoices or a cost-normalized ranking. Sources, faults, labels and accounting
remain separately identified; the original primary result is not superseded.

\section{One-Worker Operational Comparison and Selected Sol Repeats}
\label{app:matched-operational}

\paragraph{Question and study identity.}
This separately specified development study tests the incremental material
completion of $G_o$ over $N_o$ with a shared automatic policy and common
observable execution constraints. It adds 288 fresh GPT-4o mini run identities
and 288 Sol identities. The earlier Claude 288 is reused once after exact
core-file membership/hash, dependency/interpreter, runtime, task-content,
policy and exported-evidence checks. An application gate also verifies the
same owned GitLab version and scoped reader/writer contract. Claude remains
an earlier-time cohort, not a contemporaneously randomized model substitution.
The original 1,944-assignment comparison remains this paper's primary study.
The earlier two-worker GPT 288 and 1,512-assignment Sol campaign remain separate
(Appendices~\ref{app:operational-replications} and~\ref{app:sol-campaign}).

All three operational grids assign the same three previously examined
supplied-payload tasks, two native frameworks, eight conditions, three arms
and two repetitions: 288 per model, 96 per arm and 96 matched $G_o-N_o$ pairs.
The latest GPT and Sol stages execute serially, GPT first. All three profiles
use one worker, 16 model calls, 128 steps, 32,768 framed input bytes/call,
2,048 total output tokens/call, 180~s worker time, 60~s oversight bound and
zero SDK retries. Controller, guards, fault definitions, task bytes, scoring
and logical treatment declarations agree. The requested models are
\texttt{gpt-4o-mini-2024-07-18}, \texttt{gpt-6.1-sol} and
\texttt{claude-opus-5-5}. GPT-4o mini retains its qualified fresh Chat Completions
transport; Sol uses native Responses with low reasoning and consolidated
system messages; Claude uses its earlier Messages profile. Native reasoning,
serialization, seed/temperature semantics, realized compute and collection
time are not equalized. Common contracts do not justify a model-only ranking.

The analysis specified before collection makes material $G_o-N_o$ the primary contrast
\emph{within this development study}, strict completion secondary, and $E_o$
a baseline. All assignments, clean cases, blocked cases, unknowns, delivered
information, actual exposure and costs are retained. There is no success-based
progression gate or outcome-based choice of a better model report. No new
human labels or held-out task instances are supplied.

\paragraph{Information and automatic repairs.}
The fault/receipt boundary and $E_o/N_o/G_o$ meanings remain those in
Appendix~\ref{app:operational-replications}. In particular, $G_o$ adds
executor-retained receipt information as well as next-action advice;
$N_o$ is not information-equated. Automatic recovery handles FM13/FM18/FM20
without feedback offers, so their successes are not attributed to advice.
The same resource ceilings do not equalize realized calls or successful
continuation opportunity.

\begin{table*}[t]\centering\small
\setlength{\tabcolsep}{5pt}
\begin{tabular}{@{}lrrrrrrrrr@{}}\toprule
 & \multicolumn{3}{c}{GPT-4o mini} & \multicolumn{3}{c}{GPT-6.1 Sol}
 & \multicolumn{3}{c}{Claude (reused)}\\
Condition & $E_o$ & $N_o$ & $G_o$ & $E_o$ & $N_o$ & $G_o$ & $E_o$ & $N_o$ & $G_o$\\\midrule
Clean & 12 & 12 & 12 & 12 & 12 & 12 & 12 & 12 & 12\\
FM11 & 1 & 5 & 10 & 11$^{u}$ & 12 & 12 & 12 & 12 & 12\\
FM12 & 8 & 11 & 12 & 12 & 11$^{u}$ & 12 & 12 & 12 & 12\\
FM13 & 12 & 12 & 12 & 12 & 12 & 12 & 12 & 12 & 12\\
FM16 & 10 & 8 & 12 & 12 & 12 & 12 & 12 & 12 & 12\\
FM18 & 12 & 12 & 12 & 12 & 12 & 12 & 12 & 12 & 12\\
FM20 & 12 & 12 & 12 & 12 & 12 & 12 & 12 & 12 & 12\\
FM21 & 0 & 0 & 0 & 0 & 0 & 0 & 0 & 0 & 0\\
\bottomrule\end{tabular}
\caption{Latest operational comparison: material successes per 12 assigned
cases in every model/arm/condition. Superscript $u$ retains one provider-unknown
outcome in that cell: 11 known, rather than 12. It is not a model failure.
Strict differs from material only for GPT-4o mini $E_o/N_o$ on FM12 (7/7) and
FM16 (7/6); all other displayed cells have identical material/strict counts.
The earlier two-worker GPT grid is retained in
Table~\ref{tab:operational-conditions}; it is not pooled with this one.}
\label{tab:matched-operational-conditions}\end{table*}

\paragraph{Full assigned and complete-pair outcomes.}
All 576 new assignments are admitted and terminal: GPT has 288 known outcomes;
Sol has 286 known and two infrastructure unknowns. No assigned cell is unrun.
With the 288 reused Claude outcomes, the comparative record has 864 assignments
and 862 known outcomes. Table~\ref{tab:operational-replications} gives full-arm
material and strict counts; Table~\ref{tab:matched-operational-conditions}
keeps automatic recovery visible by condition.

GPT-4o mini material completion is $E_o$ 67/96, $N_o$ 72/96 (75.00\%) and
$G_o$ 82/96 (85.42\%): ten guided-only wins, no losses and 86 ties (+10.42~pp).
Strict completion is 63/66/82: 16 wins, no losses and 80 ties (+16.67~pp).
Five material gains occur on FM11 without restored issue references; one
occurs on FM12 and four on FM16, where references are part of the intervention.
This partitions conditions, not the causal contribution of receipt disclosure.
Two guided FM11 cases still fail. Neutral versus automatic has nine material
wins/four losses/83 ties (+5.21~pp); guided versus automatic has 15/zero/81
(+15.63~pp). These are new observations, not replacements of the earlier
GPT material +16.67~pp development result.

Sol has 83 successes/95 known in each of $E_o$ and $N_o$, with one additional
assigned unknown in each; $G_o$ has 84/96. All 95 known $G_o-N_o$ pairs tie.
The remaining unknown pair gives full-96-assignment identification bounds
[0,+1.04]~pp, not a confidence interval. Its task truth is not imputed as
failure. Claude has 84/96 in every arm and 96 ties. In both models, every known
clean/recoverable case already succeeds under the automatic baseline;
all 12 permission-blocked cases per arm remain incomplete. Thus no incremental
completion gain is observed on these Sol/Claude configurations,
without establishing equivalence or recovery on harder tasks.

\begin{table*}[t]\centering\small
\setlength{\tabcolsep}{4pt}
\begin{tabular}{@{}llrrrrrrr@{}}\toprule
Model & Endpoint & Complete pairs & Wins & Losses & Ties & Unknown & Difference (pp) & Assigned bounds (pp)\\\midrule
GPT-4o mini & Material & 96 & 10 & 0 & 86 & 0 & +10.42 & [+10.42,+10.42]\\
GPT-4o mini & Strict & 96 & 16 & 0 & 80 & 0 & +16.67 & [+16.67,+16.67]\\
GPT-6.1 Sol & Material & 95 & 0 & 0 & 95 & 1 & 0.00 & [+0.00,+1.04]\\
GPT-6.1 Sol & Strict & 95 & 0 & 0 & 95 & 1 & 0.00 & [+0.00,+1.04]\\
Claude Opus 5.5 & Material & 96 & 0 & 0 & 96 & 0 & 0.00 & [+0.00,+0.00]\\
Claude Opus 5.5 & Strict & 96 & 0 & 0 & 96 & 0 & 0.00 & [+0.00,+0.00]\\
\bottomrule\end{tabular}
\caption{$G_o-N_o$ contrasts with 96 pairs assigned per model/endpoint.
Differences use complete pairs; bounds retain every unknown pair. Bounds express
uncertainty due to missing outcomes, not sampling uncertainty. With only three related
authored tasks, the exploratory two-sided task-cluster sign values are .25
for GPT-4o mini and 1 for Sol/Claude; these do not establish population efficacy
or equivalence.}
\label{tab:matched-operational-pairs}\end{table*}

\paragraph{Actual fault exposure, delivery and semantic flags.}
Changed injection hooks occur in 252 GPT, 250 Sol and 252 Claude traces;
stricter boundary witnesses occur in 108 per model. These are distinct exposure
measures, not interchangeable denominators. Every arm has 60 automatic tool
retries: 12 rate-limit, 12 genuine-verification and 36 server-burst events.
There are no feedback offers in $E_o$. GPT and Claude each have 48 offers,
literal returned messages and accepted native-request bindings per $N_o/G_o$
arm. Sol has 47 in $N_o$ and 48 in $G_o$, all included verbatim in accepted model requests; the
interrupted neutral run does not receive feedback. Missing delivery therefore
does not explain a null on the 95 known Sol pairs.

Among feedback recipients, material success is GPT 24/48 versus 34/48,
Sol 35/47 versus 36/48, and Claude 36/48 in both. On recoverable
FM11/FM12/FM16 only, GPT is 24/36 versus 34/36; the remaining recipients face
persistent permission denial. Recipient subsets are descriptive, not the
full-assigned primary contrast. Across all six recoverable conditions, GPT
$N_o/G_o$ is 60/72 versus 70/72; clean and blocked cases remain in the main score.

The fresh GPT cohort retains original automated unsafe/silent flags 0/1;
Sol retains 0/0; Claude's reused 1/0 remains subject to the semantic concern
in Appendix~\ref{app:operational-replications}. GPT's silent flag belongs to an
$E_o$ FM16 case with correct material state but an incomplete strict contract.
No original label is rewritten. Author-run program checks reproduce 862 known
trace labels and directly reconstruct material state from fixed task JSON;
they do not provide independent human construct or safety validation.

\paragraph{Infrastructure unknowns and selected repeat attempts.}
The two Sol unknowns are native provider-request failures: CrewAI/$E_o$/FM11
on \texttt{audit\_export\_preservation}, and LangGraph/$N_o$/FM12 on
\texttt{dependency\_upgrade\_review}, both repetition 1. Their exact original
network/provider causes are unobserved. Each retains a conservative full-run
reservation of USD~1.376256 rather than a fabricated zero cost or task failure.

A separately authorized and fixed two-case follow-up repeats exactly these
configurations once, with the same model, recorded run seed 20261110,
repetition 1, task,
fault, arm, native profile and resource constraints, in fresh projects/run IDs.
Both material and strict completion succeed: 2/2 known, no unknown or unrun
case. The automatic FM11 repeat has two changed hook events and no feedback;
the neutral FM12 repeat has one changed hook and one feedback message included verbatim in an accepted model request
neutral message. Neither uses targeted advice. Both retain zero automated
unsafe/silent flags. The selected repeats demonstrate later completion of
the interrupted configurations, not an unbiased advice effect or a diagnosis
of the original request failure. The recorded seed is not sent as a portable
native provider seed; these API calls are not assumed deterministic.

Original unknowns, assigned bounds and denominators above remain unchanged.
Repeats are not substituted into the main comparison. Prelaunch representation
checks and a Docker-unavailable readiness stop admitted no paid run; their
zero-cost records remain. Docker Desktop and the existing GitLab container
were subsequently started with version/configuration/credential checks;
no image, task, policy or scoring changes were made. This adds a timing boundary
to these selected supplemental attempts.

\paragraph{Costs, provenance and interpretation.}
Recorded conservative charges are USD~0.55 GPT and USD~11.18 Sol, total
USD~11.73 for the 576 new assignments; the two selected repeats add USD~0.07.
These are overlapping-ledger reconciliations, not separately additive charges
per funding scope or provider invoices. Earlier Claude and historical GPT
costs remain in Appendix~\ref{app:operational-replications}. Original histories,
protected reserves, raw labels and source/evidence seals are preserved; all
new holds settle after worker drain. No additional Claude calls occur.

The full study and follow-up have distinct protocol identifiers
\texttt{f7e6a71e85090fe0f} and \texttt{f28599c1a64a97a2e}
(SHA-256 prefixes). Credential-free run records, traces, task inputs, native
binding projections, checksums and separate result audits retain their lineage;
private provider bodies, caches and ownership journals are excluded. These
author-run audits check consistency, not outside reproducibility or human labels.
The data support a configuration-specific GPT-4o mini development gain for the
information/advice bundle and no observed increment above the known-case
Sol/Claude baseline ceiling. They do not establish held-out transfer, isolated
advice-wording efficacy, a repaired earlier adverse result or a general model ranking.

\section{Separate Sol Campaign on the Paper Grids}
\label{app:sol-campaign}

\paragraph{Scope, inclusion and comparability.}
This completed source-fixed campaign extends the paper's task grids to a
separately configured native model. We report its full tested set, including
adverse, null and infrastructure-unknown outcomes. It remains separate from
the original primary comparison and from the later operational controller.
The requested model is \texttt{gpt-6.1-sol}. Its 1,512 assignments comprise a
fresh-grid 864 (the original three arms' 648 plus 216 joint-revision cells),
a 576-assignment known-task ladder and a 72-assignment component study.
All 1,512 were admitted: 1,509 known completed outcomes, three infrastructure
unknowns, none unrun. The separate 12-case integration pilot and inert advice
qualification are excluded. No paid outcome is replaced.

All 648 common fresh-grid and 576 ladder assignment keys match the original
task/framework/condition/arm/repetition grids, with matching task-pack hashes.
This is not a model-only substitution: Sol uses native Responses, low reasoning,
consolidated system messages, 4,096 total output tokens/call and a 240~s worker
bound, versus 2,048/180~s in the original comparison. Concurrency is two versus
four. Both retain 16 calls, 128 steps and 32,768 framed input bytes/call.
Sol sends no portable seed/temperature; original Chat Completions requests them.
Source freezes differ, without a full semantic-equivalence audit of changed
code. Within-Sol arms share one profile; historical cross-model rankings are
confounded by these source, transport, resource and time differences.

\begin{table}[t]\centering\small
\setlength{\tabcolsep}{3pt}
\begin{tabular}{@{}lrrr@{}}\toprule
Fresh-grid arm & Success/known & Unknown & Offers/bound\\\midrule
Previous O & 142/214 & 2 & 0/0\\
Neutral N & 161/216 & 0 & 35/35\\
Obligation G & 144/215 & 1 & 36/36\\
Joint revision J & 144/216 & 0 & 36/36\\\bottomrule
\end{tabular}
\caption{Separate Sol fresh grid: 216 assigned cases per arm; strict and
material success counts coincide. Bound denotes accepted native feedback
input, not comprehension. J is the prospectively specified tool-naming/stop-
clause revision, not the newer operational $G_o$. Unknowns remain assigned.}
\label{tab:sol-fresh}\end{table}

\paragraph{Adverse fresh-grid result and actual delivery.}
G$-$N has zero wins, 17 losses and 198 ties over 215 known of 216 assigned
pairs ($-$7.91~pp); logical bounds including missing outcomes are [$-$7.87,$-$7.41]~pp,
not sampling intervals. J$-$N has zero wins/17 losses/199 ties over all 216
pairs ($-$7.87~pp). J$-$G ties on all 215 complete pairs, with bounds
[$-$0.46,0.00]~pp. All 17 losses are server-burst FM20: N 17/36 versus G 0/36
and J 0/36. Sixteen losing pairs have native-bound continuation in both arms;
one has guided but no neutral offer. Aggregate localization does not identify
a behavioral mechanism or justify attributing unseen text an effect.

Clean, both 429 conditions and forged verification each score 36/36 in N/G/J,
without late offers. Permission revocation scores 0/36 in N/J and 0/35 known
in G, whose remaining case is infrastructure unknown. The other two unknowns
are in O. Across the whole fresh grid, all 107 offers have accepted native
bindings (N 35, G 36, J 36), all in FM20. Ceiling successes in other conditions
must not be credited to delivered advice.

\paragraph{Ladder and component outcomes without intervention.}
The ladder has 576/576 known outcomes: B 96/144, R 96/144, N 97/144, G 97/144,
with identical material counts and zero offers in every arm. G$-$N has
144 ties. N$-$B and N$-$R each have one win/zero losses/143 ties; because no
late advice occurs, these are trajectory differences, not message effects.
All six component arms have 12/12 strict/material successes, 72/72 known
in total, again with zero offers. These arms cannot identify the value of
removing the stop clause, naming tools or supplying neutral continuation.

\paragraph{Cost and inference limits.}
Stage conservative charges are USD 36.30 fresh, USD 19.16 ladder and USD 2.84
components; the campaign total including excluded pilot/qualification is
USD 58.67; the excluded pilot/qualification contributes USD 0.37. These are
recorded conservative charges, not invoices or wallet balances. Original
reports, source snapshots, unknowns and labels remain retained. The Sol study
tests the earlier guidance policies. The newer operational controller is
evaluated on a separate 288-assignment grid, not on this full 1,512-assignment
campaign (Appendix~\ref{app:matched-operational}). The adverse direction and no-offer ceilings add
coverage, not a formal model-family interaction, population effect, general
recovery improvement, or independent human validation.

\section{New Instances for the Recovery Policy and Removal of Issue IDs}
\label{app:operational-transfer}
This supporting study follows the operational development comparisons. It
retains the same controller, model snapshot, native frameworks, supplied-payload
task requirements, automatic policy and per-run resource ceilings. Six existing
authored instances unused by this controller are selected as the complete
complement of its three development task IDs within the earlier nine-instance
set. Their exact task bodies are unchanged: two instances per familiar
incident family. This is instance transfer within the same workflow, not a
new domain, blinded task design or independent validation. The original primary
comparison and its adverse results remain unchanged.

The run plan, fixed before collection, assigns 576 cases across six instances, two
frameworks, eight conditions, three arms and two repetitions. Another 48 cases
remove only issue references from guided feedback on FM12/FM16 in the same
randomized serial campaign. The internal rule for triggering feedback, automatic recovery, other
execution information and next-step advice remain unchanged.
The feedback formatter removes the issue IID and prevents it from being
passed through another argument. Padding with spaces keeps the original
message-size check unchanged. This is
a test of removing issue IDs, not a comparison that holds all information
constant and changes wording alone.
Normal guided controls are collected in the same campaign. Repetitions share
the declared native seed pair, without assuming deterministic provider output.

\paragraph{Run history and missing outcomes.}
The initial sandbox check admits no calls. A separate orchestration-order
failure also admits no calls. The next attempt admits two cases: one completes;
the second is an infrastructure unknown when transport validation rejects the
declared second-repetition seed. Its original label and conservative full
reservation remain. A new fixed transport option admits that exact seed
without changing the request payload, controller or task. Only the remaining
622 not previously started identities are collected under that continuation; neither
paid identity is repeated. The 624-assignment analysis combines disjoint
identities and preserves every infrastructure unknown and unrun case. This
operational interruption adds a timing/source boundary, not evidence of a model
failure or an outcome-dependent replacement.

The full registry has 624 admitted, 623 strict-known and 0 unrun assignments.
Material E/N/G success counts are 135/192, 142/192 and 161 successes plus one
unknown/192. Strict counts are 129/192, 134/192 and 161 plus one unknown/192.
The no-IID arm has 47/48 material and strict successes.
Table~\ref{tab:transfer-operational-conditions} reports conditions separately.
FM11 improves from 9/24 neutral to 20/24 guided successes without restored issue
references. FM12/FM16 account for nine further net material gains, with two
guided-only losses on FM16. Across their 48 assignments, normal guided and
no-IID guidance complete 46 and 47, respectively: zero guided-only wins, one
no-IID-only win and 47 ties. This small, near-ceiling contrast neither establishes
equivalence nor shows a reference benefit; it leaves other receipt information
and advice bundled.

The G--N material complete-pair difference is +10.47 pp over 191 pairs; the
all-assigned range [+9.90,+10.42] pp uses all 192 assignments and varies only
missing truth. These have different denominators and are not confidence
intervals. Exploratory task/family permutation values are .0625/.25 for
material and .03125/.25 for strict completion. Related instances, only three
incident families and multiple endpoints preclude a population-level efficacy
claim. The original primary study and its success measure are unchanged.

\begin{table*}[t]\centering\small
\setlength{\tabcolsep}{4pt}
\begin{tabular}{@{}llrrrr@{}}\toprule
Condition & Interpretation & $E_o$ & $N_o$ & $G_o$ & No-IID $G_o$\\\midrule
FM0 & Clean & 24/24 & 24/24 & 24/24 & --\\
FM11 & Request not sent & 5/5 & 9/9 & 20/20 & --\\
FM12 & Response lost after commit & 16/12 & 18/17 & 24/24 & 24/24\\
FM13 & Rate limit & 24/24 & 24/24 & 23/23$^{u1}$ & --\\
FM16 & Lost receipt + stale read & 18/16 & 19/12 & 22/22 & 23/23\\
FM18 & Forged verification & 24/24 & 24/24 & 24/24 & --\\
FM20 & Server-error burst & 24/24 & 24/24 & 24/24 & --\\
FM21 & Permission revoked & 0/0 & 0/0 & 0/0 & --\\
\bottomrule\end{tabular}
\caption{Material/strict success counts per 24 assigned E/N/G cases per condition;
cells contain two counts, not fractions.
The no-IID arm has 24 assignments on each lost-response condition only.
$u$ marks unknown outcomes, not failures. Persistent denial remains in coverage;
incompletion is not by itself an unsafe outcome. Strict additionally checks
reads, current genuine proof and final reporting, including the actual issue IID.}
\label{tab:transfer-operational-conditions}\end{table*}

\begin{table*}[t]\centering\small
\setlength{\tabcolsep}{4pt}
\begin{tabular}{@{}llrrrrll@{}}\toprule
Contrast & Endpoint & Known/A & W/L/T & Unknown & Diff. pp & Assigned bounds & $p_{task}/p_{family}$\\\midrule
$G_o-N_o$ & Material & 191/192 & 22/2/167 & 1 & +10.47 & [+9.90,+10.42] & 0.0625/0.2500\\
$G_o-N_o$ & Strict & 191/192 & 30/2/159 & 1 & +14.66 & [+14.06,+14.58] & 0.03125/0.25\\
$G_o-G_o^{\mathrm{noIID}}$ & Material & 48/48 & 0/1/47 & 0 & -2.08 & [-2.08,-2.08] & 1.0000/1.0000\\
$G_o-G_o^{\mathrm{noIID}}$ & Strict & 48/48 & 0/1/47 & 0 & -2.08 & [-2.08,-2.08] & 1.0000/1.0000\\
\bottomrule\end{tabular}
\caption{Matched contrasts. Differences use complete pairs; assigned bounds vary
missing truth and are not confidence intervals. Task and incident-family sign permutations are
exploratory tests under symmetry assumptions, not evidence of an effect
in a wider task population. Six related
instances do not provide six independent incident families. The IID contrast
retains other receipt information and measures more than advice wording.}
\label{tab:transfer-operational-contrasts}\end{table*}

\paragraph{Evidence, costs and limits.}
The author-run program verifies file hashes, nonoverlapping run identities, original
labels, material state reconstructed directly from task JSON, actual changed
tool calls actually affected by injection, feedback included verbatim in
accepted model requests, and removal of issue IDs.
Finite author program checks do not provide independent human construct or
safety validation. The study has no controller-off arm, so baseline success
does not identify the automatic controller's causal contribution. Delivery
subsets remain descriptive; the primary contrast retains all assigned pairs.

E/N/G/no-IID feedback offers, literal input deliveries and accepted native
bindings are respectively 0/96/96/48 for each measure. Changed-hook runs are
168/168/167/48, with 364/397/428/72 changed events; strict boundary-witness
runs are 72/72/71/0. Automatic retries are 120/120/119/0. The original
infrastructure interruption explains the missing guided exposure, not a
recovery failure. All clean cases complete (24/24 in each main arm); all
persistent-denial cases remain incomplete (0/24). Original automated unsafe
and silent-failure flags are zero across all arms; these are not human harm
judgments. All 48 no-IID offers omit the issue reference and retain their
verbatim delivery in accepted model requests.

The two collected source versions have protocol SHA-256 prefixes
\texttt{8b4cfe03c44888c8} and \texttt{813ce595a506291d}.
Original reports, closed admission/run record rows, task bodies, credential-free
traces, native binding projections, disjoint-key merge and author-run check retain
their checksums. Financial reconciliation confirms worker drain, settled
holds, unchanged historical prefixes and protected reserves; no further
provider calls or outcome retries are made during analysis.

Conservative lifetime charge is USD~1.30, including the original infrastructure reservation, under the USD~5 cap. Overlapping parent ledgers are not additive provider charges. Raw outcomes, source versions and all failed preparation/collection records are preserved; no paid outcome is replaced.

\end{document}